\documentclass{applemlr}

\usepackage{amsmath}
\usepackage{enumerate}
\usepackage{algorithm}
\usepackage{algpseudocode}
\usepackage{amsfonts}
\usepackage{amsthm}
\usepackage{cleveref}
\usepackage{diagbox}
\usepackage{colortbl}
\usepackage{amssymb}
\usepackage{xspace}
\usepackage{wrapfig}
\usepackage{adjustbox}
\usepackage{tabularx}
\usepackage{booktabs}
\usepackage{mathtools}
\usepackage{tikz}
\usepackage{enumitem}
\usepackage{silence}
\usepackage{dsfont}
\usepackage[table]{xcolor}
\usepackage[dvipsnames]{xcolor}
\usepackage{multirow}
\usepackage{makecell}
\usepackage{xfakebold}

\usepackage{amsmath,amsfonts,bm}

\def\eqref#1{equation~\ref{#1}}

\def\1{\bm{1}}

\DeclareMathAlphabet{\mathsfit}{\encodingdefault}{\sfdefault}{m}{sl}
\SetMathAlphabet{\mathsfit}{bold}{\encodingdefault}{\sfdefault}{bx}{n}

\definecolor{textgray}{HTML}{6E6E73}
\usetikzlibrary{positioning, calc}
\usetikzlibrary{decorations.pathmorphing}

\makeatletter
\patchcmd{\wrong@fontshape}{\@gobbletwo}{}{}{}
\makeatother
\numberwithin{equation}{section}
\makeatletter
\AtBeginDocument{
  \urlstyle{sf}
  
}
\makeatother

\definecolor{light}{RGB}{125, 125, 125}
\crefname{tcb@cnt@pbox}{code}{code}
\Crefname{tcb@cnt@pbox}{Code}{Code}
\crefname{assumption}{assumption}{assumption}
\Crefname{assumption}{Assumption}{Assumptions}

\newtcolorbox[auto counter]{pbox}[2][]{
  colback=white,
  title=Code~\thetcbcounter: #2,
  #1,fonttitle=\sffamily,
  fontupper=\sffamily,
  arc=2pt,
  colframe=bgcolor,
  coltitle=fgcolor,
  colbacktitle=bgcolor,
  toptitle=0.25cm,
  bottomtitle=0.125cm
}

\makeatletter
\newcommand\applefootnote[1]{%
  \begingroup
  \renewcommand\thefootnote{}%
  \renewcommand\@makefntext[1]{\noindent##1}%
  \footnote{#1}%
  \addtocounter{footnote}{-1}%
  \endgroup
}
\makeatother

\definecolor{cverbbg}{gray}{0.90}

\usepackage{nicefrac}
\usepackage[rightcaption]{sidecap}
\usepackage[normalem]{ulem}

\apptocmd{\appendix}{%
  \pretocmd{\section}{\FloatBarrier}{}{}%
  \pretocmd{\subsection}{\FloatBarrier}{}{}%
}{}{}

\title{Mix, Don't Tune: Bilingual Pre-Training Outperforms Hyperparameter Search in Data-Constrained Settings}

\author[1,2,*,\ddagger]{Paul Jeha}
\author[1]{Anastasiia Sedova}
\author[1]{Louis Bethune}
\author[1]{Skyler Seto}
\author[2,\dagger]{Jes Frellsen}
\author[1,\dagger]{Pierre Ablin}
\author[1,\dagger]{Natalie Schluter}

\affiliation[1]{Apple}
\affiliation[2]{DTU}

\contribution[\dagger]{Equal senior authorship}
\contribution[\ddagger]{Work done while at Apple}

\abstract{
For most languages of the world, language model pre-training operates in a data-constrained regime where models must repeat their training data many times, degrading generalization. Two remedies exist: aggressive hyperparameter tuning such as high weight decay, and mixing in data from a high-resource auxiliary language to directly aid the low-resource target. While hyperparameter tuning regularizes the model by shrinking weights to restrict network capacity, auxiliary data mixing uses a tunable mixing ratio to expand the training distribution and diversify the training signal with new knowledge. Both offer a principled way to improve training in a data-constrained domain. We compare these levers systematically across four model scales from 150M to 1.43B parameters, using Arabic as the low-resource target and English as the auxiliary, over approximately 1000 pre-training runs. Three findings emerge. First, mixing yields larger improvements than hyperparameter tuning on both validation loss and downstream task accuracy, and the gap grows with model size. Second, we quantify how much mixing helps: it boosts performance by an amount equivalent to 2--3$\times$ the unique target data on validation loss and 2--13$\times$ on downstream task accuracy, with the gain scaling steeply with model size. Third, this divergence reveals that target-language validation loss systematically underestimates mixing's value. Mixing regularizes by diversifying the training signal and contributes knowledge the repeated target corpus cannot supply; validation loss captures only the first effect. Our practical recommendations are: mix in a high-resource language, prioritize the mixing ratio over hyperparameter tuning, and transfer hyperparameters from a small proxy model via $\mu$P.
}

\metadata[Correspondence]{\sffamily Paul Jeha: \url{pauje@dtu.dk}}
\date{\sffamily\today}

\begin{document}

\maketitle

\section{Introduction}
\label{sec:intro}

As compute budgets grow faster than available text, language model pre-training increasingly operates in a data-constrained regime where models must repeat their training data many times \citep{muennighoff2025scalingdataconstrainedlanguagemodels}. Beyond a threshold, repetition actively damages model quality, disproportionately degrading generalization while shifting capacity toward memorization, with larger models hitting the degradation region at fewer repeated epochs \citep{hernandez2022scalinglawsinterpretabilitylearning}. Data scarcity is especially acute for low-resource languages \citep{penedo2025fineweb2}, but arises whenever a high-value corpus is too small to fill the compute budget. Prior work proposes two remedies: \citet{kim2025pretraininginfinitecompute} show that under extreme repetition, naive scaling produces a U-shaped loss curve as model size grows, but aggressive regularization (weight decay up to $30\times$ standard practice) restores monotonically decreasing scaling, while \citet{seto2025trainingbilinguallmsdata} show that mixing in data from a high-resource auxiliary language can boost target-language performance without explicitly modifying the model or objective.

These results suggest two levers for a practitioner training on scarce data: regularize harder to get more out of what you have, or bring in auxiliary data to diversify the training signal. The two mechanisms are not equivalent. Regularization curbs overfitting to the repeated target corpus; auxiliary data also adds capabilities the target corpus cannot teach. Validation loss, computed on a held-out sample of the same scarce target distribution, directly reflects the first effect. Throughout, we use ``regularization'' in this strict sense: an intervention that reduces overfitting to the target corpus, measured by target-language validation loss. Because proxy metrics can fail to track the behaviors practitioners care about \citep{theis2016noteevaluationgenerativemodels, huszar2015nottraingenerativemodel, nalisnick2019deepgenerativemodelsknow}, comparing the two remedies demands evaluating them on both validation loss and downstream performance. We do this across four model scales (150M--1.43B parameters), using Arabic as the low-resource target and English as the auxiliary, and ask three questions: (i) when pre-training with limited target-language data, which lever improves the model more, hyperparameter (HP) tuning or mixing in a high-resource language? (ii) to what extent can language mixing make up for the lack of target-language data? (iii) does target-language validation loss account for downstream performance, and if not, in what way?

\section{Related work}
\label{sec:related}

\paragraph{Data-constrained scaling laws.}
\citet{muennighoff2025scalingdataconstrainedlanguagemodels} generalize Chinchilla scaling laws to the data-constrained regime, showing that repeated data follows an exponential saturation curve with diminishing returns beyond roughly 4 epochs. \citet{hernandez2022scalinglawsinterpretabilitylearning} study partial repetition and find that it damages induction heads, shifting capacity from generalization to memorization. \citet{kim2025pretraininginfinitecompute} show that under extreme repetition, naive scaling produces a U-shaped loss curve as model size grows, but aggressive regularization (weight decay up to $30\times$ standard) restores monotonically decreasing scaling. \citet{gadre2024languagemodelsscalereliably} show that scaling laws extrapolate reliably in the over-training regime and can predict downstream performance. All four studies are English-only and do not consider cross-lingual data augmentation as a strategy for data efficiency. We add this dimension: bilingual mixing as a lever alongside regularization, and we quantify their relative contributions via variance decomposition.
\vspace{-1.5ex}
\paragraph{Bilingual and multilingual pre-training.}
Multilingual scaling laws have been studied at the level of language families \citep{he2024scalinglawsmultilinguallanguage}, in the curse-of-multilinguality regime across 250 languages \citep{chang2023multilingualitycurselanguagemodeling}, and via adaptive transfer laws spanning hundreds of languages \citep{longpre2026atlasadaptivetransferscaling}. These studies target compute-optimal sampling across many languages rather than the data-constrained bilingual regime. \citet{seto2025trainingbilinguallmsdata} train bilingual models from scratch with a data-constrained target language (German, 250M tokens) and English auxiliary, sweep several mixing ratios at 1B and 3B, and discuss how auxiliary data substitutes for fresh target data. Broader multilingual pre-training \citep{lin2022xglm, conneau2020unsupervisedcrosslingualrepresentationlearning} targets language coverage rather than controlled mixing analysis, and the continued pre-training approach of \citet{lin2024mala500massivelanguageadaptation} cannot isolate the value of bilingual mixing because it starts from a strong English model. We complement this line by decomposing variance against HP tuning, sweeping four model scales from 150M to 1.43B, formalizing the data multiplier, and identifying a divergence between validation loss and benchmark performance.
\vspace{-1.5ex}
\paragraph{Data mixture optimization.}
A growing literature optimizes domain mixture weights for English pre-training: DoReMi \citep{xie2023doremioptimizingdatamixtures} uses distributionally robust optimization, RegMix \citep{liu2025regmixdatamixtureregression} uses regression from small proxy models, \citet{ye2025datamixinglawsoptimizing} derive parametric mixing laws predicting loss as a function of domain proportions, and \citet{shukor2025scalinglawsoptimaldata} derive scaling laws for optimal data mixtures that extrapolate across scales and to unseen domain weights. These methods operate in a compute-constrained, many-domain setting and optimize the mixture weights. A complementary line of work studies dynamic schedules in which mixture proportions evolve during training \citep{liang2025boosting, dong2024abilitieslargelanguagemodels}, an axis orthogonal to the static-ratio analysis we conduct here. We ask a different question: in a data-constrained, bilingual setting with a single mixing parameter, how much value does mixing provide compared to hyperparameter tuning?
\vspace{-1.5ex}
\paragraph{Hyperparameter sensitivity analysis.}
Decomposing performance variance into per-factor contributions is the standard tool for hyperparameter importance \citep{pmlr-v32-hutter14, probst2018tunabilityimportancehyperparametersmachine, watanabe2023pedanovaefficientlyquantifyinghyperparameter}. The LLM literature has instead used scaling laws \citep{shukor2025scalinglawsoptimaldata, gadre2024languagemodelsscalereliably} and regression surrogates over mixture weights \citep{liu2025regmixdatamixtureregression, ye2025datamixinglawsoptimizing}, which predict the mean response surface but do not partition variance into main and interaction effects. We apply variance decomposition (classical ANOVA, since our grid is exhaustively trained) to ask whether a suboptimal choice of data mixing or hyperparameters costs more, and how the answer shifts with model scale.

\section{Experimental setup}
\label{sec:setup}

\paragraph{Data.}
We use Arabic as the low-resource (LR) target language and English as the high-resource (HR) auxiliary. The Arabic corpus is a subset of FineWeb2 \citep{penedo2025fineweb2}, which contains approximately 55B Arabic tokens. We subsample to $D_\text{LR} = 200\text{M}$ unique tokens, following the data-constrained regime of \citet{kim2025pretraininginfinitecompute}. The validation set is a disjoint subset of FineWeb2. The English corpus comes from FineWeb \citep{penedo2024fineweb} and is effectively unlimited at our training scales (approximately 350B tokens).
In the monolingual setting, models train on Arabic only; we fix the maximum number of times the Arabic data is repeated during training to $R_\text{max} = 100$, so every model sees $100 \times D_\text{LR}$ tokens regardless of scale. In the bilingual setting, each sample in a batch is drawn from Arabic with probability $\alpha$ and from English with probability $1 - \alpha$. We fix the total training budget to $D = 100 \times N_\text{nonemb}$ tokens, where $N_\text{nonemb}$ is the number of non-embedding parameters (2.3B, 12.2B, 28.5B, and 97.3B tokens for the 150M, 380M, 600M, and 1.43B models, respectively). This budget splits into an LR portion $\alpha D = R_\text{max} \, D_\text{LR}$ (Arabic repeated $R_\text{max}$ times) and an HR portion $(1-\alpha) D = D_\text{HR}$ (fresh English), so $\alpha = R_\text{max} \, D_\text{LR} / D$ ties the per-batch mixing fraction to the Arabic repetition budget $R_\text{max}$. We sweep $R_\text{max} \in \{1, 2, 5, 10, 20, 50, 100\}$. To quantify how much unique Arabic data bilingual mixing is equivalent to (\S\ref{sec:data_equiv}), we also train monolingual baselines at varying corpus sizes, subsampling FineWeb2 to create Arabic corpora of $D_\text{LR} \in \{25\text{M}, 50\text{M}, 100\text{M}, 200\text{M}, 500\text{M}, 1\text{B}, 2\text{B}\}$ tokens.
\vspace{-1.5ex}
\paragraph{Compute cost of mixing.}
We study the data-constrained regime where compute outruns unique target-language data \citep{muennighoff2025scalingdataconstrainedlanguagemodels, kim2025pretraininginfinitecompute}: the scarce resource is Arabic, not GPU hours, so our comparison is between repeating Arabic under heavy regularization and mixing in English on a fixed Arabic budget.
\vspace{-1.5ex}
\paragraph{Models.}
We train LLaMA-style decoder-only transformers at four scales (150M, 380M, 600M, 1.43B) with the XGLM tokenizer \citep{lin2022xglm} and sequence length 2048. Full architecture details are in App.~\ref{app:configs}.
\vspace{-1.5ex}
\paragraph{Hyperparameters.}
We optimize with AdamW using default $\beta$-values. We tune two hyperparameters, weight decay ($\lambda$) and learning rate ($\eta$), over the grid $\lambda \in \{0.01, 0.1, 0.3, 1.0, 3.0\}$ and $\eta \in \{10^{-4}, 3{\times}10^{-4}, 10^{-3}, 3{\times}10^{-3}, 10^{-2}\}$.
We use Maximal Update Parametrization ($\mu$P) \citep{yang2022tensorprograms} to tune once at a cheap proxy and transfer the base pair to larger scales. Concretely, a base $(\lambda, \eta)$ chosen at proxy width $d_\text{base} = 512$ (our 150M model) is rescaled at target width $d$ by $m_N = d/d_\text{base}$ as $\eta_{\mu\text{P}} = \eta/m_N$ and $\lambda_{\mu\text{P}} = \lambda \cdot m_N$ on hidden-layer parameters (App.~\ref{app:configs}). For example, transferring $(\lambda = 0.1, \eta = 0.01)$ from the 150M proxy to the 1.43B model ($d = 2048$, $m_N = 4$) yields effective values $\lambda_{\mu\text{P}} = 0.4$ and $\eta_{\mu\text{P}} = 0.0025$. Every grid cell below refers to a base $(\lambda, \eta)$ that $\mu$P rescales at the target scale.
We distinguish two HP regimes. \textit{Basic HP}: we run the $5 \times 5$ grid only on the 150M proxy, pick the best base $(\lambda, \eta)$ by Arabic validation loss ($\lambda = 0.1$, $\eta = 0.01$), and apply $\mu$P to transfer those values to every larger scale. \textit{Tuned HP}: we run the $\mu$P-scaled $5 \times 5$ grid \emph{at each model scale} and pick the best base $(\lambda, \eta)$ per scale; for example, bilingual training selects $(\lambda{=}0.01, \eta{=}0.01)$ at 380M and $(\lambda{=}0.1, \eta{=}0.003)$ at 1.43B (full per-paradigm values in Table~\ref{tab:optimal_hps}, App.~\ref{app:hp_grids}). In the bilingual setting, the per-scale grid expands to $5 \times 5 \times 7$ to also sweep $R_\text{max} \in \{1, 2, 5, 10, 20, 50, 100\}$. Tuned HP represents an upper bound on what HP tuning can achieve at each scale.
\vspace{-1.5ex}
\paragraph{Four paradigms.}
We traverse data strategies (monolingual vs.\ bilingual) and HP regimes (basic vs.\ tuned) to obtain four experimental paradigms (Table~\ref{tab:paradigms}).

\begin{table}[!htbp]
  \caption{Four experimental paradigms. $R$ (Arabic repetitions at a checkpoint) is read along every run via the constant learning rate, so it is implicit in all paradigms; the ``Sweeps'' column lists only the design axes varied across runs.}
  \label{tab:paradigms}
  \centering
  \begin{tabular}{llll}
    \toprule
    Paradigm & Data & Hyperparameters & Sweeps \\
    \midrule
    Monolingual, basic HP & Arabic only & $\mu$P from 150M & --- \\
    Monolingual, tuned HP & Arabic only & grid search per scale & $\lambda$, $\eta$ \\
    Bilingual, basic HP & Arabic + English & $\mu$P from 150M & $R_\text{max}$ \\
    Bilingual, tuned HP & Arabic + English & grid search per scale & $R_\text{max}$, $\lambda$, $\eta$ \\
    \bottomrule
  \end{tabular}
\end{table}
\vspace{-1.5ex}
\paragraph{Evaluation.}
Our primary metric is Arabic validation loss (VL). Each checkpoint corresponds to a different repetition count $R$, yielding a full curve from a single run. We wanted to verify model performance beyond word order prediction, i.e., knowledge and/or reasoning, but the limited capacities of the models we train here exclude benchmarks like MMLU \citep{hendrycks2021measuringmassivemultitasklanguage}. Moreover, to our knowledge no large-scale ``easy'' Arabic equivalent exists, a common gap for low-resource languages. We therefore use ARC Easy \citep{clark2018thinksolvedquestionanswering} for downstream evaluation, translating it from English to Arabic with NLLB-200-3.3B \citep{nllbteam2022languageleftbehindscaling}.

\section{Results}
\label{sec:results}

\subsection{Data mixing dominates hyperparameter tuning}
\label{sec:main_result}

\begin{figure}[!htbp]
  \centering
  \begin{minipage}[t]{0.48\linewidth}
    \centering
    \includegraphics[width=\linewidth]{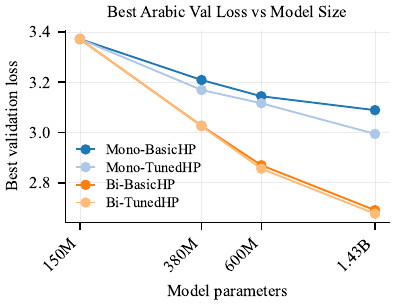}
  \end{minipage}
  \hfill
  \begin{minipage}[t]{0.48\linewidth}
    \centering
    \includegraphics[width=\linewidth]{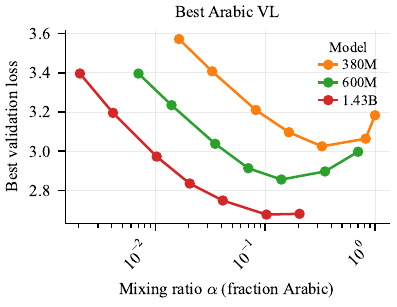}
  \end{minipage}
  \caption{\textbf{Left}: best Arabic validation loss vs.\ model size (blue: monolingual, orange: bilingual; dark: basic HP, light: tuned HP). Bilingual training separates sharply from monolingual as scale increases. \textbf{Right}: best Arabic VL vs.\ mixing ratio $\alpha$ (fraction Arabic) for each model scale, minimized over HPs and repetitions at each $(N, \alpha)$. The optimum shifts toward more English as scale grows ($\alpha^* = 0.33, 0.14, 0.10$ from 380M to 1.43B), and the basin visibly flattens at 1.43B around $\alpha \in [0.1, 0.2]$.}
  \label{fig:scaling_laws}
\end{figure}

We ask whether data mixing or hyperparameter tuning contributes more to Arabic language modeling performance. Our four paradigms (Table~\ref{tab:paradigms}) vary each factor independently, letting us isolate their effects on Arabic validation loss across scales. Our experiments provide strong empirical evidence that \textbf{data mixing is the dominant lever, and its advantage over HP tuning grows with model size} (Fig.~\ref{fig:scaling_laws}).

At 150M, all four paradigms collapse to the same run for two reasons. First, basic and tuned HP coincide at 150M by construction: basic HP is defined as the result of the $5 \times 5$ grid search at the 150M proxy, so the two regimes select the same $(\lambda, \eta)$ at this scale. Second, the 200M Arabic tokens suffice at 150M without overfitting, so the bilingual paradigms select $\alpha = 1$ (all Arabic) and reduce to the monolingual run. As models grow, mixing pulls sharply ahead. At 380M, mixing improves Arabic VL by 5.7\% over the monolingual baseline while HP tuning alone yields 1.2\%; at 600M, 8.7\% vs.\ 0.9\%; at 1.43B, 12.9\% vs.\ 3.1\%. HP tuning on top of mixing is negligible at every scale ($<$1\%). In concrete terms, a bilingual 600M model with $\mu$P HPs (VL 2.870) outperforms a monolingual 1.43B model with per-scale tuned HPs (VL 2.995), despite being 2.4$\times$ smaller.

The reason is that English data serves as an effective regularizer. Larger models overfit more aggressively on repeated Arabic data, and mixing in English counteracts this directly. The optimal Arabic fraction reflects this, declining from $\alpha = 0.33$ at 380M to $0.14$ at 600M and $0.10$ at 1.43B under tuned HPs (Fig.~\ref{fig:scaling_laws}, right). Once the model sees diverse data, there is little left for $\lambda$ and $\eta$ to improve.

Validation loss is an intermediate metric. The next section asks whether this ranking holds on downstream benchmarks, and then examines how sensitive bilingual training is to HP selection and whether practitioners can skip per-scale tuning.

\subsection{Benchmark performance across paradigms}
\label{sec:benchmarks}

\begin{figure}[!htbp]
  \centering
  \begin{minipage}[t]{0.48\linewidth}
    \centering
    \includegraphics[width=\linewidth]{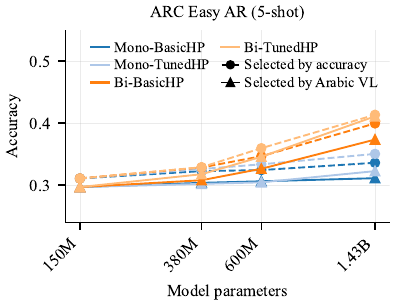}
  \end{minipage}
  \hfill
  \begin{minipage}[t]{0.48\linewidth}
    \centering
    \includegraphics[width=\linewidth]{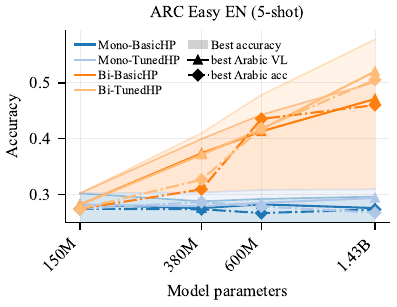}
  \end{minipage}
  \caption{ARC Easy accuracy (5-shot) vs.\ model size. \textbf{Left} (Arabic): solid = at best Arabic VL; dashed = peak Arabic (oracle). \textbf{Right} (English): solid = at best Arabic VL; dash-dot = at best Arabic accuracy; shaded = peak English (oracle). Blue: monolingual; orange: bilingual (dark = basic HP, light = tuned HP). Bilingual scales steeply on both languages, regardless of selection criterion.}
  \label{fig:benchmarks}
\end{figure}

Section~\ref{sec:main_result} established that mixing dominates HP tuning on Arabic validation loss. We now ask whether the same holds on downstream task accuracy. We evaluate all checkpoints on ARC Easy \citep{clark2018thinksolvedquestionanswering} in Arabic and English at 5-shot, selected as a knowledge-based benchmark that satisfies the four FineWeb reliability criteria \citep{penedo2024fineweb} (low seed variance, monotonic training curves, above-random signal, and discrimination between corpora of known quality) at small ablation scale. Figure~\ref{fig:benchmarks} reports accuracy under several selection rules: at the best Arabic VL checkpoint (solid), at the best Arabic accuracy checkpoint (English only, dash-dot), and at peak per-language accuracy (dashed left, shaded right). Per-language splits and full per-paradigm tables are in App.~\ref{app:detailed_results}.

\begin{figure}[!htbp]
  \centering
  \begin{minipage}[c]{0.66\linewidth}
    \begin{minipage}[t]{0.48\linewidth}
      \centering
      \includegraphics[width=\linewidth]{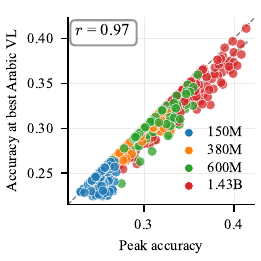}
    \end{minipage}\hfill
    \begin{minipage}[t]{0.48\linewidth}
      \centering
      \includegraphics[width=\linewidth]{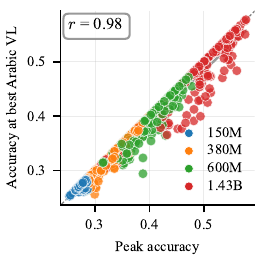}
    \end{minipage}
  \end{minipage}\hfill
  \begin{minipage}[c]{0.31\linewidth}
    \caption{Accuracy at the best Arabic VL checkpoint vs.\ peak accuracy (5-shot). \textbf{Left}: Arabic ARC Easy ($r = 0.97$). \textbf{Right}: English ARC Easy ($r = 0.98$). Dashed line: $y = x$. Solid line: linear fit. Colors indicate model scale. Points near the diagonal confirm that Arabic validation loss selects near-optimal checkpoints for both languages.}
    \label{fig:vl_proxy}
  \end{minipage}
\end{figure}

\vspace{-1.5ex}
\paragraph{Monolingual benchmark accuracy does not scale with model size.}
Despite the validation loss improvements observed in Section~\ref{sec:main_result}, monolingual benchmark accuracy is essentially flat as model size grows. Arabic accuracy rises from 29.7\% to only 31.1\% (basic HP) or 32.3\% (tuned HP) across a $10\times$ increase in parameters from 150M to 1.43B, all near the 25\% random baseline. English accuracy remains at random ($\sim$28\%) at every scale. Even with per-scale HP tuning, including weight decay up to $30\times$ standard practice \citep{kim2025pretraininginfinitecompute}, scaling a model on scarce repeated data does not translate into downstream improvement. The validation loss gains from aggressive regularization are real but do not carry over to benchmarks.
\vspace{-1.5ex}
\paragraph{Mixing scales steeply on both languages.}
Bilingual training with basic HPs ($\mu$P HPs from 150M) outperforms monolingual training with the same HPs by a margin that grows with scale: on Arabic ARC Easy, +0.4\% at 380M rising to +6.3\% at 1.43B; on English ARC Easy, +9.8\% at 380M rising to +19.5\% at 1.43B. Notably, a practitioner who mixes in English and selects checkpoints by Arabic VL obtains 47\% on English ARC Easy at 1.43B, compared to 28\% for the monolingual counterpart, without ever optimizing for English. English competence emerges as a side benefit of mixing in English data, at no additional tuning cost (though the compute budget is shared).
\vspace{-1.5ex}
\paragraph{HP tuning on top of mixing helps more on benchmarks than VL suggests.}
The value of HP tuning on top of mixing depends on the checkpoint selection rule. Under oracle selection (peak accuracy across all checkpoints), HP tuning adds little on Arabic: +1.3\% at 600M and +1.4\% at 1.43B. Under VL selection (accuracy at the best Arabic-VL checkpoint), the benefit grows with scale: +1.9\% at 600M and +3.7\% at 1.43B on Arabic; +4.9\% on English at 1.43B. Optimizing directly for English under accuracy selection yields up to +7.6\% at 1.43B (App.~\ref{app:detailed_results}). Section~\ref{sec:main_result} showed that HP tuning on top of mixing improves Arabic VL by under 1\% at every scale, so these benchmark gains are not reflected in VL. This is a first indication that Arabic VL underestimates downstream gains, a pattern \S\ref{sec:data_equiv} quantifies for mixing itself. Mixing, by contrast, improves both languages at every scale regardless of the selection criterion.
\vspace{-1.5ex}
\paragraph{Validation loss as a per-run checkpoint-selection proxy.}
The preceding comparison is \emph{across} runs, where VL gaps understate benchmark gaps between HP configurations. Within a single run, by contrast, VL remains a reliable checkpoint selector. Figure~\ref{fig:benchmarks} reports oracle (peak per-language accuracy) and VL-based (best Arabic VL checkpoint) selection, plus best-Arabic-accuracy selection on English. Figure~\ref{fig:vl_proxy} justifies the latter on a per-run basis: for each bilingual run, accuracy at the best-Arabic-VL checkpoint tracks peak accuracy of that run closely, with Pearson $r = 0.97$ (Arabic) and $0.98$ (English), and median absolute gaps of 0.21\% and 0.13\% ($n \approx 700$). Within a run, Arabic VL reliably picks near-optimal checkpoints, making it a cheap, label-free alternative to benchmark-based selection. Full statistics in Table~\ref{tab:vl_proxy_stats} (App.~\ref{app:benchmarks}).

Mixing is the only lever that scales downstream accuracy: monolingual training plateaus near random across a $10\times$ parameter sweep, while bilingual training scales steeply on both languages. HP tuning on top adds a second-order benefit partly invisible to Arabic VL, a pattern \S\ref{sec:data_equiv} returns to.

\begin{figure}[!htbp]
  \centering
  \includegraphics[width=\linewidth]{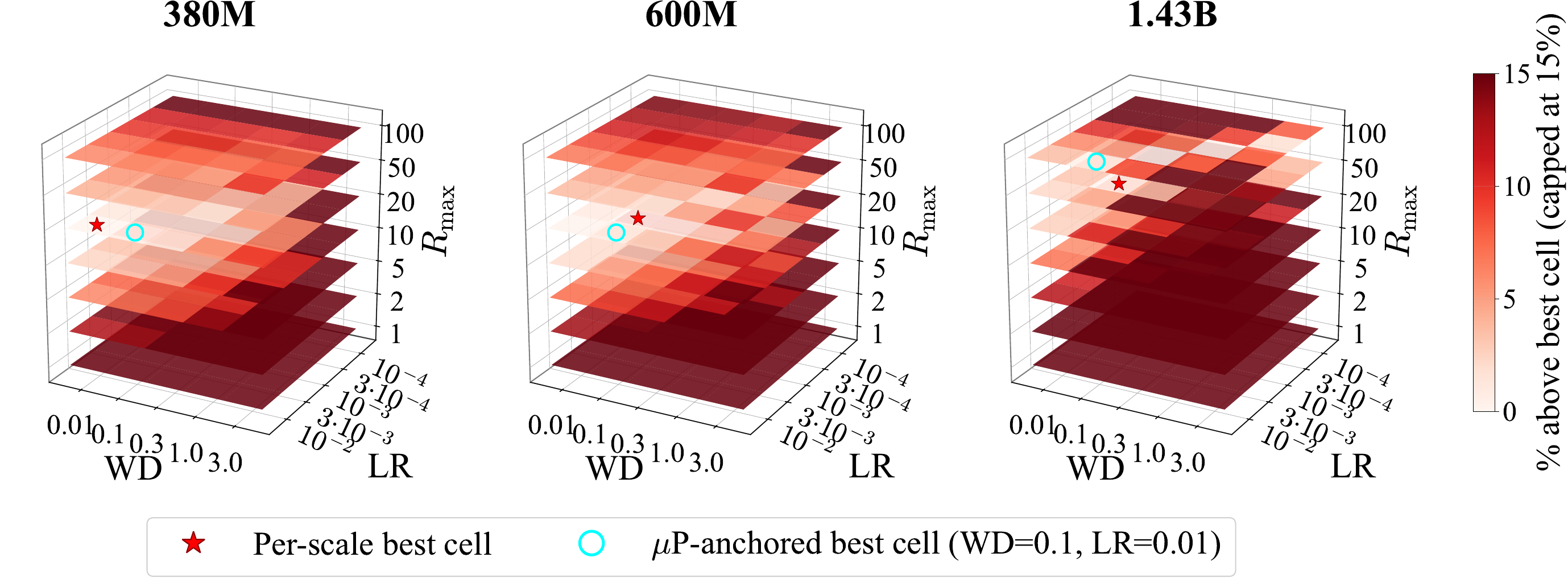}
  \caption{Arabic VL landscape over $(R_\text{max}, \lambda, \eta)$ at 380M, 600M, and 1.43B, as stacked $(\lambda, \eta)$ heatmaps along $R_\text{max}$ (color: \% above per-scale best, capped at 15\%). Red star: per-scale best; cyan circle: $\mu$P-anchored best ($\lambda{=}0.1$, $\eta{=}0.01$). Markers share a basin at 380M/600M but separate at 1.43B.}
  \label{fig:rmax_hp_heatmap}
\end{figure}

\subsection{Hyperparameter sensitivity and the case for $\mu$P}
\label{sec:hp_sensitivity}
\label{sec:analysis}
\label{sec:hp_sensitivity_alt}

The preceding sections establish that bilingual training outperforms monolingual training on both Arabic validation loss and downstream benchmarks. We now ask how sensitive bilingual performance is to each tuning axis, and whether the search over hyperparameters can be simplified.

The optimal configuration shifts with both scale and paradigm (Table~\ref{tab:optimal_hps}, App.~\ref{app:hp_grids}). For monolingual, tuned HP, the effective weight decay $\lambda_{\mu\text{P}}$ climbs sharply with scale, from 0.1 at 150M to 2.0 and 2.5 at 380M and 600M before settling at 1.2 at 1.43B, compensating for overfitting on repeated Arabic data. For bilingual, tuned HP, $\lambda_{\mu\text{P}}$ still grows with scale (0.1, 0.02, 0.25, 0.4 from 150M to 1.43B) but an order of magnitude more gently: \emph{on Arabic VL, mixing in English reduces the need for aggressive weight decay}. Under tuned HPs, the optimal Arabic fraction $\alpha$ drops monotonically from 0.33 at 380M to 0.10 at 1.43B as the per-scale HP search finds increasingly English-heavy regimes (Fig.~\ref{fig:alpha_decay}, App.~\ref{app:hp_grids}). Bilingual training with basic HP matches or beats monolingual training with tuned HP on Arabic validation loss (Table~\ref{tab:optimal_hps}), confirming that the diverse signal from English is a more effective VL regularizer than weight decay alone. A practitioner adopting bilingual training therefore faces a joint search over mixing ratio and hyperparameters. We ask two questions to determine how much of that search is actually necessary.
\vspace{-1.5ex}
\paragraph{Q1: Does the mixing ratio or the hyperparameters drive Arabic VL more?}
We ask whether varying $R_\text{max}$ or varying $(\lambda, \eta)$ produces more variation in Arabic validation loss in the bilingual grid. We use analysis of variance (ANOVA), which partitions the total variance in a metric across a grid of configurations into the contribution of each factor (App.~\ref{app:anova_method}). The answer is not a property of the loss landscape alone, but of the landscape viewed from where the practitioner stands. ``Around the operational point'' requires defining a region of HP space, and three reasonable definitions correspond to three practitioner states (we refer to Fig.~\ref{fig:rmax_hp_heatmap} throughout). \emph{(i) Top-$\gamma\%$ of the grid (post-hoc):} keep all HP cells whose mean Arabic VL is within $\gamma\%$ of the best, wherever they sit on the heatmap, and run ANOVA on those (PED-ANOVA's lens, \citet{watanabe2023pedanovaefficientlyquantifyinghyperparameter}); requires having trained the full grid first; results in Fig.~\ref{fig:recentering_stacked_bars}. \emph{(ii) Box around the per-scale optimum (idealized):} start at the red star, expand outward through the $(\lambda, \eta)$ grid one cell at a time, and run ANOVA on each successive box; requires knowing the optimum, which is what we are trying to avoid tuning; results in Fig.~\ref{fig:expanding_grid_2way_stacked_bars}, bottom row. \emph{(iii) Box around $\mu$P (a-priori):} same construction anchored at the cyan circle (the basic HP $(\lambda{=}0.1, \eta{=}0.01)$, selected at the 150M proxy), the only view a practitioner can build before any per-scale training; results in Fig.~\ref{fig:expanding_grid_2way_stacked_bars}, top row.

\begin{SCfigure}[][!htbp]
  \centering
  \includegraphics[width=0.6\linewidth]{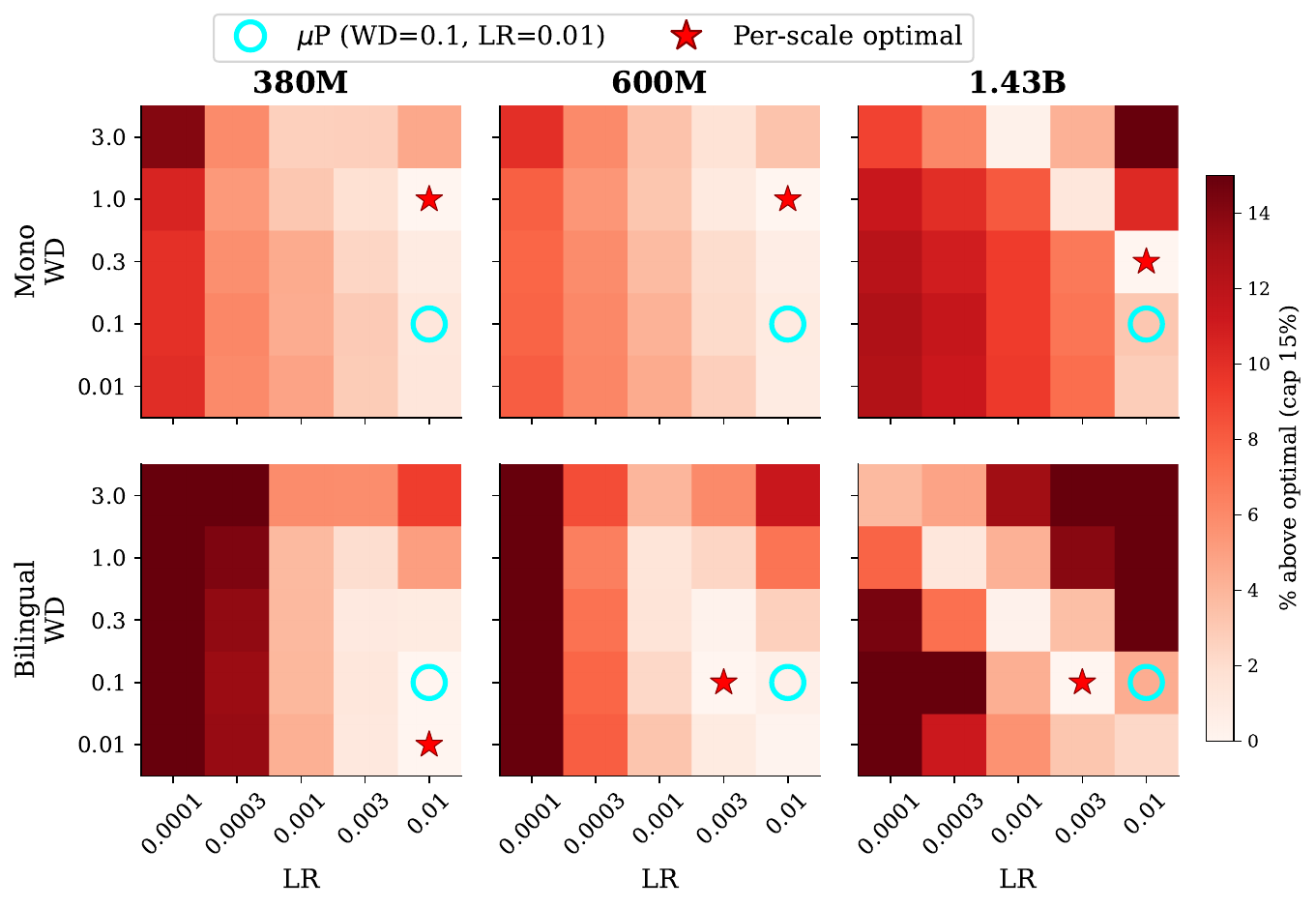}
  \caption{HP landscape (relative gap \% to per-scale optimum). Top: monolingual; bottom: bilingual. Cyan circle: $\mu$P HPs; red star: per-scale optimal. At 380M/600M, uniformly pale cells indicate a flat landscape where HP choice is forgiving. At 1.43B, dark cells emerge throughout the grid, signaling sharper sensitivity. The $\mu$P cell stays within 0.6\% (bilingual) and 3.1\% (monolingual) of optimal at all scales. Gaps quantified in App.~\ref{app:hp_landscape}.}
  \label{fig:mup_transfer}
\end{SCfigure}

\emph{View (i), threshold-based.} At every scale and across the practical range of thresholds, $R_\text{max}$ explains the majority of Arabic VL variance (Fig.~\ref{fig:recentering_stacked_bars}). At 1.43B the dominance weakens only at the loosest thresholds; tightening the threshold even slightly restores it. The full-grid view, sensitive to grid design, is reported separately in App.~\ref{app:hp_landscape}. \emph{Views (ii) and (iii), box around the optimum vs around $\mu$P.} At 380M and 600M, the red star and cyan circle sit in the same wide low-loss basin (Fig.~\ref{fig:rmax_hp_heatmap}, left and middle), so the two views agree: $R_\text{max}$ dominates at small radii around either center ($>$90\% in the smallest neighborhood). The practitioner using $\mu$P faces the same low HP sensitivity as the one who has tuned per-scale. At 1.43B, the basin fragments and the two anchors separate (Fig.~\ref{fig:rmax_hp_heatmap}, right). Standing at the red star, $R_\text{max}$ still dominates ($\approx$89\% in the smallest optimum-centered box). Standing at the cyan circle, HP variance and interactions account for roughly half of the variance in the smallest neighborhood: at 1.43B the loss landscape around $\mu$P is sharper than at smaller scales, with more variance over short HP distances (Fig.~\ref{fig:mup_transfer}).

At 380M and 600M, $\mu$P sits near the per-scale optimum in a flat basin and mixing dominates. At 1.43B, $\mu$P is still close to the optimum but lands in a region of higher local HP sensitivity. Q2 measures the resulting gap to the per-scale optimum and whether it justifies per-scale tuning.

\sidecaptionvpos{figure}{c}
\sidecaptionvpos{figure}{t}
\vspace{-1.5ex}
\paragraph{Q2: How robust is $\mu$P transfer?}
Q1 recommended $\mu$P. The cost of that recommendation is minimal: transferring the best 150M proxy HPs to larger scales via $\mu$P keeps the relative gap to the per-scale tuned optimum below 0.6\% on bilingual Arabic VL at every scale up to 1.43B (Fig.~\ref{fig:mup_transfer_gap}, left, App.~\ref{app:hp_landscape}). The bilingual gap is 0.1\% at 380M and 0.5\% at both 600M and 1.43B; the monolingual gap is larger but still bounded: 1.3\% at 380M, 0.9\% at 600M, 3.1\% at 1.43B. The heatmaps (Fig.~\ref{fig:mup_transfer}) show this: the $\mu$P cell sits in a flat, near-optimal basin at 380M and 600M; at 1.43B the basin shrinks but the $\mu$P cell remains within a few percent of the per-scale optimum (App.~\ref{app:operational_sensitivity}).

The practical recipe is therefore to transfer HPs via $\mu$P from a small proxy and spend compute on $R_\text{max}$ search. The alternative, a wide HP search at scale, exposes the practitioner to high-variance, jagged regions of HP space where the cost of a wrong choice grows substantially (App.~\ref{app:operational_sensitivity}). Practitioners with extra compute at 1.43B may invest in per-scale HP refinement, but the marginal return is uncertain compared to mixing-ratio tuning.

\subsection{To what extent can mixing compensate for data scarcity?}
\label{sec:data_equiv}

\begin{SCfigure}[][!htbp]
  \centering
  \includegraphics[width=0.45\linewidth]{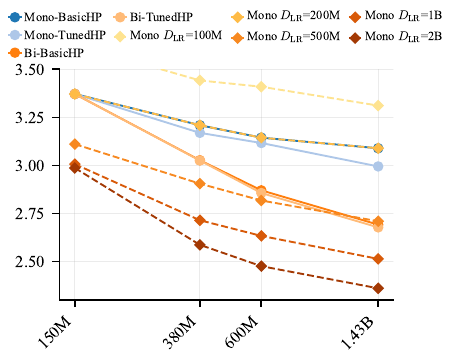}
  \caption{Best Arabic validation loss vs.\ model size. Dashed lines (yellow to brown): monolingual training at each Arabic corpus size $D_\text{LR}$ (100M--2B). Solid lines: the four paradigm results at $D_\text{LR} = 200\text{M}$ (blue: monolingual, orange: bilingual; dark: basic HP, light: tuned HP). On VL, more unique Arabic data consistently beats mixing: the mono $D_\text{LR}{=}500\text{M}$ curve already sits below bilingual, tuned HP at 600M and above.}
  \label{fig:data_equiv}
\end{SCfigure}

The preceding sections establish that data mixing outperforms monolingual training at $D_\text{LR} = 200\text{M}$ tokens, and that a practitioner using $\mu$P HPs should prioritize the mixing ratio over HP tuning. We now quantify the value of mixing: how much unique target-language data is it equivalent to?
\vspace{-1.5ex}
\paragraph{Building the baseline.}
We train monolingual models with $\mu$P HPs at seven Arabic corpus sizes: $D_\text{LR} \in \{25\text{M}, 50\text{M}, 100\text{M}, 200\text{M}, 500\text{M}, 1\text{B}, 2\text{B}\}$ tokens, each repeated up to 100 times. Figure~\ref{fig:data_equiv} plots the best Arabic validation loss against model size, with one curve per corpus size and the four paradigm results overlaid. On validation loss, more unique data dominates: 2.5$\times$ more Arabic data alone already matches the best bilingual result at 600M.

\vspace{-1.5ex}
\paragraph{The data multiplier.}
To quantify the value of mixing, we define the \emph{data multiplier} $N = D_\text{LR}^\text{equiv} / 200\text{M}$: how much unique Arabic data a monolingual model would need to match a bilingual model trained on only 200M Arabic tokens. We compute $N$ by fitting a log-linear model to the monolingual baseline sweep and inverting it at the bilingual target (App.~\ref{app:data_equiv}). We report $N$ under two conditions: a \emph{controlled} comparison where both models use identical hyperparameters ($\lambda{=}0.1$, $\eta{=}0.01$), isolating the pure effect of mixing, and an \emph{optimized} comparison where the bilingual model additionally gets a full HP sweep over $\lambda \times \eta \times R_\text{max}$.

\begin{figure}[!htbp]
  \centering
  \begin{minipage}[t]{0.48\linewidth}
    \centering
    \includegraphics[width=\linewidth]{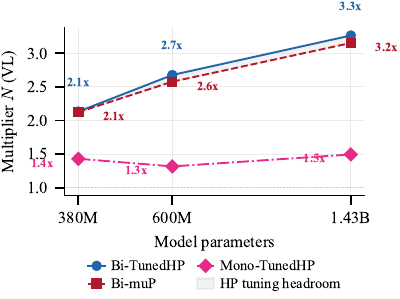}
  \end{minipage}
  \hfill
  \begin{minipage}[t]{0.48\linewidth}
    \centering
    \includegraphics[width=\linewidth]{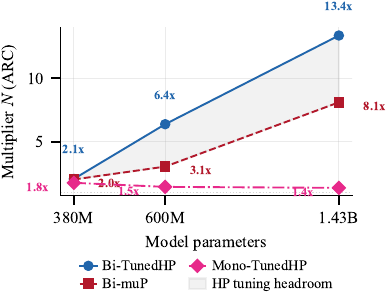}
  \end{minipage}
  \caption{Data multiplier $N$ vs.\ model size on Arabic VL (left) and ARC Easy Arabic 5-shot (right), for bilingual+tuned HPs (blue), bilingual+$\mu$P (red), monolingual+tuned HPs (pink). VL multipliers grow slowly ($1.3$–$3.3\times$); on the benchmark, bilingual reaches $13.4\times$ while monolingual stays at $1.4$--$1.8\times$, exposing the mixing gain VL underreports. App.~\ref{app:multiplier_vl}: VL-based checkpoint selection.}
  \label{fig:multiplier_comparison}
\end{figure}

The main finding is a striking divergence between validation loss and benchmarks. On VL, the multiplier is modest: $2.1\times$ at 380M, $2.6\times$ at 600M, $3.2\times$ at 1.43B under controlled HPs (Fig.~\ref{fig:multiplier_comparison}, left). On benchmarks, it is far larger and grows steeply with scale: $2.0\times$ at 380M, $3.1\times$ at 600M, $8.1\times$ at 1.43B (Fig.~\ref{fig:multiplier_comparison}, right). At 1.43B, bilingual training on 200M Arabic tokens matches the benchmark accuracy of a monolingual model trained on 1.6B unique Arabic tokens. With HP tuning on top of mixing, the multiplier reaches $13.4\times$ at 1.43B, equivalent to 2.7B unique Arabic tokens.

The benchmark multiplier increasingly exceeds the VL multiplier with scale (Fig.~\ref{fig:multiplier_comparison}): at 1.43B on ARC Easy, monolingual tuning is worth only ${\sim}1.5\times$ unique Arabic data, bilingual $\mu$P jumps to $8.1\times$, and HP tuning on top pushes the multiplier to $13.4\times$. On VL the same three paradigms span only $1.3$--$3.3\times$: validation loss saturates quickly while benchmarks still have headroom, so VL credits only a fraction of the gain mixing actually delivers. We explore this discrepancy below.

\vspace{-1.5ex}
\paragraph{The benchmark flip.}
\label{sec:flip}

\begin{SCfigure}[][!htbp]
  \centering
  \includegraphics[width=0.55\linewidth]{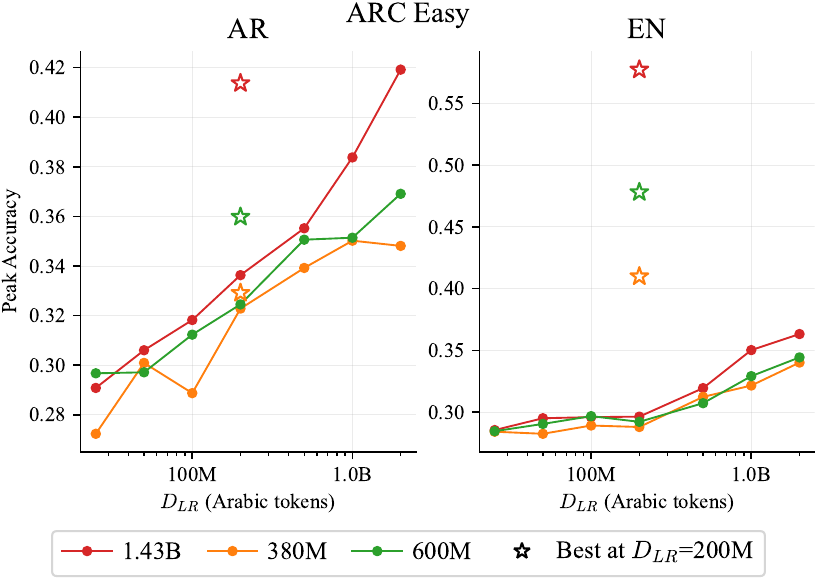}
  \caption{Peak ARC Easy accuracy (5-shot) vs.\ Arabic data budget ($D_\text{LR}$). Lines: monolingual training at each model scale. Stars: best bilingual configuration at $D_\text{LR} = 200\text{M}$ with per-scale tuned HPs. \textbf{Left} (Arabic): bilingual stars approach the high-$D_\text{LR}$ end of each monolingual curve; at 1.43B, bilingual comes within 0.6 points of monolingual at $D_\text{LR} = 2\text{B}$, which uses 14$\times$ more Arabic. \textbf{Right} (English): bilingual dominates at every $D_\text{LR}$. The $\mu$P variant (App.~\ref{app:multiplier_variants}, Fig.~\ref{fig:flip_mup_5shot}) shows the same pattern with slightly lower bilingual accuracy on Arabic.}
  \label{fig:flip}
\end{SCfigure}

On validation loss, more unique Arabic data always beats mixing (Fig.~\ref{fig:data_equiv}). On benchmarks, the ranking flips: mixing wins decisively, even against monolingual models trained on far more Arabic data (Fig.~\ref{fig:flip}). At 1.43B, monolingual training sees 18.4B Arabic tokens and reaches 33.6\% accuracy; bilingual training sees only 9.2B Arabic tokens and reaches 41.4\%. For English, the gap is even larger: 57.7\% vs.\ 29.6\%.

A plausible explanation is that the two metrics reward different things. Repetition lets the model overfit to in-distribution patterns, which lowers VL on a held-out sample of the same distribution without improving general capabilities. Benchmarks require factual knowledge largely absent from 200M Arabic tokens but abundant in English. Mixing supplies it; repetition cannot.

English data therefore plays two complementary roles \citep{seto2025trainingbilinguallmsdata}: it \emph{regularizes} by diversifying the training signal, and it \emph{adds knowledge} the Arabic corpus lacks. VL appears to capture only the first; benchmarks capture both, which explains why the benchmark multiplier (2--13$\times$) far exceeds the VL multiplier (2--3$\times$). Practically, VL is a reasonable proxy \emph{within} a paradigm but cannot rank monolingual and bilingual models against each other.

\section{Conclusion}
\label{sec:conclusion}

We studied how data mixing, hyperparameter tuning, and data scaling compare when pre-training language models on scarce target-language data. Three findings emerge. First, mixing in a high-resource language contributes more than hyperparameter tuning to Arabic performance at 380M+, and $\mu$P transfer from a small proxy model keeps the HP cost near zero. Second, mixing amplifies scarce data by 2--3$\times$ on validation loss and 2--13$\times$ on downstream task accuracy, equivalent to up to 2.7B unique Arabic tokens at 1.43B. The compute cost is real, but it replaces Arabic repetitions that would otherwise be wasted on overfitting. Third, validation loss undercounts mixing's value: it credits the regularization effect but not the knowledge the auxiliary language brings in.
\vspace{-1.5ex}
\paragraph{Practical recipe.}
For practitioners training on scarce target-language data: (1) mix in a high-resource auxiliary language and prioritize the mixing ratio over hyperparameter tuning; (2) transfer hyperparameters from a small proxy model via $\mu$P; (3) do not use validation loss alone to compare monolingual and bilingual strategies.

\bibliography{references}
\bibliographystyle{plainnat}

\appendix

\section{Learning rate schedule}
\label{app:constant_lr}

All experiments use a constant learning rate with no warmup and no decay. Constant or near-constant schedules are established practice in recent pre-training work: \citet{hägele2024scalinglawscomputeoptimaltraining} show that constant LR with a short cooldown matches cosine at matched compute; \citet{porian2025resolvingdiscrepanciescomputeoptimalscaling} reproduce the Chinchilla scaling law under a constant schedule, finding that learning-rate decay is not essential for its validity; and \citet{yano2026pretrainingllmlearningrate} pre-train 1B and 8B models under a constant LR through the end of training.

Our design also requires it. The central object of analysis is a loss-vs-repetition curve read along a single training run (Figures~\ref{fig:scaling_laws},~\ref{fig:data_equiv}). A decay schedule couples the learning rate to the training horizon, so checkpoints at different repetition counts $R$ sit at different points on the schedule and are not directly comparable within a run. Matching our sweep under a decay schedule would require one run per repetition point, multiplying compute by the length of the $R$ axis.

The cost is that absolute loss would be lower under a schedule with a cooldown. All four paradigms share the same constant schedule, so cross-paradigm comparisons and the relative conclusions reported in the paper are unaffected.

\section{Model architectures and training configurations}
\label{app:configs}

All models are LLaMA-style decoder-only transformers with SiLU activations, RMSNorm ($\epsilon = 10^{-5}$), rotary position embeddings (RoPE, base 10{,}000), and tied input/output embeddings. We use the XGLM tokenizer \citep{lin2022xglm} with a 256K vocabulary. Table~\ref{tab:arch} summarizes the architecture at each scale.

\begin{table}[h]
  \caption{Model architectures. All models use a head dimension of 128 and a context length of 2048 tokens.}
  \label{tab:arch}
  \centering
  \begin{tabular}{lcccc}
    \toprule
    & 150M & 380M & 600M & 1.43B \\
    \midrule
    Layers          & 6    & 8     & 12    & 16 \\
    Hidden dim      & 512  & 1024  & 1280  & 2048 \\
    Attn heads      & 4    & 8     & 10    & 16 \\
    KV heads        & 4    & 8     & 10    & 8 \\
    FFN dim         & 1792 & 3584  & 4480  & 7168 \\
    Parameters      & 154M & 384M  & 613M  & 1.43B \\
    \bottomrule
  \end{tabular}
\end{table}

\paragraph{Optimizer.} All models use AdamW with $\beta_1 = 0.9$, $\beta_2 = 0.95$, $\epsilon = 10^{-8}$, gradient clipping at norm 1.0, and bfloat16 precision. We use a constant learning rate schedule. An auxiliary z-loss with weight $10^{-4}$ stabilizes training.

\paragraph{$\mu$P scaling.} We use maximal update parameterization with a base dimension of 512 (the 150M model). The scaling factor $m_N = d_\text{model} / 512$ adjusts the learning rate ($\eta' = \eta_\text{base} / m_N$) and weight decay ($\lambda' = \lambda_\text{base} \times m_N$) for hidden-layer parameters. Base hyperparameters are $\eta = 10^{-2}$ and $\lambda = 0.1$, tuned at the 150M proxy scale and transferred to larger models.

\section{Per-paradigm detailed results}
\label{app:hp_grids}

\subsection{Optimal hyperparameters and Arabic fraction}

Table~\ref{tab:optimal_hps} reports the optimal hyperparameters for each paradigm and model scale, selected by best Arabic validation loss. All values are $\mu$P base hyperparameters (before scaling by $m_N$).

\begin{table}[h]
  \caption{Optimal hyperparameters per paradigm and model scale (Arabic validation loss). $\lambda$ and $\eta$ are $\mu$P base values; $\lambda_{\mu\text{P}}$ and $\eta_{\mu\text{P}}$ are the effective values after $\mu$P scaling ($\eta_{\mu\text{P}} = \eta / m_N$, $\lambda_{\mu\text{P}} = \lambda \times m_N$, where $m_N = d_\text{model}/512$). $R^*$: repetition count at the best checkpoint. $R_\text{max}$: maximum repetition budget (bilingual only). $\alpha$: Arabic fraction at the selected $R_\text{max}$ (bilingual only). All 150M bilingual runs select $\alpha = 1$ (monolingual), confirming that mixing does not help at this scale.}
  \label{tab:optimal_hps}
  \centering
  \footnotesize
  \begin{tabular}{llcccccccc}
    \toprule
    Paradigm & Scale & $\lambda$ & $\lambda_{\mu\text{P}}$ & $\eta$ & $\eta_{\mu\text{P}}$ & $R^*$ & $R_\text{max}$ & $\alpha$ & Loss \\
    \midrule
    \multirow{4}{*}{Monolingual, basic}
      & 150M  & 0.1 & 0.1  & 0.01  & 0.01    & 69 & --- & --- & 3.372 \\
      & 380M  & 0.1 & 0.2  & 0.01  & 0.005   & 27 & --- & --- & 3.209 \\
      & 600M  & 0.1 & 0.25 & 0.01  & 0.004   & 21 & --- & --- & 3.144 \\
      & 1.43B & 0.1 & 0.4  & 0.01  & 0.0025  & 7  & --- & --- & 3.089 \\
    \midrule
    \multirow{4}{*}{Monolingual, tuned}
      & 150M  & 0.1 & 0.1  & 0.01  & 0.01    & 69 & --- & --- & 3.372 \\
      & 380M  & 1.0 & 2.0  & 0.01  & 0.005   & 47 & --- & --- & 3.169 \\
      & 600M  & 1.0 & 2.5  & 0.01  & 0.004   & 39 & --- & --- & 3.116 \\
      & 1.43B & 0.3 & 1.2  & 0.01  & 0.0025  & 23 & --- & --- & 2.995 \\
    \midrule
    \multirow{4}{*}{Bilingual, basic}
      & 150M  & 0.1 & 0.1  & 0.01  & 0.01    & 69 & 100 & 1.00 & 3.372 \\
      & 380M  & 0.1 & 0.2  & 0.01  & 0.005   & 20 & 20  & 0.33 & 3.027 \\
      & 600M  & 0.1 & 0.25 & 0.01  & 0.004   & 20 & 20  & 0.14 & 2.870 \\
      & 1.43B & 0.1 & 0.4  & 0.01  & 0.0025  & 77 & 100 & 0.21 & 2.692 \\
    \midrule
    \multirow{4}{*}{Bilingual, tuned}
      & 150M  & 0.1 & 0.1  & 0.01  & 0.01    & 69 & 100 & 1.00 & 3.372 \\
      & 380M  & 0.01& 0.02 & 0.01  & 0.005   & 20 & 20  & 0.33 & 3.025 \\
      & 600M  & 0.1 & 0.25 & 0.003 & 0.0012  & 20 & 20  & 0.14 & 2.856 \\
      & 1.43B & 0.1 & 0.4  & 0.003 & 0.00075 & 50 & 50  & 0.10 & 2.678 \\
    \bottomrule
  \end{tabular}
\end{table}

\begin{figure}[h]
  \centering
  \includegraphics[width=0.6\linewidth]{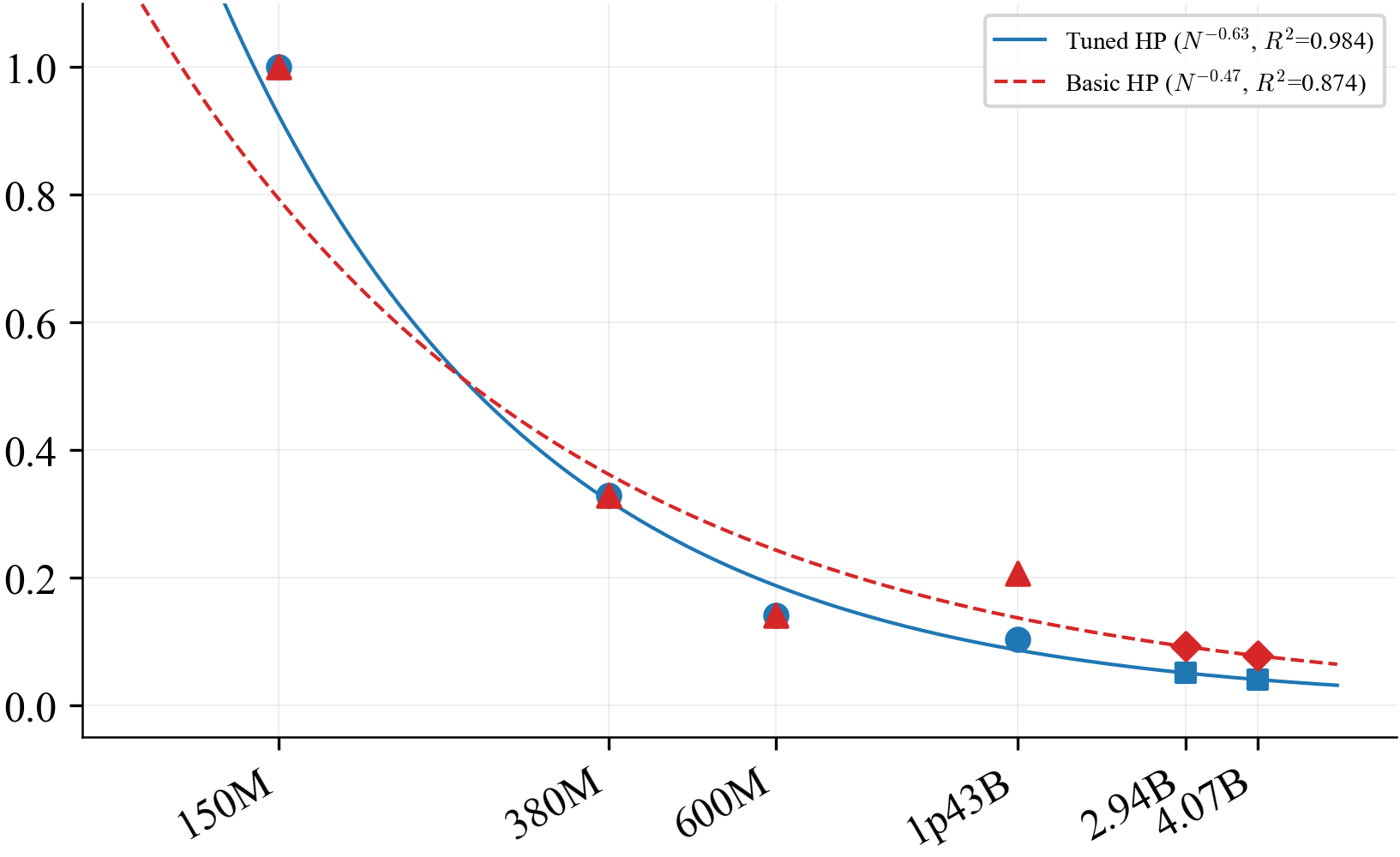}
  \caption{Optimal Arabic fraction $\alpha^*$ vs.\ model size. Both HP regimes follow a power-law decay, with the tuned-HP curve declining faster ($\beta = 0.63$) than basic HP ($\beta = 0.47$). Extrapolations to 2.94B and 4.07B shown.}
  \label{fig:alpha_decay}
\end{figure}

Table~\ref{tab:optimal_hps} reports the configurations that minimize Arabic validation loss. Figure~\ref{fig:alpha_decay} visualizes the optimal Arabic fraction $\alpha^*$ from this table as a function of model size, showing a power-law decay in both HP regimes.

\subsection{Optimizing for English instead of Arabic}

\begin{figure}[h]
  \centering
  \begin{minipage}[t]{0.48\linewidth}
    \centering
    \includegraphics[width=\linewidth]{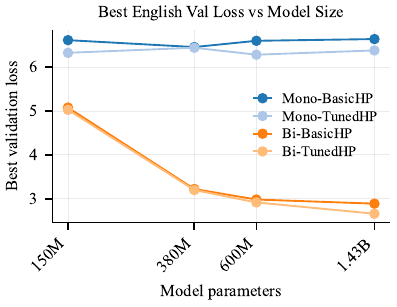}
    \caption{Best English validation loss vs.\ model size for each paradigm, selecting the checkpoint and HPs that minimize English VL. Monolingual Arabic models (top two curves) show essentially flat English loss across scales (6.3--6.6), since they never see English tokens. Bilingual models improve English dramatically with scale, from $\sim$5.1 at 150M to 2.66 at 1.43B with tuned HPs. HP tuning provides a modest additional gain in the bilingual setting (2.89 $\to$ 2.66 at 1.43B).}
    \label{fig:english_val_scaling}
  \end{minipage}
  \hfill
  \begin{minipage}[t]{0.48\linewidth}
    \centering
    \includegraphics[width=\linewidth]{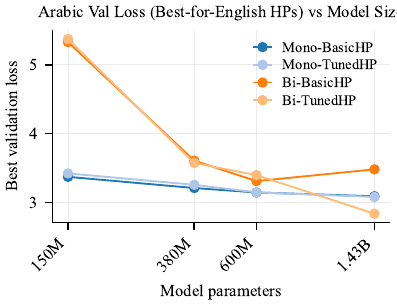}
    \caption{Arabic validation loss at the best-English checkpoint vs.\ model size. For monolingual runs, the best-English checkpoint corresponds to fewer repetitions than the best-Arabic one (e.g.\ $R = 10$ vs.\ $R = 27$ at 380M for Monolingual, basic), so Arabic loss is slightly worse. For bilingual runs, the best-English HPs select $R_\text{max} \leq 2$ at all scales below 1.43B ($\alpha < 0.09$), producing high Arabic loss at small scales. At 1.43B with tuned HPs, the gap narrows: best-English bilingual achieves Arabic loss of 2.84 vs.\ 2.68 when optimizing directly for Arabic, confirming that Arabic and English objectives converge at larger scale with proper tuning.}
    \label{fig:cross_arabic_val_scaling}
  \end{minipage}
\end{figure}

So far every result selects hyperparameters and checkpoints to minimize Arabic validation loss. We now flip the objective. Figure~\ref{fig:english_val_scaling} reports the best English validation loss achievable by each paradigm when HPs and checkpoints are selected to minimize English VL. Figure~\ref{fig:cross_arabic_val_scaling} reports the Arabic validation loss attained at those same English-optimal configurations, quantifying the cost in Arabic performance when a practitioner prioritizes the auxiliary language.

\subsection{Benchmark accuracy by paradigm}
\label{app:detailed_results}

\begin{figure}[!htbp]
  \centering
  \begin{minipage}[t]{0.48\linewidth}
    \centering
    \includegraphics[width=\linewidth]{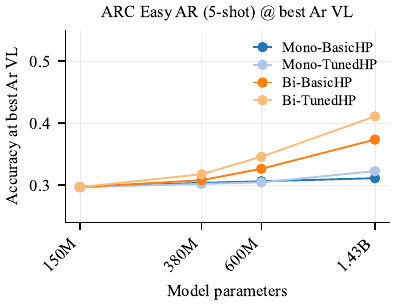}
  \end{minipage}
  \hfill
  \begin{minipage}[t]{0.48\linewidth}
    \centering
    \includegraphics[width=\linewidth]{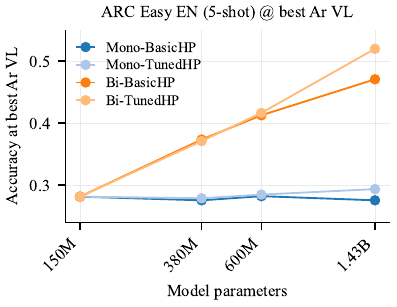}
  \end{minipage}
  \caption{ARC Easy accuracy (5-shot) vs.\ model size at the checkpoint with best Arabic validation loss (dashed lines in Fig.~\ref{fig:benchmarks}). \textbf{Left}: Arabic. \textbf{Right}: English. Blue: monolingual; orange: bilingual (dark = basic HP, light = tuned HP). Monolingual training is flat; bilingual scales steeply on both languages.}
  \label{fig:benchmark_by_arabic_vl}
\end{figure}

\begin{figure}[!htbp]
  \centering
  \begin{minipage}[t]{0.48\linewidth}
    \centering
    \includegraphics[width=\linewidth]{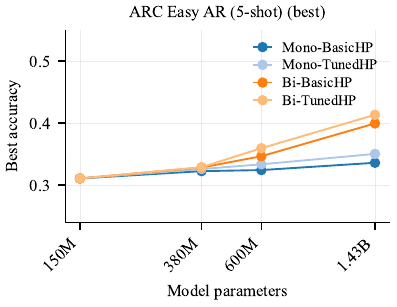}
  \end{minipage}
  \hfill
  \begin{minipage}[t]{0.48\linewidth}
    \centering
    \includegraphics[width=\linewidth]{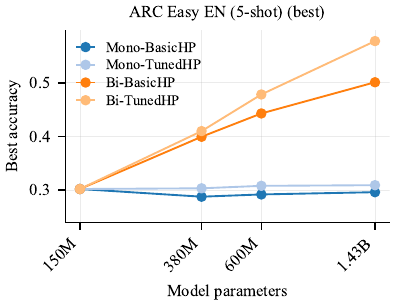}
  \end{minipage}
  \caption{ARC Easy accuracy (5-shot) vs.\ model size under oracle selection (best accuracy across all checkpoints; solid lines in Fig.~\ref{fig:benchmarks}). \textbf{Left}: Arabic. \textbf{Right}: English. HP tuning provides a larger benefit under oracle selection, especially on English at 1.43B (+7.7\%), but this requires optimizing for each language separately.}
  \label{fig:benchmark_by_accuracy}
\end{figure}

This subsection breaks down the summary in Fig.~\ref{fig:benchmarks} by paradigm, scale, language, and checkpoint-selection rule. Tables~\ref{tab:benchmark_ar_full} (Arabic) and~\ref{tab:benchmark_en_full} (English) report ARC Easy 5-shot accuracy under two rules: best accuracy across all checkpoints (left columns, oracle) and accuracy at the checkpoint minimizing Arabic VL (right columns, the cheap practitioner rule). Figures~\ref{fig:benchmark_by_arabic_vl} and~\ref{fig:benchmark_by_accuracy} plot the same numbers against model size. Three patterns recur across every panel: (i) at 150M all four paradigms collapse to the same accuracy, since the model consumes the full 200M Arabic budget without overfitting; (ii) monolingual accuracy is near-flat with scale on both languages while bilingual scales steeply; (iii) HP tuning adds a small, consistent margin on top of mixing, larger under oracle selection than under VL selection.

\begin{table}[h]
  \caption{ARC Easy Arabic (5-shot) accuracy per paradigm and model scale. \textbf{Left columns}: best accuracy across all checkpoints (selected by accuracy). \textbf{Right columns}: accuracy at the checkpoint with best Arabic validation loss (selected by VL).}
  \label{tab:benchmark_ar_full}
  \centering
  \begin{tabular}{lcccccccc}
    \toprule
    & \multicolumn{4}{c}{Selected by accuracy} & \multicolumn{4}{c}{Selected by Arabic VL} \\
    \cmidrule(lr){2-5} \cmidrule(lr){6-9}
    Paradigm & 150M & 380M & 600M & 1.43B & 150M & 380M & 600M & 1.43B \\
    \midrule
    Monolingual, basic HP & .311 & .323 & .325 & .336 & .297 & .304 & .306 & .311 \\
    Monolingual, tuned HP & .311 & .326 & .334 & .351 & .297 & .302 & .305 & .323 \\
    Bilingual, basic HP   & .311 & .329 & .347 & .400 & .297 & .308 & .327 & .374 \\
    Bilingual, tuned HP   & .311 & .329 & .360 & .414 & .297 & .318 & .346 & .411 \\
    \bottomrule
  \end{tabular}
\end{table}

\begin{table}[h]
  \caption{ARC Easy English (5-shot) accuracy per paradigm and model scale. Same layout as Table~\ref{tab:benchmark_ar_full}. Monolingual English accuracy is flat and near random ($\sim$28--30\%). Bilingual training scales steeply.}
  \label{tab:benchmark_en_full}
  \centering
  \begin{tabular}{lcccccccc}
    \toprule
    & \multicolumn{4}{c}{Selected by accuracy} & \multicolumn{4}{c}{Selected by Arabic VL} \\
    \cmidrule(lr){2-5} \cmidrule(lr){6-9}
    Paradigm & 150M & 380M & 600M & 1.43B & 150M & 380M & 600M & 1.43B \\
    \midrule
    Monolingual, basic HP & .302 & .288 & .292 & .296 & .282 & .276 & .282 & .276 \\
    Monolingual, tuned HP & .302 & .304 & .308 & .309 & .282 & .279 & .285 & .294 \\
    Bilingual, basic HP   & .302 & .399 & .443 & .501 & .282 & .374 & .413 & .471 \\
    Bilingual, tuned HP   & .302 & .410 & .478 & .577 & .282 & .371 & .417 & .520 \\
    \bottomrule
  \end{tabular}
\end{table}

\subsection{Validation loss as a per-run checkpoint selector}
\label{app:benchmarks}

\paragraph{Validation loss as a proxy.}
VL is a better model selection proxy for bilingual models than for monolingual ones. In monolingual training, the mean VL-accuracy gap is 2.1\% for Arabic and 1.8\% for English, with peak accuracy occurring after the best-VL checkpoint 84\% of the time. In bilingual training, the mean gap drops to 0.6\%, and peak accuracy occurs before or at the best-VL checkpoint 87\% of the time.

\begin{figure}[h]
  \centering
  \begin{minipage}[t]{0.48\linewidth}
    \centering
    \includegraphics[width=\linewidth]{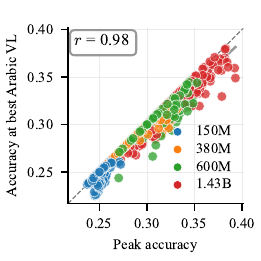}
  \end{minipage}
  \hfill
  \begin{minipage}[t]{0.48\linewidth}
    \centering
    \includegraphics[width=\linewidth]{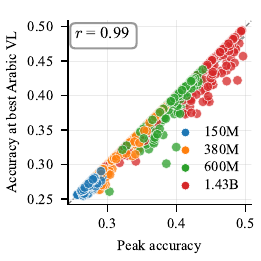}
  \end{minipage}
  \caption{Accuracy at the best Arabic VL checkpoint vs.\ peak accuracy (0-shot). \textbf{Left}: Arabic ARC Easy ($r = 0.98$). \textbf{Right}: English ARC Easy ($r = 0.99$). Same format as Fig.~\ref{fig:vl_proxy}.}
  \label{fig:vl_proxy_0shot}
\end{figure}

\paragraph{Validation loss proxy: full statistics.}
Table~\ref{tab:vl_proxy_stats} reports Pearson $r$, RMSE, and median absolute gap for all four evaluation settings, broken down by model scale. Figure~\ref{fig:vl_proxy_0shot} shows the 0-shot version of the scatter plots (5-shot is in the main text, Fig.~\ref{fig:vl_proxy}).

\begin{table}[h]
  \caption{Validation loss as a model selection proxy: Pearson $r$, RMSE (\%), and median $|$gap$|$ (\%) between accuracy at the best Arabic VL checkpoint and peak accuracy, per evaluation setting and model scale.}
  \label{tab:vl_proxy_stats}
  \centering
  \scriptsize
  \begin{tabular}{llcccc}
    \toprule
    Setting & Metric & 150M & 380M & 600M & 1.43B \\
    \midrule
    \multirow{2}{*}{AR 0-shot ($r = 0.98$)}
      & RMSE (\%)          & 0.84 & 0.64 & 1.04 & 1.13 \\
      & Median $|$gap$|$ (\%) & 0.19 & 0.00 & 0.04 & 0.30 \\
    \midrule
    \multirow{2}{*}{AR 5-shot ($r = 0.97$)}
      & RMSE (\%)          & 1.16 & 0.88 & 1.27 & 1.53 \\
      & Median $|$gap$|$ (\%) & 0.19 & 0.00 & 0.42 & 0.42 \\
    \midrule
    \multirow{2}{*}{EN 0-shot ($r = 0.99$)}
      & RMSE (\%)          & 0.60 & 0.77 & 1.50 & 1.84 \\
      & Median $|$gap$|$ (\%) & 0.13 & 0.00 & 0.13 & 0.38 \\
    \midrule
    \multirow{2}{*}{EN 5-shot ($r = 0.98$)}
      & RMSE (\%)          & 0.68 & 1.05 & 1.94 & 2.77 \\
      & Median $|$gap$|$ (\%) & 0.19 & 0.00 & 0.13 & 0.42 \\
    \bottomrule
  \end{tabular}
\end{table}

\subsection{Labeled-point configurations}

Tables~\ref{tab:points_ar_acc}--\ref{tab:points_en_vl} list the configuration ($\lambda$, $\eta$, $R_\text{max}$) and accuracy for each labeled point in Figures~\ref{fig:benchmark_by_arabic_vl} and~\ref{fig:benchmark_by_accuracy}. $\lambda$ and $\eta$ are base (pre-$\mu$P) values.

\begin{table}[h]
  \centering
  \begin{minipage}[t]{0.48\linewidth}
    \caption{Arabic ARC Easy points, selected by best accuracy (Fig.~\ref{fig:benchmark_by_accuracy}, left).}
    \label{tab:points_ar_acc}
    \centering
    \scriptsize
    \begin{tabular}{clccccc}
      \toprule
      ID & Paradigm & Model & $\lambda$ & $\eta$ & $R$ & Acc \\
      \midrule
      1  & Mono, basic & 150M  & 0.1  & 0.01  & 100 & 0.311 \\
      2  & Mono, basic & 380M  & 0.1  & 0.01  & 100 & 0.323 \\
      3  & Mono, basic & 600M  & 0.1  & 0.01  & 100 & 0.325 \\
      4  & Mono, basic & 1.43B & 0.1  & 0.01  & 100 & 0.336 \\
      \midrule
      5  & Mono, tuned & 150M  & 0.1  & 0.01  & 100 & 0.311 \\
      6  & Mono, tuned & 380M  & 0.01 & 0.003 & 100 & 0.326 \\
      7  & Mono, tuned & 600M  & 3   & 0.001 & 100 & 0.334 \\
      8  & Mono, tuned & 1.43B & 1   & 0.003 & 100 & 0.351 \\
      \midrule
      9  & Bi, basic & 150M  & 0.1  & 0.01  & 100 & 0.311 \\
      10 & Bi, basic & 380M  & 0.1  & 0.01  & 50  & 0.329 \\
      11 & Bi, basic & 600M  & 0.1  & 0.01  & 20  & 0.347 \\
      12 & Bi, basic & 1.43B & 0.1  & 0.01  & 100 & 0.400 \\
      \midrule
      13 & Bi, tuned & 150M  & 0.1  & 0.01  & 100 & 0.311 \\
      14 & Bi, tuned & 380M  & 1   & 0.003 & 50  & 0.329 \\
      15 & Bi, tuned & 600M  & 0.3  & 0.001 & 20  & 0.360 \\
      16 & Bi, tuned & 1.43B & 0.1  & 0.003 & 50  & 0.414 \\
      \bottomrule
    \end{tabular}
  \end{minipage}%
  \hfill
  \begin{minipage}[t]{0.48\linewidth}
    \caption{Arabic ARC Easy points, selected by best Arabic VL (Fig.~\ref{fig:benchmark_by_arabic_vl}, left).}
    \label{tab:points_ar_vl}
    \centering
    \scriptsize
    \begin{tabular}{clccccc}
      \toprule
      ID & Paradigm & Model & $\lambda$ & $\eta$ & $R$ & Acc \\
      \midrule
      1  & Mono, basic & 150M  & 0.1  & 0.01  & 100 & 0.297 \\
      2  & Mono, basic & 380M  & 0.1  & 0.01  & 100 & 0.304 \\
      3  & Mono, basic & 600M  & 0.1  & 0.01  & 100 & 0.306 \\
      4  & Mono, basic & 1.43B & 0.1  & 0.01  & 100 & 0.311 \\
      \midrule
      5  & Mono, tuned & 150M  & 0.1  & 0.01  & 100 & 0.297 \\
      6  & Mono, tuned & 380M  & 1   & 0.01  & 100 & 0.302 \\
      7  & Mono, tuned & 600M  & 1   & 0.01  & 100 & 0.305 \\
      8  & Mono, tuned & 1.43B & 0.3  & 0.01  & 100 & 0.323 \\
      \midrule
      9  & Bi, basic & 150M  & 0.1  & 0.01  & 100 & 0.297 \\
      10 & Bi, basic & 380M  & 0.1  & 0.01  & 20  & 0.308 \\
      11 & Bi, basic & 600M  & 0.1  & 0.01  & 20  & 0.327 \\
      12 & Bi, basic & 1.43B & 0.1  & 0.01  & 100 & 0.374 \\
      \midrule
      13 & Bi, tuned & 150M  & 0.1  & 0.01  & 100 & 0.297 \\
      14 & Bi, tuned & 380M  & 0.01 & 0.01  & 20  & 0.318 \\
      15 & Bi, tuned & 600M  & 0.1  & 0.003 & 20  & 0.346 \\
      16 & Bi, tuned & 1.43B & 0.1  & 0.003 & 50  & 0.411 \\
      \bottomrule
    \end{tabular}
  \end{minipage}
\end{table}

\begin{table}[h]
  \centering
  \begin{minipage}[t]{0.48\linewidth}
    \caption{English ARC Easy points, selected by best accuracy (Fig.~\ref{fig:benchmark_by_accuracy}, right).}
    \label{tab:points_en_acc}
    \centering
    \scriptsize
    \begin{tabular}{clccccc}
      \toprule
      ID & Paradigm & Model & $\lambda$ & $\eta$ & $R$ & Acc \\
      \midrule
      1  & Mono, basic & 150M  & 0.1  & 0.01  & 100 & 0.302 \\
      2  & Mono, basic & 380M  & 0.1  & 0.01  & 100 & 0.288 \\
      3  & Mono, basic & 600M  & 0.1  & 0.01  & 100 & 0.292 \\
      4  & Mono, basic & 1.43B & 0.1  & 0.01  & 100 & 0.296 \\
      \midrule
      5  & Mono, tuned & 150M  & 0.1  & 0.01  & 100 & 0.302 \\
      6  & Mono, tuned & 380M  & 1   & 0.003 & 100 & 0.304 \\
      7  & Mono, tuned & 600M  & 1   & 0.003 & 100 & 0.308 \\
      8  & Mono, tuned & 1.43B & 3   & 0.001 & 100 & 0.309 \\
      \midrule
      9  & Bi, basic & 150M  & 0.1  & 0.01  & 100 & 0.302 \\
      10 & Bi, basic & 380M  & 0.1  & 0.01  & 1   & 0.399 \\
      11 & Bi, basic & 600M  & 0.1  & 0.01  & 10  & 0.443 \\
      12 & Bi, basic & 1.43B & 0.1  & 0.01  & 50  & 0.501 \\
      \midrule
      13 & Bi, tuned & 150M  & 0.1  & 0.01  & 100 & 0.302 \\
      14 & Bi, tuned & 380M  & 0.01 & 0.003 & 1   & 0.410 \\
      15 & Bi, tuned & 600M  & 0.01 & 0.01  & 5   & 0.478 \\
      16 & Bi, tuned & 1.43B & 0.01 & 0.001 & 1   & 0.577 \\
      \bottomrule
    \end{tabular}
  \end{minipage}%
  \hfill
  \begin{minipage}[t]{0.48\linewidth}
    \caption{English ARC Easy points, selected by best Arabic VL (Fig.~\ref{fig:benchmark_by_arabic_vl}, right). Same checkpoints as Table~\ref{tab:points_ar_vl}.}
    \label{tab:points_en_vl}
    \centering
    \scriptsize
    \begin{tabular}{clccccc}
      \toprule
      ID & Paradigm & Model & $\lambda$ & $\eta$ & $R$ & Acc \\
      \midrule
      1  & Mono, basic & 150M  & 0.1  & 0.01  & 100 & 0.282 \\
      2  & Mono, basic & 380M  & 0.1  & 0.01  & 100 & 0.276 \\
      3  & Mono, basic & 600M  & 0.1  & 0.01  & 100 & 0.282 \\
      4  & Mono, basic & 1.43B & 0.1  & 0.01  & 100 & 0.276 \\
      \midrule
      5  & Mono, tuned & 150M  & 0.1  & 0.01  & 100 & 0.282 \\
      6  & Mono, tuned & 380M  & 1   & 0.01  & 100 & 0.279 \\
      7  & Mono, tuned & 600M  & 1   & 0.01  & 100 & 0.285 \\
      8  & Mono, tuned & 1.43B & 0.3  & 0.01  & 100 & 0.294 \\
      \midrule
      9  & Bi, basic & 150M  & 0.1  & 0.01  & 100 & 0.282 \\
      10 & Bi, basic & 380M  & 0.1  & 0.01  & 20  & 0.374 \\
      11 & Bi, basic & 600M  & 0.1  & 0.01  & 20  & 0.413 \\
      12 & Bi, basic & 1.43B & 0.1  & 0.01  & 100 & 0.471 \\
      \midrule
      13 & Bi, tuned & 150M  & 0.1  & 0.01  & 100 & 0.282 \\
      14 & Bi, tuned & 380M  & 0.01 & 0.01  & 20  & 0.371 \\
      15 & Bi, tuned & 600M  & 0.1  & 0.003 & 20  & 0.417 \\
      16 & Bi, tuned & 1.43B & 0.1  & 0.003 & 50  & 0.520 \\
      \bottomrule
    \end{tabular}
  \end{minipage}
\end{table}

\section{Data multiplier and equivalence}
\label{app:data_equiv}

\subsection{Monolingual training across Arabic corpus sizes}
\label{app:monolingual_sweep}

We train monolingual models at seven Arabic corpus sizes, $D_\text{LR} \in \{25\text{M}, 50\text{M}, 100\text{M}, 200\text{M}, 500\text{M}, 1\text{B}, 2\text{B}\}$ tokens, each repeated up to 100 times, with $\mu$P HPs. We report Arabic validation loss (at 150M, 380M, 600M, 1.43B) and ARC Easy 5-shot accuracy (at 380M, 600M, 1.43B).

\begin{figure}[h]
  \centering
  \begin{minipage}[t]{0.48\linewidth}
    \centering
    \includegraphics[width=\linewidth]{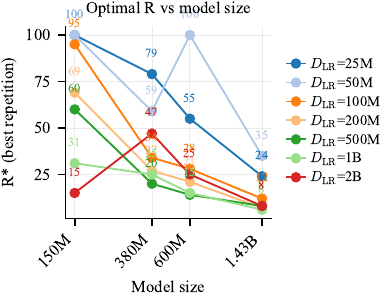}
    \caption{Optimal repetition count $R^*$ vs.\ Arabic corpus size $D_\text{LR}$ for each model scale.}
    \label{fig:best_R}
  \end{minipage}
  \hfill
  \begin{minipage}[t]{0.48\linewidth}
    \centering
    \includegraphics[width=\linewidth]{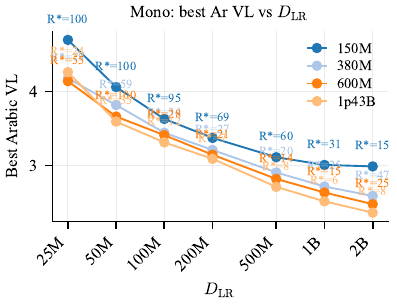}
    \caption{Best Arabic validation loss vs.\ Arabic corpus size $D_\text{LR}$ for each model scale (monolingual, basic HP).}
    \label{fig:loss_vs_dlr}
  \end{minipage}
\end{figure}

\paragraph{Validation loss.}
Table~\ref{tab:data_equiv} reports the best Arabic VL at each ($D_\text{LR}$, scale) pair alongside the optimal repetition count $R^*$ at which that minimum is reached. Figure~\ref{fig:loss_vs_dlr} plots the same loss against $D_\text{LR}$. VL decreases steadily with more unique data at every scale, with gains diminishing in the high-$D_\text{LR}$ regime.

\begin{table}[h]
  \centering
  \caption{Best Arabic validation loss for monolingual training with $\mu$P HPs at each corpus size. $R^*$ is the optimal repetition count.}
  \label{tab:data_equiv}
  \scriptsize
  \begin{tabular}{lcccccccc}
    \toprule
    & \multicolumn{2}{c}{150M} & \multicolumn{2}{c}{380M} & \multicolumn{2}{c}{600M} & \multicolumn{2}{c}{1.43B} \\
    \cmidrule(lr){2-3} \cmidrule(lr){4-5} \cmidrule(lr){6-7} \cmidrule(lr){8-9}
    $D_\text{LR}$ & Loss & $R^*$ & Loss & $R^*$ & Loss & $R^*$ & Loss & $R^*$ \\
    \midrule
    25M   & 4.698 & 100 & 4.177 & 79  & 4.140 & 55  & 4.262 & 24 \\
    50M   & 4.060 & 100 & 3.817 & 59  & 3.661 & 100 & 3.593 & 35 \\
    100M  & 3.629 & 95  & 3.442 & 34  & 3.409 & 28  & 3.311 & 12 \\
    200M  & 3.372 & 69  & 3.209 & 27  & 3.144 & 21  & 3.089 & 7 \\
    500M  & 3.111 & 60  & 2.905 & 20  & 2.817 & 14  & 2.709 & 8 \\
    1B    & 3.007 & 31  & 2.715 & 25  & 2.633 & 15  & 2.514 & 6 \\
    2B    & 2.987 & 15  & 2.587 & 47  & 2.476 & 25  & 2.361 & 8 \\
    \bottomrule
  \end{tabular}
\end{table}

\paragraph{Optimal repetition count.}
Figure~\ref{fig:best_R} visualizes how $R^*$ shrinks as the corpus grows: the 1.43B model peaks at $R^* = 24$ with 25M tokens but at $R^* = 8$ with 2B tokens. More unique data means less repetition is needed before overfitting dominates.

\paragraph{Downstream accuracy.}
Table~\ref{tab:dlr_benchmarks} reports peak ARC Easy 5-shot accuracy for the same sweep, split by language. Arabic accuracy rises from $27$--$30\%$ at $D_\text{LR} = 25\text{M}$ to $35$--$41\%$ at $D_\text{LR} = 2\text{B}$. English accuracy also rises modestly with corpus size even though no dedicated English data is seen. These monolingual numbers are the reference against which bilingual accuracy at $D_\text{LR} = 200\text{M}$ is compared in the multiplier analysis: how much monolingual Arabic data is equivalent to the 200M-token bilingual run?

\begin{table}[h]
  \caption{ARC Easy accuracy (5-shot) for monolingual training at each Arabic corpus size and model scale.}
  \label{tab:dlr_benchmarks}
  \centering
  \begin{tabular}{lcccccc}
    \toprule
    & \multicolumn{3}{c}{Arabic} & \multicolumn{3}{c}{English} \\
    \cmidrule(lr){2-4} \cmidrule(lr){5-7}
    $D_\text{LR}$ & 380M & 600M & 1.43B & 380M & 600M & 1.43B \\
    \midrule
    25M   & 0.272 & 0.297 & 0.291 & 0.284 & 0.285 & 0.285 \\
    50M   & 0.301 & 0.297 & 0.306 & 0.282 & 0.290 & 0.295 \\
    100M  & 0.289 & 0.312 & 0.318 & 0.289 & 0.297 & 0.296 \\
    200M  & 0.323 & 0.325 & 0.336 & 0.288 & 0.292 & 0.296 \\
    500M  & 0.339 & 0.351 & 0.355 & 0.312 & 0.307 & 0.319 \\
    1B    & 0.350 & 0.351 & 0.384 & 0.322 & 0.329 & 0.350 \\
    2B    & 0.348 & 0.368 & 0.410 & 0.340 & 0.344 & 0.363 \\
    \bottomrule
  \end{tabular}
\end{table}

\subsection{Variants: 0-shot and $\mu$P peak-accuracy plots}
\label{app:multiplier_variants}

Figure~\ref{fig:flip} in the main body plots peak ARC Easy 5-shot accuracy against $D_\text{LR}$, with per-scale monolingual curves and bilingual-tuned-HP stars at $D_\text{LR} = 200\text{M}$ sitting above those curves. That layout is the visual hint behind the multiplier: the horizontal gap between a star and the corresponding monolingual curve is $D_\text{LR}^\text{equiv}$, and the multiplier simply reports it as a ratio to $200\text{M}$. The three panels below repeat the same layout under two natural variants, 0-shot evaluation and $\mu$P (rather than per-scale-tuned) HPs for the bilingual stars.

\begin{figure}[h]
  \centering
  \begin{minipage}[t]{0.48\linewidth}
    \centering
    \includegraphics[width=\linewidth]{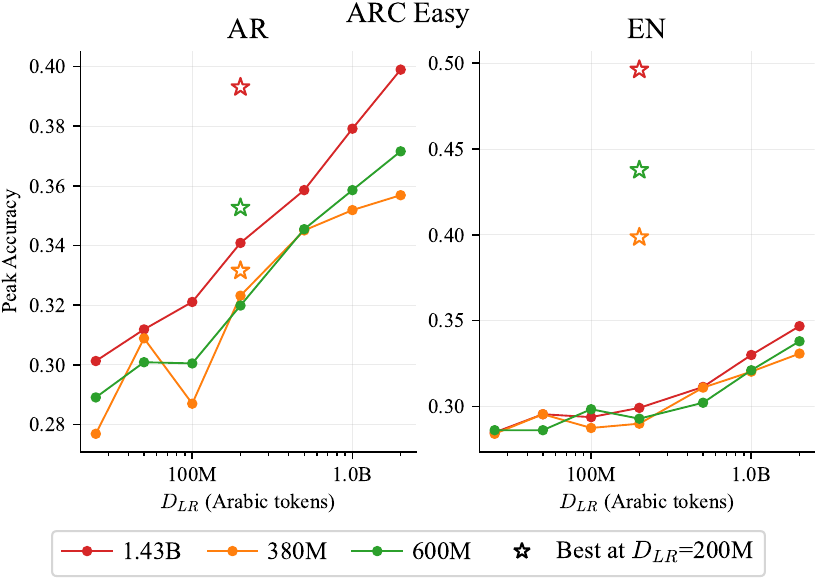}
    \caption{Peak ARC Easy accuracy (0-shot) vs.\ Arabic data budget ($D_\text{LR}$). Same layout as Fig.~\ref{fig:flip} but with 0-shot evaluation; stars use per-scale tuned HPs. The Arabic-side pattern mirrors 5-shot: bilingual approaches but does not exceed monolingual at $D_\text{LR} = 2\text{B}$; English-side bilingual dominance persists.}
    \label{fig:flip_0shot}
  \end{minipage}
  \hfill
  \begin{minipage}[t]{0.48\linewidth}
    \centering
    \includegraphics[width=\linewidth]{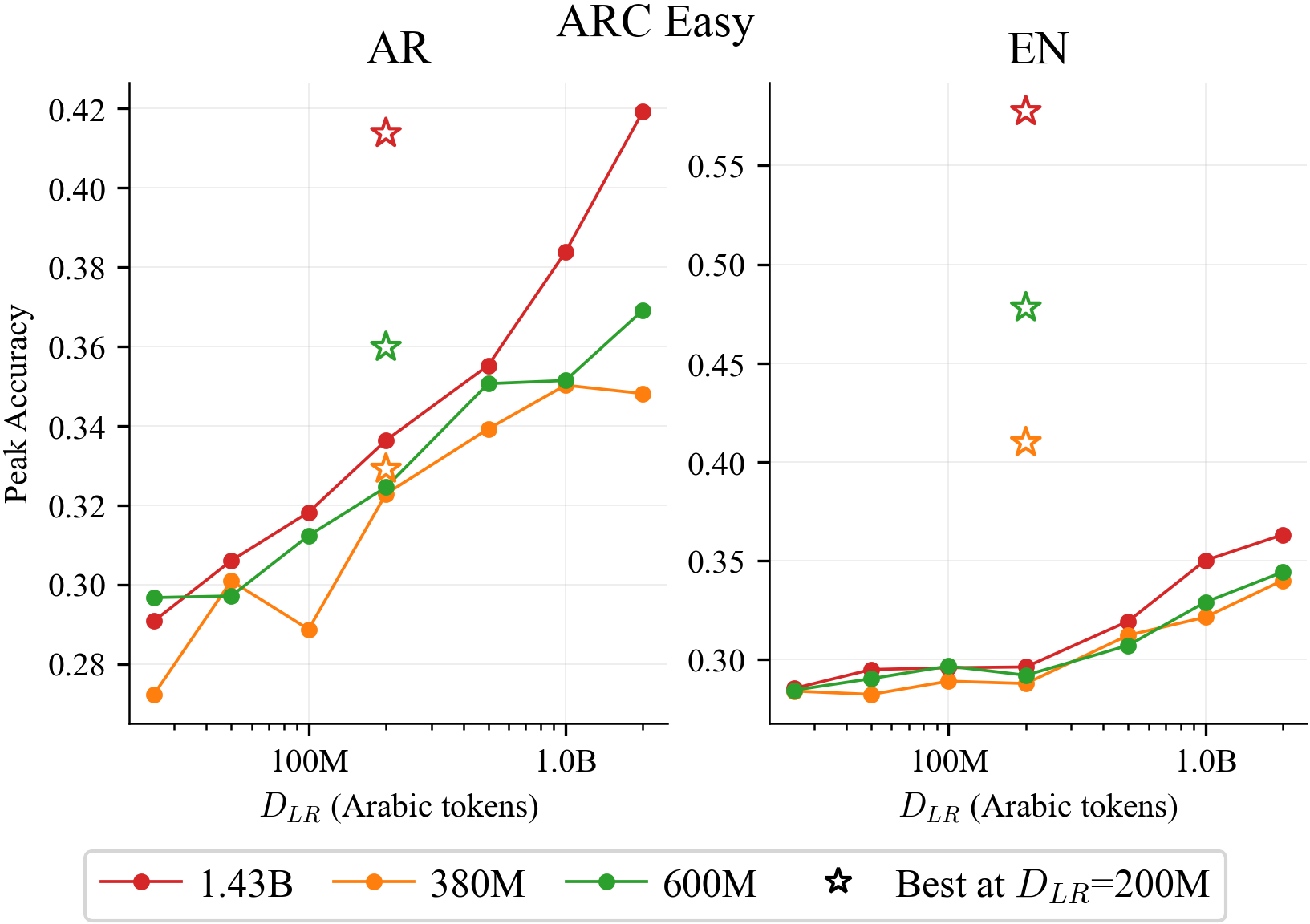}
    \caption{Peak ARC Easy accuracy (5-shot) vs.\ Arabic data budget ($D_\text{LR}$) with $\mu$P HPs for the bilingual runs. Same layout and monolingual lines as Fig.~\ref{fig:flip}; only the bilingual stars change. Compared with tuned HPs (Fig.~\ref{fig:flip}), $\mu$P stars sit slightly below on Arabic and on English at 1.43B, confirming that per-scale tuning gains a few accuracy points on top of $\mu$P transfer.}
    \label{fig:flip_mup_5shot}
  \end{minipage}
\end{figure}

\begin{figure}[h]
  \centering
  \begin{minipage}[c]{0.6\linewidth}
    \centering
    \includegraphics[width=\linewidth]{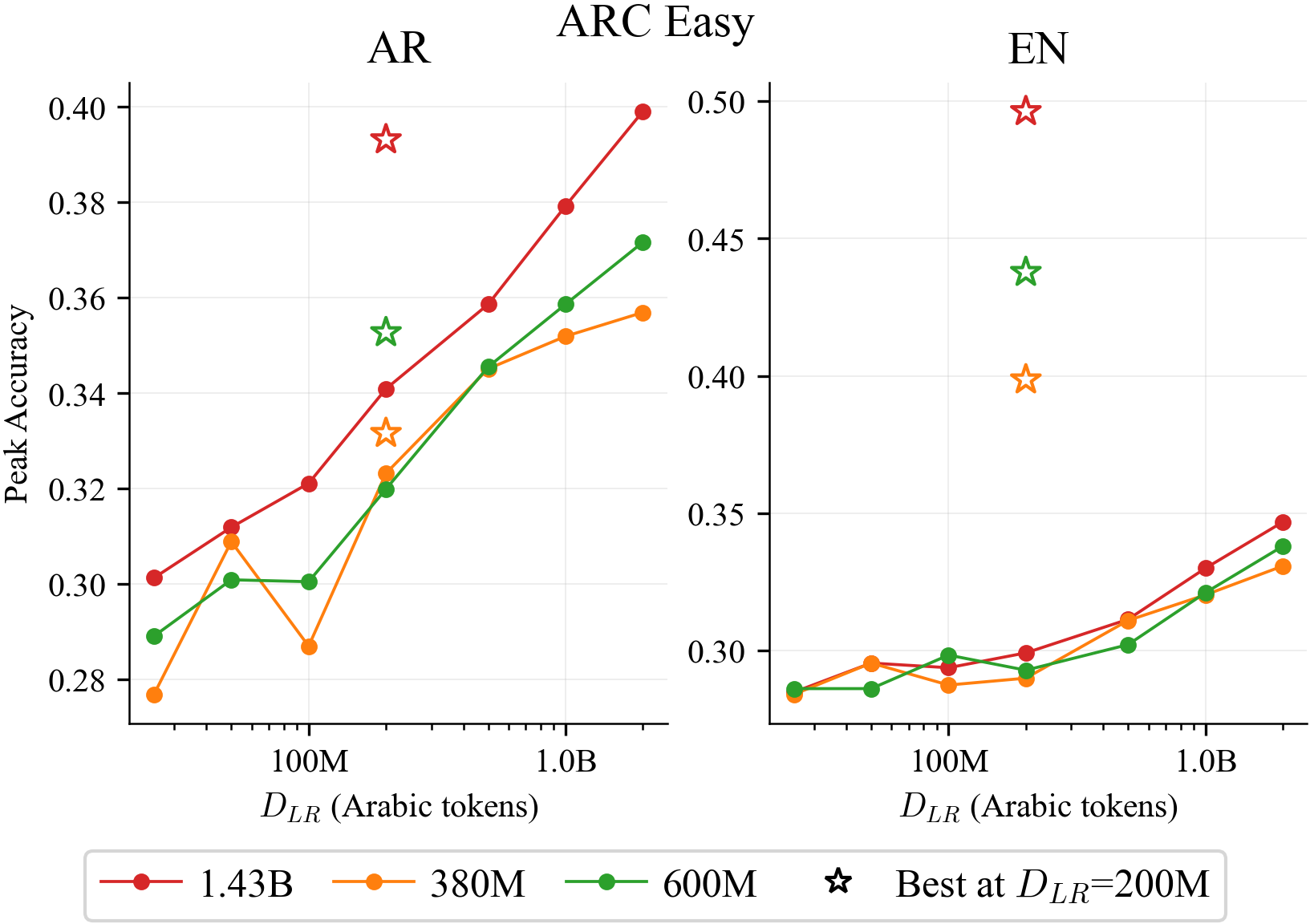}
  \end{minipage}%
  \hfill
  \begin{minipage}[c]{0.38\linewidth}
    \caption{Peak ARC Easy accuracy (0-shot) vs.\ Arabic data budget ($D_\text{LR}$) with $\mu$P HPs for the bilingual runs. Same layout as Fig.~\ref{fig:flip_mup_5shot} but with 0-shot evaluation.}
    \label{fig:flip_mup_0shot}
  \end{minipage}
\end{figure}

\subsection{Data multiplier construction and results}
\label{app:multiplier_construction}

\paragraph{Multiplier construction.}
Recall from \S\ref{sec:data_equiv} that the multiplier is $N = D_\text{LR}^\text{equiv}/200\text{M}$: the unique Arabic corpus size a monolingual model would need to match a bilingual model trained on only 200M Arabic tokens. To compute it on a given metric (Arabic validation loss or ARC Easy accuracy), we need two ingredients: (i) a reference curve mapping $D_\text{LR}$ to best monolingual performance, and (ii) a bilingual target value to invert against. The reference curve is built from the monolingual sweep in \S\ref{app:monolingual_sweep}: for each $D_\text{LR} \in \{25\text{M}, 50\text{M}, 100\text{M}, 200\text{M}, 500\text{M}, 1\text{B}, 2\text{B}\}$ we take the best performance achieved over repetitions under $\mu$P HPs, then fit a log-linear model $y = a \ln(D_\text{LR}) + b$ to the seven points. Fits are tight at every scale ($r^2 \geq 0.90$; coefficients in Table~\ref{tab:fit_coefficients}). The bilingual target is the best value achieved at $D_\text{LR} = 200\text{M}$ under either $\mu$P HPs (basic) or a full HP sweep (tuned). Inverting the fit at the target value yields $D_\text{LR}^\text{equiv}$, and $N$ follows by division.

\begin{figure}[h]
  \centering
  \includegraphics[width=\linewidth]{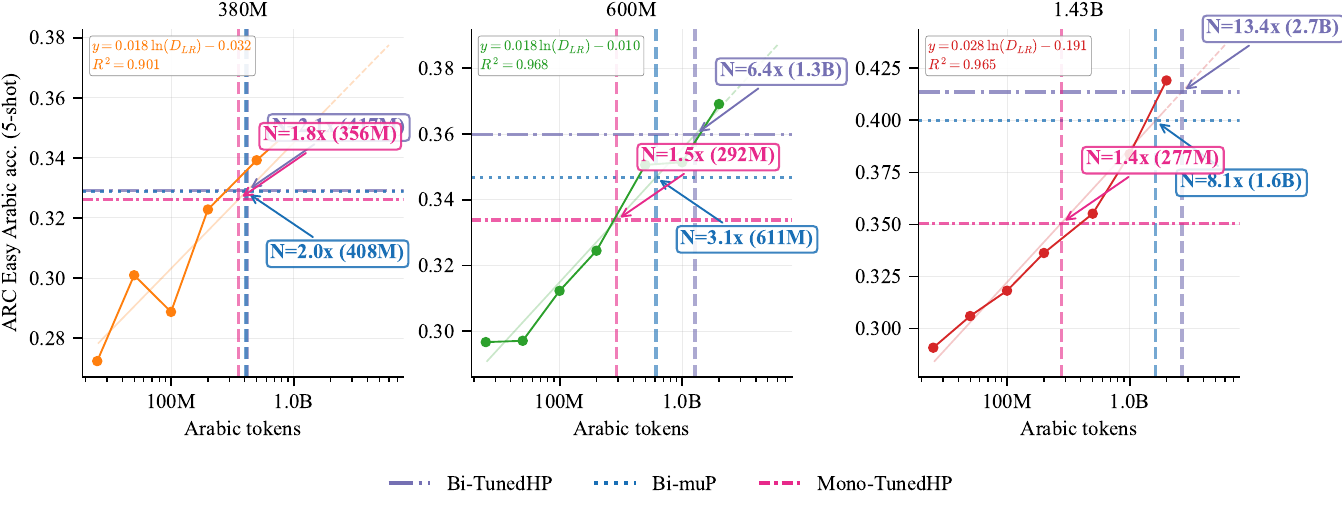}
  \caption{Data multiplier methodology (ARC Easy accuracy). Solid curves: monolingual accuracy vs.\ $D_\text{LR}$, with log-linear fit. Horizontal lines: bilingual performance at $D_\text{LR} = 200\text{M}$ under $\mu$P HPs (blue) and full HP sweep (purple). Vertical dashed lines: equivalent $D_\text{LR}^\text{equiv}$. The gap between the two targets grows with scale.}
  \label{fig:multiplier_method}
\end{figure}

\begin{figure}[h]
  \centering
  \includegraphics[width=\linewidth]{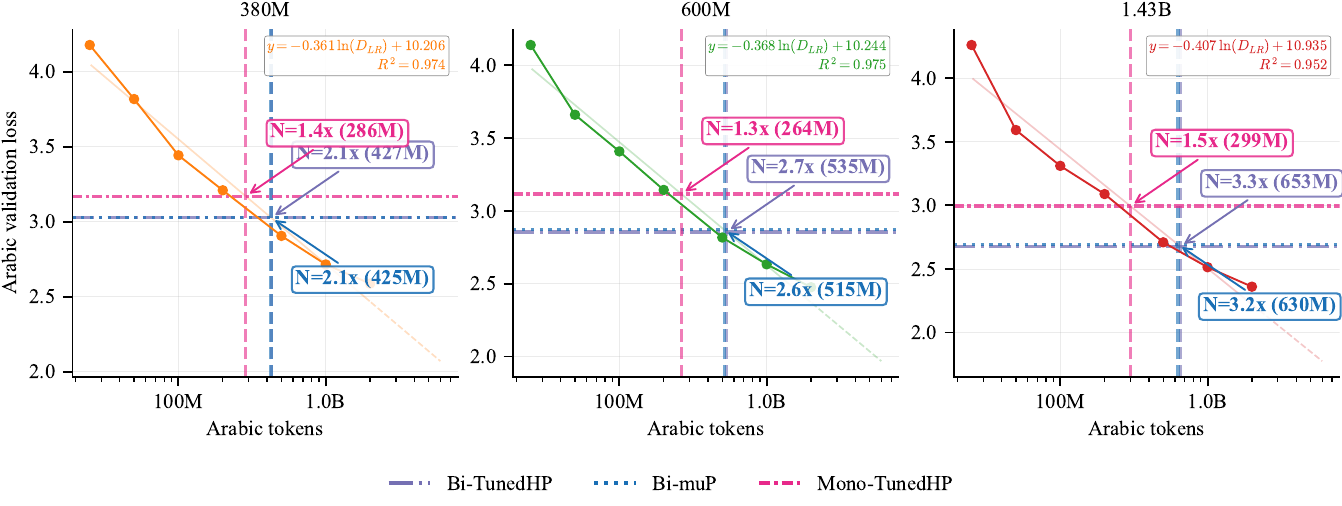}
  \caption{Data multiplier methodology (Arabic validation loss). Same format as Fig.~\ref{fig:multiplier_method}. The VL multiplier stays in a narrow $2$--$3\times$ band across scales, substantially smaller than the benchmark multiplier.}
  \label{fig:multiplier_method_vl}
\end{figure}

\paragraph{Reading the methodology figures.}
Each of Figures~\ref{fig:multiplier_method} and~\ref{fig:multiplier_method_vl} has one panel per model scale (380M, 600M, 1.43B). Within a panel: the x-axis is Arabic corpus size $D_\text{LR}$ (log scale), the y-axis is performance (accuracy in Fig.~\ref{fig:multiplier_method}, Arabic VL in Fig.~\ref{fig:multiplier_method_vl}), and the solid curve through the markers is the log-linear fit to the monolingual sweep. Two horizontal lines mark the bilingual targets achieved at $D_\text{LR} = 200\text{M}$ under $\mu$P HPs (blue) and the full HP sweep (purple). To recover a multiplier from one of those lines, follow it horizontally until it meets the monolingual curve, then drop down to read $D_\text{LR}^\text{equiv}$ on the x-axis (marked by a vertical dashed line); $N = D_\text{LR}^\text{equiv}/200\text{M}$. The further the intersection sits to the right of $D_\text{LR} = 200\text{M}$, the larger the multiplier.

\paragraph{Results.}
On accuracy (Fig.~\ref{fig:multiplier_method}), the horizontal gap between the bilingual targets and the monolingual reference grows steeply with scale, yielding $N$ from $2\times$ at 380M to $13\times$ at 1.43B. On Arabic VL (Fig.~\ref{fig:multiplier_method_vl}), the gap stays narrow, yielding $N$ in a $2$--$3\times$ band across all scales. Table~\ref{tab:multiplier} summarises the accuracy multipliers under both controlled (identical HPs) and optimized (full HP sweep for bilingual) conditions. These two views are the basis of the divergence reported in \S\ref{sec:data_equiv}.

\begin{table}[h]
\centering
\begin{minipage}[t]{0.48\linewidth}
  \centering
  \captionof{table}{Log-linear fit coefficients for the data multiplier. Model: $y = a \cdot \ln(D_\text{LR}) + b$, where $y$ is either ARC Easy accuracy or Arabic validation loss.}
  \label{tab:fit_coefficients}
  \small
  \begin{tabular}{llccc}
    \toprule
    Model & Metric & $a$ & $b$ & $r^2$ \\
    \midrule
    380M  & Accuracy & $+0.018$ & $-0.032$ & 0.90 \\
    380M  & Arabic VL & $-0.361$ & $+10.21$ & 0.97 \\
    \midrule
    600M  & Accuracy & $+0.017$ & $-0.006$ & 0.97 \\
    600M  & Arabic VL & $-0.368$ & $+10.24$ & 0.98 \\
    \midrule
    1.43B & Accuracy & $+0.027$ & $-0.168$ & 0.98 \\
    1.43B & Arabic VL & $-0.407$ & $+10.94$ & 0.95 \\
    \bottomrule
  \end{tabular}
\end{minipage}
\hfill
\begin{minipage}[t]{0.48\linewidth}
  \centering
  \captionof{table}{Data multiplier $N$: unique Arabic data (as a multiple of $D_\text{LR} = 200\text{M}$) a monolingual model would need to match bilingual accuracy on ARC Easy (5-shot). Controlled: identical HPs. Optimized: bilingual gets a full HP sweep.}
  \label{tab:multiplier}
  \small
  \begin{tabular}{lcccc}
    \toprule
    & \multicolumn{2}{c}{Controlled} & \multicolumn{2}{c}{Optimized} \\
    \cmidrule(lr){2-3} \cmidrule(lr){4-5}
    Model & Acc & $N$ & Acc & $N$ \\
    \midrule
    380M  & 0.329 & 2.0$\times$ & 0.329 & 2.1$\times$ \\
    600M  & 0.347 & 3.1$\times$ & 0.360 & 6.4$\times$ \\
    1.43B & 0.400 & 8.1$\times$ & 0.414 & 13.4$\times$ \\
    \bottomrule
  \end{tabular}
\end{minipage}
\end{table}

\begin{SCfigure}[][h]
  \centering
  \includegraphics[width=0.55\linewidth]{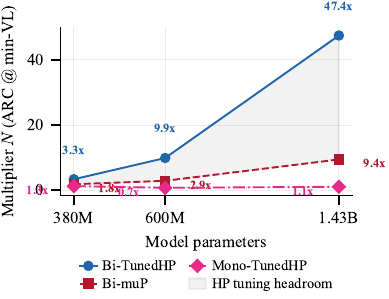}
  \caption{Data multiplier $N$ vs.\ model size, with checkpoints selected by min Arabic VL rather than peak accuracy. Same paradigms as Fig.~\ref{fig:multiplier_comparison}, right panel. The bilingual-tuned-HP multiplier inflates from $13.4\times$ to $47.4\times$ at 1.43B; the 1.43B value is an extrapolation of the log-linear fit $4.7\times$ past the largest monolingual corpus ($D_\text{LR} = 2\text{B}$).}
  \label{fig:multiplier_at_vl}
\end{SCfigure}

\paragraph{Multiplier under validation-loss-based checkpoint selection.}
\label{app:multiplier_vl}
The main-text multiplier (Fig.~\ref{fig:multiplier_comparison}) selects checkpoints by peak accuracy on the target benchmark, which requires running the benchmark at every checkpoint. A cheaper rule picks the min-VL checkpoint and reports accuracy there. Figure~\ref{fig:multiplier_at_vl} recomputes the multiplier under this rule.

\begin{figure}[h]
  \centering
  \includegraphics[width=\linewidth]{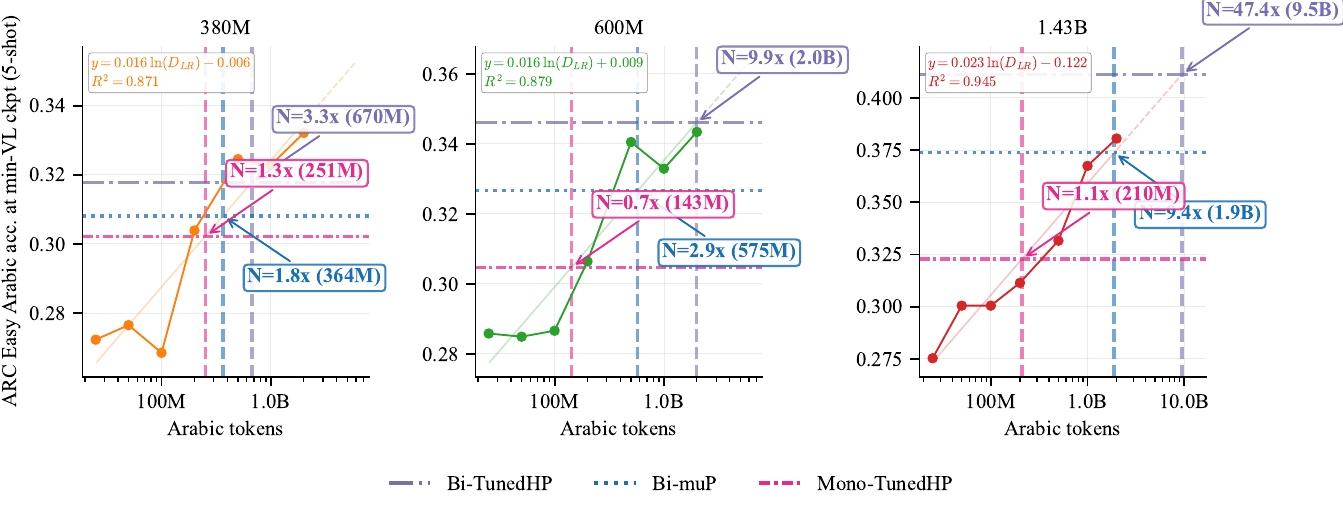}
  \caption{Data multiplier methodology under min-VL checkpoint selection (Arabic ARC Easy 5-shot). For each paradigm, accuracy is read at the checkpoint that minimizes Arabic VL rather than at peak accuracy. The monolingual reference curve shifts down relative to Fig.~\ref{fig:multiplier_method}, while the bilingual targets barely move, producing the inflated multipliers in Fig.~\ref{fig:multiplier_at_vl}.}
  \label{fig:multiplier_method_at_vl}
\end{figure}

\subparagraph{Observation.}
Under VL-based selection, the Bilingual-tuned-HP multiplier rises from $13.4\times$ to $47.4\times$ at 1.43B. The rise is driven by an asymmetry in how much accuracy each paradigm loses when forced to commit to min-VL. At 1.43B, the best-VL bilingual run loses $0.003$ accuracy relative to its own peak-accuracy checkpoint ($0.4137 \to 0.4112$); the monolingual run at $D_\text{LR} = 2\text{B}$ loses $0.039$ ($0.4192 \to 0.3805$), an order of magnitude more. The same pattern holds across the monolingual sweep: accuracy at min-VL is $0.015$--$0.040$ below peak at every $D_\text{LR}$ and every scale, while bilingual gaps stay below $0.005$.

\subparagraph{Effect on the fit.}
Switching from peak-accuracy to min-VL checkpoint selection shifts the two sides of the multiplier comparison asymmetrically. The bilingual target barely drops (best-VL is within $0.003$ of best-accuracy for bilingual runs), while the monolingual reference curve drops by roughly $0.025$ at every $D_\text{LR}$ (monolingual runs lose much more accuracy when forced onto their min-VL checkpoint). With the target held roughly fixed and the reference curve pushed down, monolingual now needs far more data to reach the target, inflating $N$ from $13.4\times$ to $47.4\times$ at 1.43B. At 1.43B, even $D_\text{LR} = 2\text{B}$ on the shifted curve falls short of the target, so $D_\text{LR}^\text{equiv}$ is read from the log-linear fit extrapolated $4.7\times$ past the fit range.

\subparagraph{Caveats.}
The $4.7\times$ extrapolation at 1.43B assumes the log-linear fit's slope holds past $D_\text{LR} = 2\text{B}$; we treat $13.4\times$ (from peak-accuracy selection, Fig.~\ref{fig:multiplier_comparison}) as the empirical lower bound and the VL-based numbers as upper-bound estimates contingent on fit extrapolation.

\section{ANOVA methodology}
\label{app:anova_method}

Throughout this paper, we use analysis of variance (ANOVA) to measure how much of the total spread in Arabic validation loss is attributable to each tuning axis. Variance-based HP sensitivity analysis has substantial precedent in the HPO literature \citep{pmlr-v32-hutter14, probst2018tunabilityimportancehyperparametersmachine, watanabe2023pedanovaefficientlyquantifyinghyperparameter}; we use classical ANOVA on a structured grid of trained models. Our bilingual grid measures Arabic validation loss at every combination of $R_\text{max} \in \{1, 2, 5, 10, 20, 50, 100\}$ and the 25 hyperparameter cells ($5\,\lambda \times 5\,\eta$). The headline decomposition in \S\ref{sec:hp_sensitivity} is two-way: $R_\text{max}$ against a single 25-level HP factor. The three-way decomposition in App.~\ref{app:hp_landscape} splits the HP factor into $\lambda$ and $\eta$ with their interaction. We describe the two-way version below; the three-way extension is standard.

\paragraph{Setup.}
Suppose we have a grid of outcomes $y_{rh}$ indexed by two factors: a row factor $r \in \{1, \ldots, R\}$ (e.g.\ $R_\text{max}$ levels) and a column factor $h \in \{1, \ldots, H\}$ (e.g.\ $(\lambda, \eta)$ combinations). Each cell $(r, h)$ contains one observation: the best Arabic validation loss for that configuration. The grand mean is $\bar{y} = \frac{1}{N}\sum_{r,h} y_{rh}$, where $N$ is the number of observed cells.

\paragraph{Decomposition.}
ANOVA asks: when we look at all the variation in this grid, how much comes from changing the row factor vs.\ changing the column factor? It decomposes the total sum of squared deviations from the grand mean into additive terms:
\begin{align}
  \text{SS}_\text{total} &= \text{SS}_\text{row} + \text{SS}_\text{col} + \text{SS}_\text{residual}
\end{align}

Each term measures how much variation one factor explains:
\begin{itemize}
  \item $\text{SS}_\text{row} = \sum_r n_r \,(\bar{y}_{r\cdot} - \bar{y})^2$, where $\bar{y}_{r\cdot}$ is the mean of row $r$ and $n_r$ is the number of observed cells in that row. This captures how much the outcome changes when we vary the row factor (e.g.\ $R_\text{max}$), averaging over all column settings.
  \item $\text{SS}_\text{col} = \sum_h n_h \,(\bar{y}_{\cdot h} - \bar{y})^2$, defined analogously for columns. This captures the effect of varying HPs, averaging over all data settings.
  \item $\text{SS}_\text{residual} = \text{SS}_\text{total} - \text{SS}_\text{row} - \text{SS}_\text{col}$, the leftover. In a design with one observation per cell (as ours), this confounds the true interaction between factors with observation noise.
\end{itemize}

We report each term as a percentage of $\text{SS}_\text{total}$. If a factor explains 80\% of the total variance, then choosing the wrong level of that factor is far more costly than choosing the wrong level of a factor that explains 16\%.

\paragraph{Interpretation and caveats.}
A high $\text{SS}_\text{row}$ fraction means the row means are spread far apart relative to the overall spread, i.e., the outcome is sensitive to the row factor. Conversely, a low fraction means rows have similar means, so the outcome is insensitive to that factor regardless of the column setting.

Two caveats apply. First, because each cell contains a single training run (no replication), the residual term cannot separate true factor interaction from noise. In practice the two-way residual is 4--6\% across 380M, 600M, and 1.43B, and the three-way residual stays below 7\% (App.~\ref{app:hp_landscape}), so this limitation does not affect our conclusions. Second, the decomposition depends on which factor levels are included in the grid. We address this by reporting three operational views around the per-scale optimum and the $\mu$P HP (\S\ref{sec:hp_sensitivity} and App.~\ref{app:operational_sensitivity}), which makes the conclusions robust to grid boundary choices.

\section{Sensitivity around the operational point: the choice of grid matters}
\label{app:operational_sensitivity}

How sensitive is bilingual training performance to HP choice around the operational point? The answer depends on how we define ``around'', and the choice matters more than one might expect. We report three operational views of the same data and show that they tell complementary stories, all converging on the same practical recommendation. Views 2 and 3 diverge sharply at 1.43B between an optimum-centered and a $\mu$P-centered analysis; App.~\ref{app:hp_landscape} investigates that divergence and the full-grid pattern that produces it.

\paragraph{Three views of operational sensitivity.}
We consider three ways to restrict the ANOVA to an operational region:
\begin{enumerate}
  \item \textbf{Threshold-based (Lebesgue split).} Keep the top-$\gamma\%$ of grid cells by performance. This is the construction used by PED-ANOVA \citep{watanabe2023pedanovaefficientlyquantifyinghyperparameter} and corresponds to a \emph{post-hoc} sensitivity analysis: it answers ``among the configurations that turned out to be good, which factor mattered most?'' The kept cells can be scattered anywhere in HP space.
  \item \textbf{Contiguous neighborhood around the per-scale optimum (Riemann split).} Pick the per-scale optimal HP and expand a contiguous sub-grid radially outward. This answers ``in a neighborhood of the actual best HP, which factor matters most?'' It is an idealized lens, since identifying the per-scale optimum requires the full grid we are trying to analyze.
  \item \textbf{Contiguous neighborhood around $\mu$P (Riemann split).} Same as (2) but anchored at the $\mu$P HP from a small proxy. This is the \emph{a-priori} practitioner view: a real practitioner has a $\mu$P guess and wants to know how sensitive performance is to HP variation around it.
\end{enumerate}

The threshold-based and contiguous views answer different questions. Threshold-based requires loss values for every cell (post-hoc), while contiguous-around-$\mu$P requires only a starting point (a-priori). The two contiguous views (around optimum vs around $\mu$P) coincide when $\mu$P transfers to the per-scale optimum exactly, and diverge when it does not.

\begin{figure}[h]
  \centering
  \includegraphics[width=\linewidth]{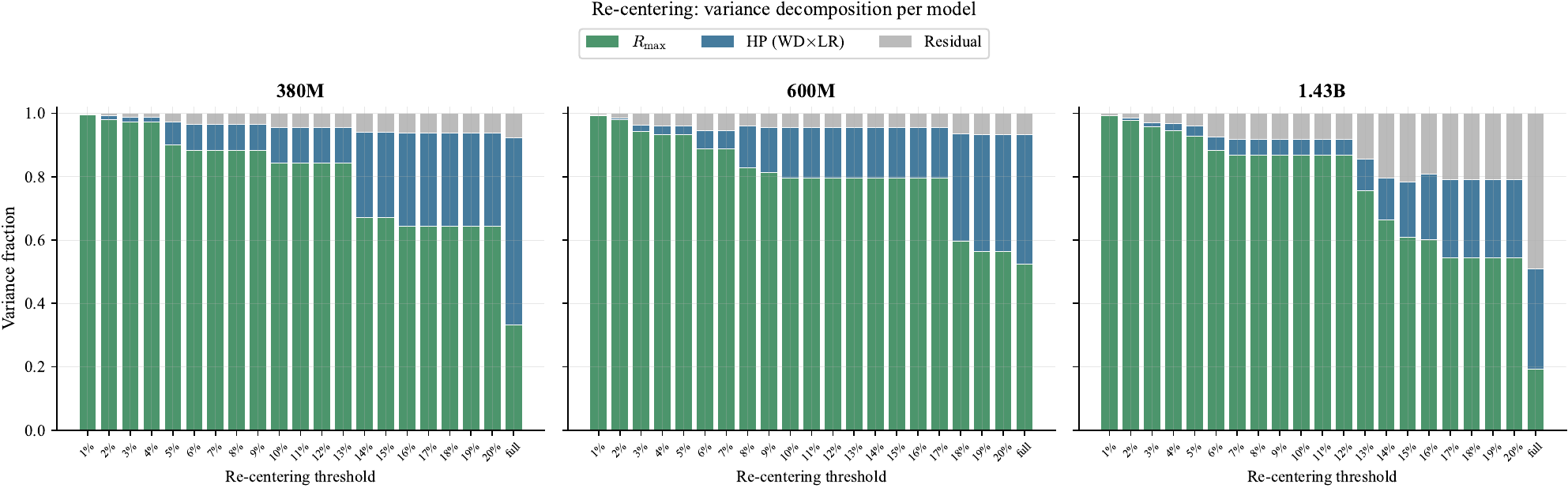}
  \caption{Threshold-based re-centering: two-way ANOVA variance shares vs.\ relative threshold $\tau$ (\% above best HP mean loss). At every scale, $R_\text{max}$ (green) dominates within reasonable thresholds. The full-grid HP-dominance at 1.43B (rightmost bar) disappears as soon as $\tau$ tightens.}
  \label{fig:recentering_stacked_bars}
\end{figure}

\paragraph{View 1: threshold-based re-centering.}
For each scale, we sort all HP configurations by their mean Arabic VL across $R_\text{max}$ levels and keep those within a relative threshold $\tau$ of the best. We then run two-way ANOVA on the kept cells. Figure~\ref{fig:recentering_stacked_bars} shows the variance decomposition across thresholds.

The pattern is consistent across scales: $R_\text{max}$ explains over 95\% of variance at the tightest threshold ($\tau = 1\%$), 80--90\% at $\tau = 10\%$, and remains the dominant factor up to $\tau \approx 14\%$. At 1.43B specifically, the full-grid pattern (HP dominance) reverses cleanly under any reasonable re-centering threshold. Within the operational region a practitioner who has done basic HP screening would occupy, mixing dominates.

\begin{figure}[h]
  \centering
  \includegraphics[width=0.6\linewidth]{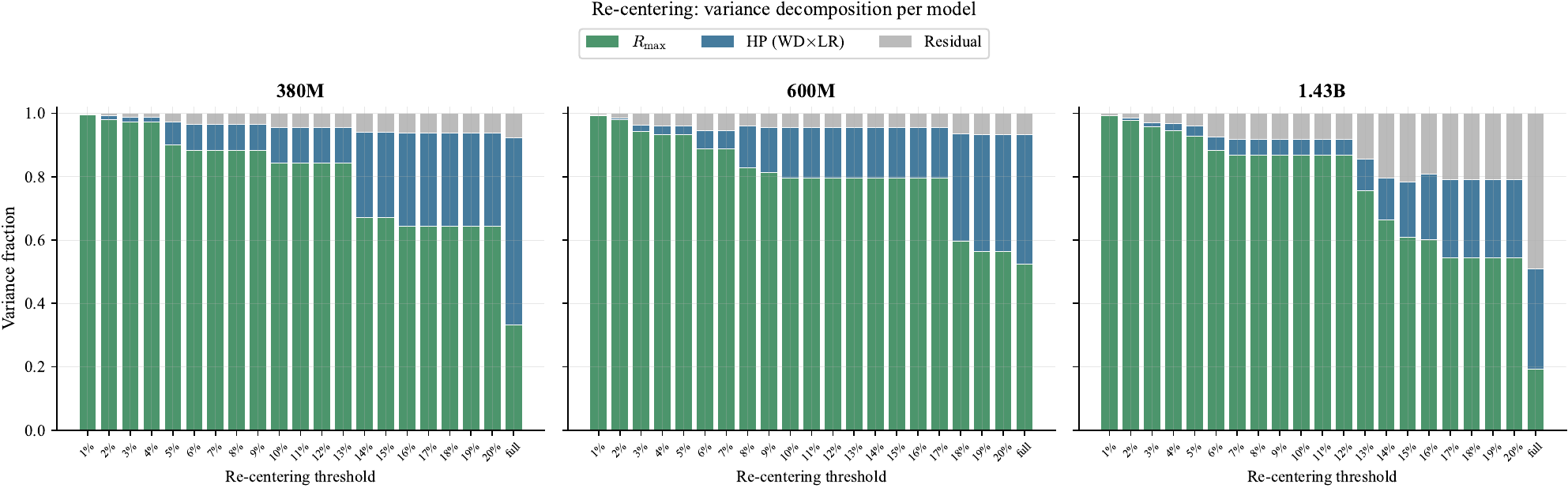}
  \caption{Companion view to Fig.~\ref{fig:recentering_stacked_bars}: $R_\text{max}$ ANOVA share as a function of the threshold $\tau$ (solid lines, left axis), with the number of HP configurations retained at each threshold overlaid (dashed lines, right axis). All three curves start near 100\% at $\tau = 1\%$. 1.43B (red) tracks 380M and 600M closely up to $\tau \approx 12\%$, then drops sharply between $\tau = 12\%$ ($R_\text{max} = 87\%$, 13/25 HPs) and $\tau = 14\%$ ($R_\text{max} = 52\%$, 15/25 HPs): two configurations entering the grid at 13--14\% above optimal are enough to cut $R_\text{max}$'s share substantially. Used by App.~\ref{app:hp_landscape} to argue that the 1.43B sharpening is broad rather than outlier-driven.}
  \label{fig:hp_recentering}
\end{figure}

The caveat is that this view assumes the practitioner can rank HPs by performance, which requires already having trained them all. It is informative for diagnosis but does not directly answer ``what should I do before running the full grid.''

\begin{figure}[h]
  \centering
  \includegraphics[width=\linewidth]{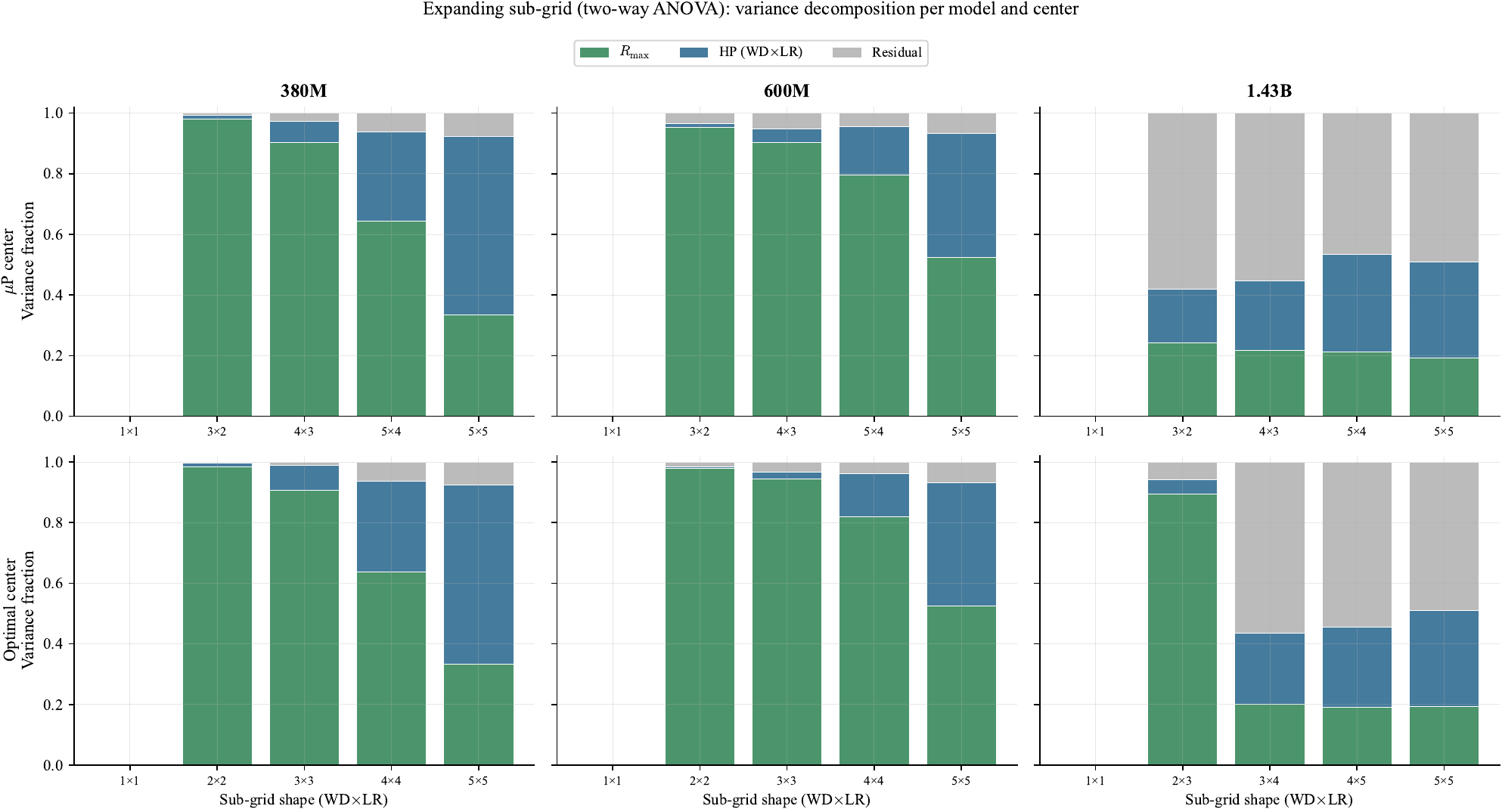}
  \caption{Contiguous neighborhood re-centering: two-way ANOVA variance shares as the sub-grid expands from the chosen center. \emph{Top row}: $\mu$P center. \emph{Bottom row}: per-scale optimum center. At 380M and 600M, the two centers give similar pictures: $R_\text{max}$ dominates at small radii. At 1.43B, the two centers diverge sharply: optimum-centered (bottom right) shows clean $R_\text{max}$ dominance at radius 1, but $\mu$P-centered (top right) shows mixed $R_\text{max}$/HP variance and a large residual ($\sim$50\%) at every radius.}
  \label{fig:expanding_grid_2way_stacked_bars}
\end{figure}

\paragraph{Views 2 and 3: contiguous neighborhood around optimum vs $\mu$P.}
For each scale we pick a center HP (the per-scale optimum, or the $\mu$P HP) and expand a contiguous sub-grid radially: the $r$-th sub-grid contains $(r{+}1) \times (r{+}1)$ HP cells centered on the chosen point. We run two-way ANOVA on each sub-grid. Figure~\ref{fig:expanding_grid_2way_stacked_bars} shows the result.

At 380M and 600M, the two centers tell the same story: at small radii ($\leq 3{\times}3$), $R_\text{max}$ explains over 90\% of variance; HP variance only catches up as the sub-grid widens to include increasingly suboptimal configurations. The HP landscape is locally flat around both centers.

\begin{figure}[h]
  \centering
  \includegraphics[width=\linewidth]{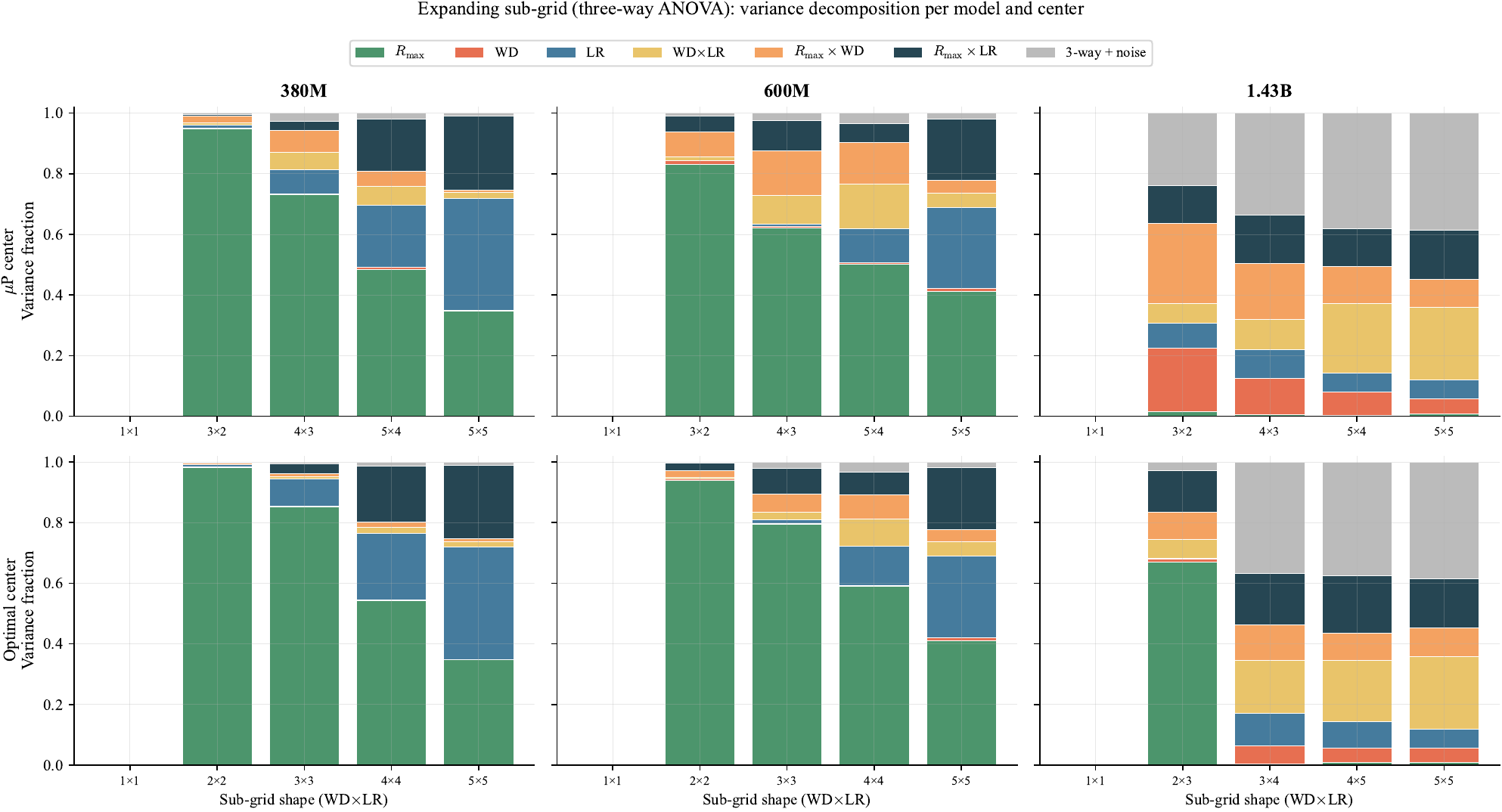}
  \caption{Same as Fig.~\ref{fig:expanding_grid_2way_stacked_bars} but with the three-way decomposition. At 1.43B, the $\mu$P-centered view (top right) is dominated by $R_\text{max} \times \text{LR}$ and $\text{WD} \times \text{LR}$ interactions; the optimum-centered view (bottom right) is cleaner.}
  \label{fig:expanding_grid_3way_stacked_bars}
\end{figure}

At 1.43B, the two centers diverge sharply. Centered on the per-scale optimum ($\lambda = 0.01$, $\eta = 0.003$), the smallest sub-grid (radius 1, $2{\times}3$ cells) shows $R_\text{max} = 89\%$, HP $= 5\%$. Centered on the $\mu$P HP ($\lambda = 0.1$, $\eta = 0.01$), the smallest sub-grid shows $R_\text{max} = 24\%$, HP $= 18\%$, with a large residual ($\sim$50\%). The three-way decomposition (Fig.~\ref{fig:expanding_grid_3way_stacked_bars}) confirms that the residual mass is concentrated in the $\lambda{\times}\eta$ and $R_\text{max}{\times}\lambda$ interactions, not in main effects.

Two findings emerge. First, at 380M and 600M, the $\mu$P HP and the per-scale optimum sit in the same locally flat basin: a practitioner using $\mu$P faces the same low HP sensitivity as a practitioner who has tuned. Second, at 1.43B, the two points have separated. The $\mu$P HP is in a region with substantially higher local HP sensitivity than the per-scale optimum. The basin around the actual optimum has shrunk below the precision of $\mu$P transfer.

What changes between 380M/600M and 1.43B that pushes the $\mu$P point outside the per-scale optimum's basin? The three-way decomposition of the $\mu$P-centered sub-grid above already hints at the answer: the residual mass is dominated by $\lambda \times \eta$ and $R_\text{max} \times \lambda$ interactions that are absent at smaller scales. Appendix~\ref{app:hp_landscape} investigates this mechanism on the full grid and shows that what looks like ``HP variance'' at 1.43B is actually interaction variance: the loss landscape develops a coupled $(\lambda, \eta)$ structure as scale grows.

\paragraph{Why $\mu$P remains the right default in all three views.}
The three views give different answers about how much HP sensitivity matters at 1.43B (low, low-but-cleaner, and substantial), but they converge on the same recommendation:

\begin{itemize}
  \item If HP sensitivity is low (380M, 600M, or threshold-based view at any scale), $\mu$P HPs are within the flat basin, so per-scale tuning yields negligible improvement. Compute is better spent on mixing-ratio search.
  \item If HP sensitivity is high (1.43B, $\mu$P-centered view), the local landscape is jagged. The per-scale optimum is reachable but not by any cheap tuning protocol: a wide grid search at 1.43B exposes the practitioner to large variance from interaction-dominated regions of HP space. $\mu$P provides a known-decent point at zero search cost; further tuning is a high-variance bet that may or may not pay off.
\end{itemize}

In both regimes, $\mu$P is the safe default. When HP does not matter, the choice is free; when HP matters, $\mu$P avoids a costly and risky search. Practitioners with extra compute at scale can invest in per-scale HP refinement, but the marginal return is uncertain compared to the well-characterized return on mixing-ratio tuning.

\begin{figure}[h]
  \centering
  \includegraphics[width=0.95\linewidth]{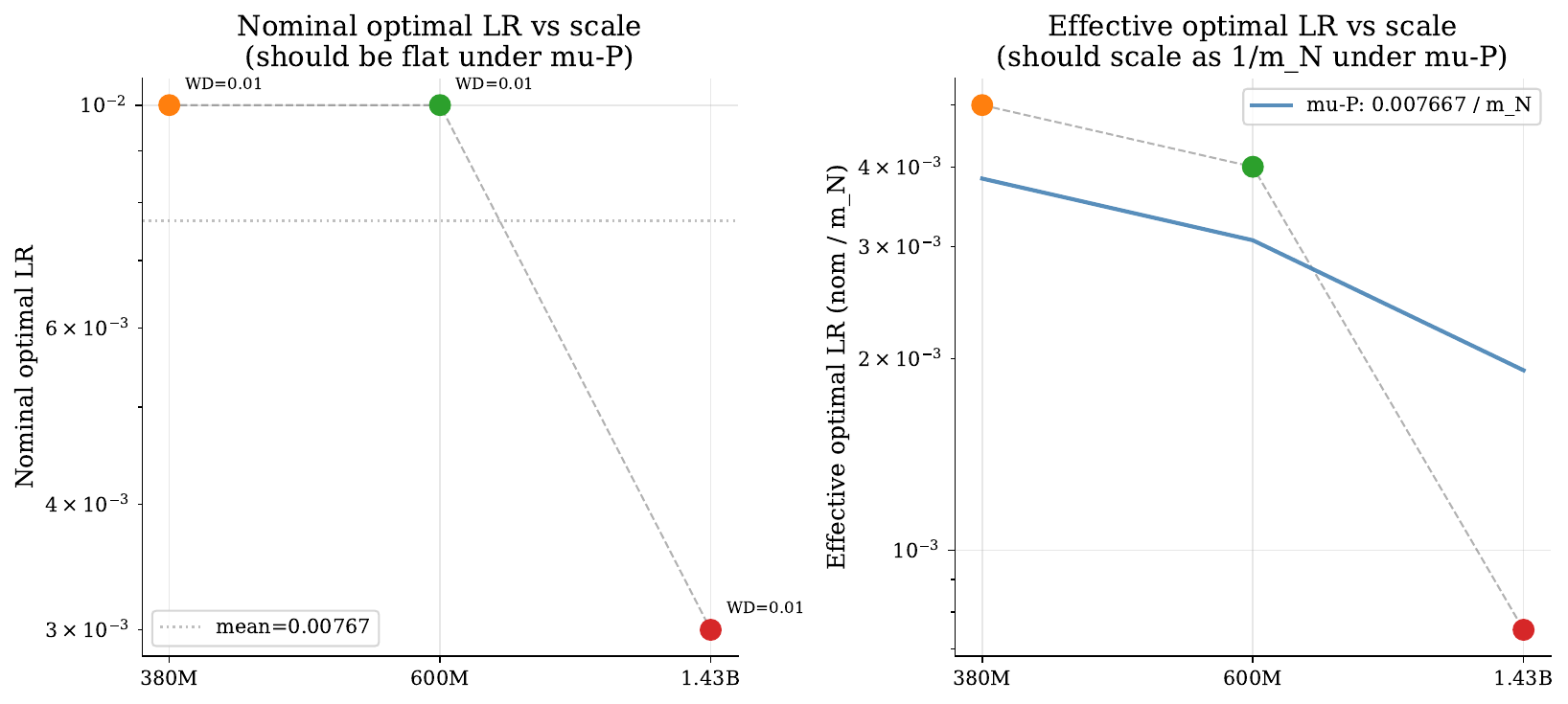}\\[0.4em]
  \includegraphics[width=0.95\linewidth]{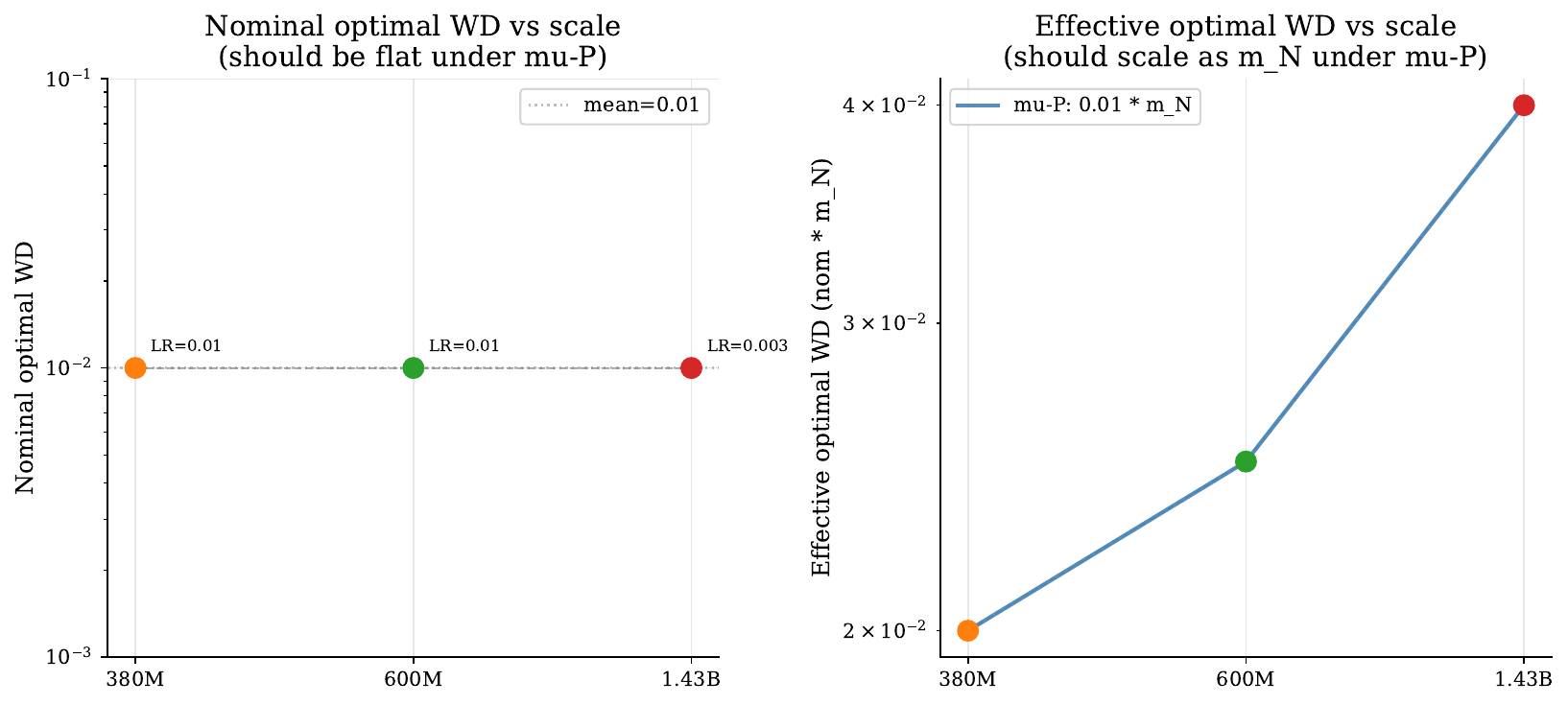}
  \caption{$\mu$P transfer test for LR (top) and weight decay (bottom), both as nominal (left column) vs effective (right column). \emph{Top row, LR}: nominal $\eta^\star$ is the best base grid value; effective is $\eta^\star / m_N$. Under perfect $\mu$P the nominal is flat and the effective scales as $\eta_\text{base} / m_N$ (blue reference curve). Observed: $\eta^\star = 10^{-2}$ at 380M and 600M (on curve), $3 \times 10^{-3}$ at 1.43B (below curve), so $\mu$P fails at 1.43B for LR. \emph{Bottom row, WD}: nominal $\lambda^\star$ is flat at $10^{-2}$ at every scale, and the effective $\lambda^\star \cdot m_N$ sits exactly on the $\mu$P reference curve $10^{-2} \cdot m_N$ at 380M, 600M, and 1.43B, so $\mu$P transfer for weight decay holds at every scale. The $\mu$P failure at 1.43B is therefore LR-specific. Caveat: $\lambda = 0.01$ is the smallest grid value, so the apparent flatness is consistent with but not uniquely diagnostic of perfect transfer.}
  \label{fig:optimal_hp_drift}
\end{figure}

\paragraph{A note on grid design.}
Our $5 \times 5$ HP grid was fixed across all model scales: $\lambda \in \{0.01, 0.1, 0.3, 1.0, 3.0\}$ and $\eta \in \{0.0001, 0.0003, 0.001, 0.003, 0.01\}$. Figure~\ref{fig:optimal_hp_drift} tests whether the per-scale optimal HPs match the $\mu$P prediction, separately for LR and weight decay. $\mu$P \citep{yang2022tensorprograms} defines base HPs at a reference scale (150M in our setup, $d_\text{model} = 512$) and a width multiplier $m_N = d_\text{model}(N) / 512$; the per-layer LR at scale $N$ is $\eta_\text{base} / m_N$ and the per-layer weight decay is $\lambda_\text{base} \cdot m_N$. The central promise is that the base HPs are scale-invariant, so tuning once at 150M transfers unchanged. Our grid values are the base HPs (pre-$\mu$P); $\mu$P applies the rescaling internally during training. For each HP we plot the nominal optimum (the best base grid value) and the effective optimum (the per-layer HP actually applied). Under perfect $\mu$P, the nominal curve is flat and the effective curve follows the $\mu$P scaling ($\eta_\text{base} / m_N$ for LR, $\lambda_\text{base} \cdot m_N$ for WD).

For weight decay, $\mu$P transfers perfectly at every scale (Fig.~\ref{fig:optimal_hp_drift}, bottom row): the nominal optimum $\lambda^\star$ is flat at $10^{-2}$ across 380M, 600M, and 1.43B, and the effective $\lambda^\star \cdot m_N$ sits exactly on the $\mu$P reference curve $10^{-2} \cdot m_N$ at every scale. One caveat, $\lambda = 0.01$ is the smallest value in our grid, so the true optimum could lie below it and we would not see it; the flatness we observe is consistent with perfect $\mu$P transfer but does not uniquely identify it. For LR (Fig.~\ref{fig:optimal_hp_drift}, top row), $\mu$P holds at 380M and 600M: nominal $\eta^\star = 10^{-2}$ at both scales, and the effective sits on the $\mu$P reference curve $\eta_\text{base} / m_N$. At 1.43B the nominal drops to $\eta^\star = 3 \times 10^{-3}$, a $\sim$3$\times$ shift, and the effective correspondingly falls below the $\mu$P curve. The $\mu$P failure at 1.43B is therefore LR-specific, not a general breakdown of $\mu$P transfer.

The LR-specific drift matches the downward shift of the marginal $\eta$ minimum in Fig.~\ref{fig:hp_axis_decomposition} (top-left), which reaches $\eta \approx 10^{-3}$ at 1.43B under a different aggregation, so the effect is robust to aggregation choice and is not a grid-noise artifact. One consequence is that the per-scale optimum frequently lands on grid boundaries: $\eta = 0.01$ (the largest grid value) at smaller scales, and $\lambda = 0.01$ (the smallest grid value) at every scale.

The high-level conclusion is unaffected: practitioners should default to $\mu$P, and the mixing-vs-HP comparison in \S\ref{sec:hp_sensitivity} remains valid because both axes are measured on the same grid.

\FloatBarrier

\section{HP landscape sharpening at scale}
\label{app:hp_landscape}

Appendix~\ref{app:operational_sensitivity} and the main-body HP landscape figure (Fig.~\ref{fig:mup_transfer}) show the same pattern: at 380M and 600M the HP landscape is broadly flat, and the $\mu$P HP sits in the same wide low-loss basin as the per-scale optimum. At 1.43B the near-optimal basin shrinks around the per-scale optimum. $\mu$P still reaches near-optimal loss (Table~\ref{tab:mup_transfer}). This section investigates what drives the narrowing. The short answer: at 1.43B, pairwise couplings across $R_\text{max}$, $\lambda$, and $\eta$ become material on the full grid, and an additional three-way residual, as we show below, is largely concentrated in a handful of pathological cells. Main effects (the sum of the individual contributions of $R_\text{max}$, $\lambda$, and $\eta$ taken one at a time) explain only 12\% of full-grid variance, versus 84\% at 380M. Pairwise interactions add 50\% spread across all three pairs, and a three-way residual of 39\% (essentially absent at smaller scales) captures joint structure across $R_\text{max}$, $\lambda$, and $\eta$ that no single pair can explain. We walk through the evidence below.

Appendix~\ref{app:operational_sensitivity} already decomposed the HP contribution at one specific anchor: the $\mu$P-centered neighborhood at 1.43B, where $\lambda \times \eta$ and $R_\text{max} \times \lambda$ interactions dominated. This section extends that analysis to the full grid at each scale, so we can track how each piece shifts as the landscape narrows: the marginal effect of $\lambda$, the marginal effect of $\eta$, and the $\lambda \times \eta$ interaction. We first quantify the narrowing itself, then decompose the HP factor axis by axis, then separate main effects from the interaction with a three-way ANOVA, and finally check that the result is not an artifact of a few pathological cells that inflate $\text{SS}_\text{HP}$.

\paragraph{Measuring landscape flatness.}
For each model scale, we compute the mean Arabic validation loss of each $(\lambda, \eta)$ combination across all $R_\text{max}$ levels where it has data. Let $\ell^*$ denote the best such mean. For a relative threshold $\tau$ (expressed as a percentage), we count the number of HP configurations $h$ satisfying $\bar{\ell}_h \leq \ell^* \cdot (1 + \tau)$. A flat landscape has many configurations near the optimum; a sharp one has few.

\begin{table}[t]
  \centering
  \begin{minipage}[t]{0.52\linewidth}
    \centering
    \captionof{table}{Number of HP configurations (out of 25) within a relative threshold of the best. Higher = flatter landscape.}
    \label{tab:hp_landscape_flatness}
    \begin{tabular}{lcccccc}
      \toprule
      Model & 1\% & 2\% & 5\% & 10\% & 15\% & 20\% \\
      \midrule
      380M  & 2 & 6  & 9  & 15 & 18 & 20 \\
      600M  & 3 & 5  & 12 & 20 & 20 & 24 \\
      1.43B & 2 & 4  & 10 & 13 & 16 & 21 \\
      \bottomrule
    \end{tabular}
  \end{minipage}%
  \hfill
  \begin{minipage}[t]{0.44\linewidth}
    \centering
    \captionof{table}{HP axis range as \% of the best marginal mean loss. At 380M and 600M, $\eta$ dominates and $\lambda$ is negligible. At 1.43B, both axes become sensitive: $\eta$ rises back to 27\% and $\lambda$ jumps from $\approx$4\% to 20\%.}
    \label{tab:hp_axis_decomposition}
    \begin{tabular}{lccc}
      \toprule
      Model & $\eta$ (\%) & $\lambda$ (\%) & $\eta$ share \\
      \midrule
      380M  & 32.4 & 4.0 & 89\% \\
      600M  & 16.1 & 3.9 & 81\% \\
      1.43B & 27.3 & 19.9 & 58\% \\
      \bottomrule
    \end{tabular}
  \end{minipage}
\end{table}

Table~\ref{tab:hp_landscape_flatness} and Fig.~\ref{fig:hp_recentering} (dashed lines, right axis) show the same pattern. We say the landscape \emph{sharpens} with scale when fewer HP configurations survive a given threshold $\tau$: the near-optimal basin holds fewer candidates as the model grows. By this measure, 1.43B sharpens against 600M at every threshold we tested. The comparison with 380M is tighter, with the two scales competitive at the tightest and loosest thresholds, so the narrowing is clearest against 600M. We turn next to which HP axis drives it.

\paragraph{Which HP axis drives the sharpening?}
The HP landscape has two axes, $\lambda$ and $\eta$. For each axis, the sharpening at 1.43B could come from one of two changes. (i) The marginal minimum along that axis drifts away from its $\mu$P base value, so $\mu$P no longer lands at the axis optimum. (ii) The \emph{range} of the marginal loss along the axis (worst minus best value) grows, so deviating from the minimum costs more than it did at smaller scales. We check both quantities for both axes below.

\begin{figure}[t]
  \centering
  \includegraphics[width=\linewidth]{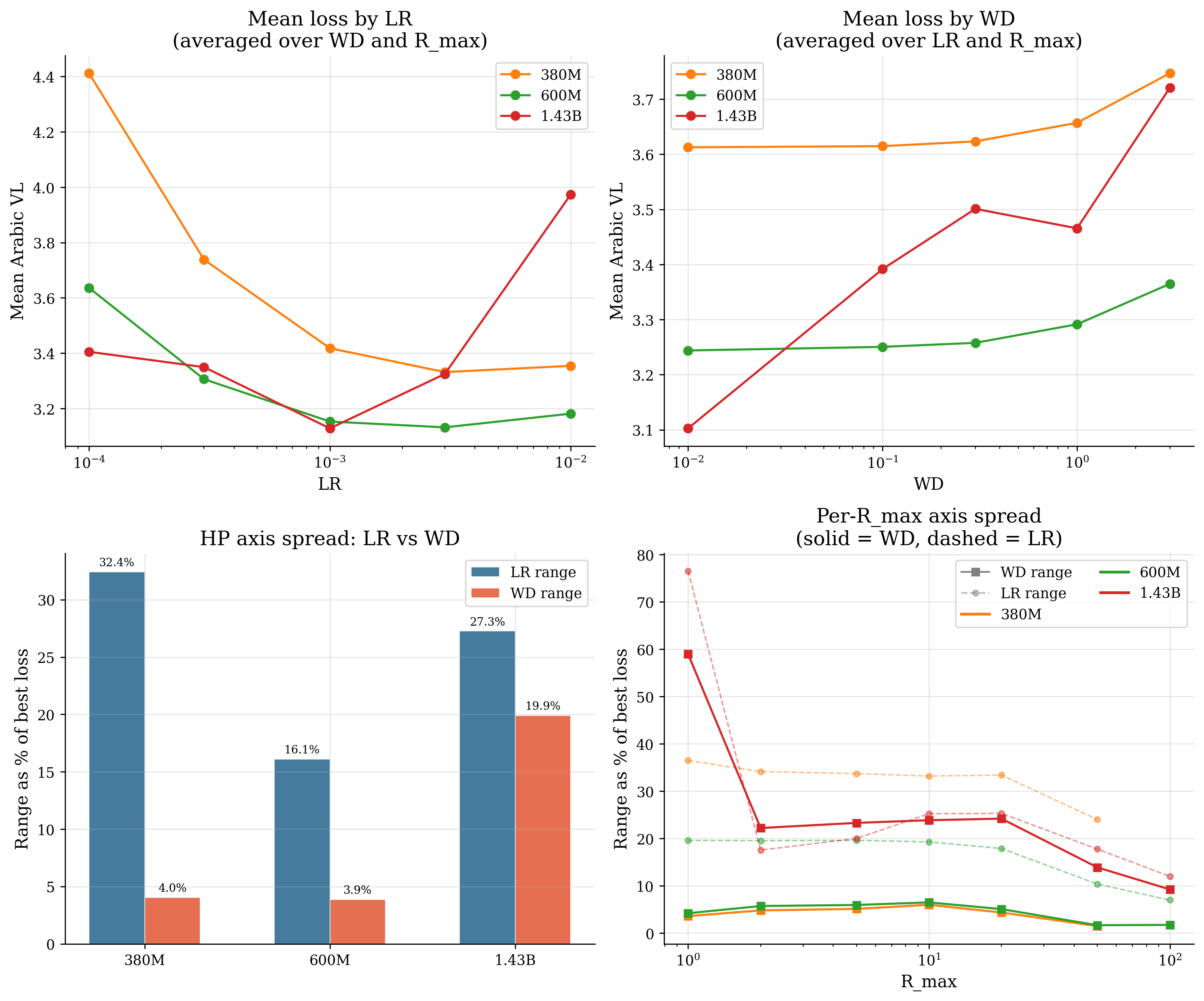}
  \caption{HP axis decomposition. Let $\bar\ell(\eta) = \operatorname{mean}_{\lambda, R_\text{max}} \ell$ denote the mean Arabic VL along the $\eta$ axis (averaged over $\lambda$ and $R_\text{max}$), and $\bar\ell(\lambda)$ analogously. The \emph{axis range} is $\operatorname{Range}(\eta) = \bigl(\max_\eta \bar\ell(\eta) - \min_\eta \bar\ell(\eta)\bigr) / \min_\eta \bar\ell(\eta)$ (in \%); $\operatorname{Range}(\lambda)$ defined analogously. \emph{Top row}: $\bar\ell(\eta)$ (left) and $\bar\ell(\lambda)$ (right) at each scale. \emph{Bottom-left}: the two axis ranges per scale. A 20\% range means the worst marginal along that axis is 20\% above the best. \emph{Bottom-right}: the same two ranges, but recomputed at each $R_\text{max}$ separately (solid = $\lambda$, dashed = $\eta$); this checks whether the full-grid pattern holds conditional on $R_\text{max}$.}
  \label{fig:hp_axis_decomposition}
\end{figure}

The top panels of Fig.~\ref{fig:hp_axis_decomposition} report where each axis's marginal minimum sits. On $\eta$, the minimum drifts roughly an order of magnitude below $\eta_\text{base}$ between 380M and 1.43B (consistent with the nominal optimum drop in Fig.~\ref{fig:optimal_hp_drift}); on $\lambda$, it stays pinned at the smallest grid value at every scale. The bottom-left panel and Table~\ref{tab:hp_axis_decomposition} report each axis's range: $\eta$'s range is comparable across scales, so $\eta$ is \emph{not} becoming sharper; $\lambda$'s range jumps roughly $5\times$ between 600M and 1.43B. The bottom-right panel recomputes both ranges at each $R_\text{max}$ separately and reproduces the same pattern, so the $\lambda$ jump is not an averaging artifact.

Two distinct effects therefore drive the sharpening at 1.43B. On $\eta$, the minimum drifts away from $\eta_\text{base}$, so the $\mu$P choice sits off-optimum along that axis. On $\lambda$, the range widens, so weight-decay values that were indistinguishable at smaller scales now differ materially. The primary driver is $\lambda$; $\eta$'s drift is a secondary contribution.

\FloatBarrier

\paragraph{Three-way ANOVA: decomposing the HP factor.}
The two-way ANOVA treats $(\lambda, \eta)$ as a single 25-level factor. This conflates the individual effects of $\lambda$ and $\eta$ with their interaction. To disentangle these, we run a three-way ANOVA with $R_\text{max}$, $\lambda$, and $\eta$ as separate factors, including all two-way interactions. The residual captures the three-way interaction and noise.

\begin{figure}[t]
  \centering
  \includegraphics[width=\linewidth]{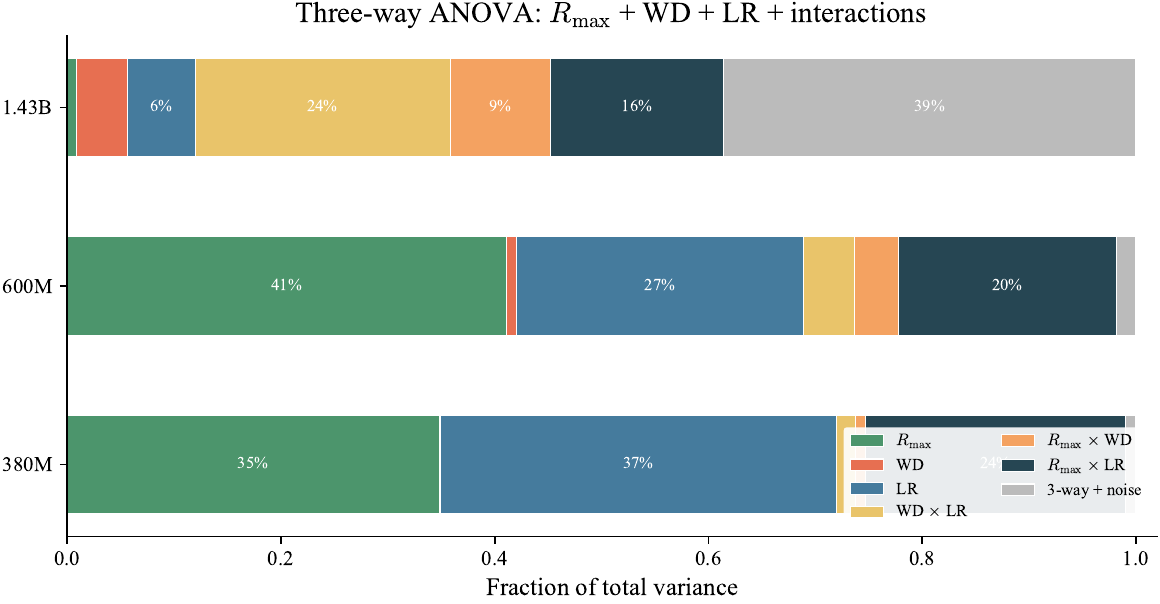}
  \caption{Three-way ANOVA decomposition of Arabic VL variance into main effects ($R_\text{max}$, $\lambda$, $\eta$), two-way interactions ($\lambda \times \eta$, $R_\text{max} \times \lambda$, $R_\text{max} \times \eta$), and a three-way residual (three-way interaction plus noise). At 380M and 600M, main effects dominate ($>$69\%). At 1.43B, main effects drop to 12\%, pairwise interactions add 50\% (spread across all three pairs), and a three-way residual of 39\% captures joint structure that no single pair explains.}
  \label{fig:three_way_anova}
\end{figure}

\begin{table}[t]
  \caption{Three-way ANOVA: fraction of Arabic VL variance explained by each main effect, pairwise interaction, and the three-way residual (three-way interaction plus noise). At 380M and 600M, main effects (especially $R_\text{max}$ and $\eta$) dominate and the three-way residual is negligible. At 1.43B, no single term dominates: variance is spread across pairwise interactions (50\% total) and a large three-way residual (39\%).}
  \label{tab:three_way_anova}
  \centering
  \begin{tabular}{lccccccc}
    \toprule
    Model & $R_\text{max}$ & $\lambda$ & $\eta$ & $\lambda \times \eta$ & $R_\text{max} \times \lambda$ & $R_\text{max} \times \eta$ & Residual \\
    \midrule
    380M  & 30.2\% & 0.1\% & 53.8\% & 2.9\%  & 1.0\% & 11.0\% & 1.1\% \\
    600M  & 41.1\% & 1.0\% & 26.8\% & 4.8\%  & 4.1\% & 20.4\% & 1.8\% \\
    1.43B & 0.9\%  & 4.8\% & 6.3\%  & 23.9\% & 9.4\% & 16.2\% & 38.6\% \\
    \bottomrule
  \end{tabular}
\end{table}

Figure~\ref{fig:three_way_anova} and Table~\ref{tab:three_way_anova} show the results. At 380M, main effects explain 84\% of variance, with $\eta$ as the dominant factor (54\%) and $R_\text{max}$ second (30\%). At 600M, $R_\text{max}$ grows to 41\% and $\eta$ drops to 27\%, but main effects still account for 69\%. At 1.43B, the picture is qualitatively different. Main effects explain only 12\% of variance. Pairwise interactions add 50\%, distributed across $\lambda \times \eta$ (24\%), $R_\text{max} \times \eta$ (16\%), and $R_\text{max} \times \lambda$ (9\%). The remaining 39\% sits in the three-way residual, which at smaller scales was negligible ($<$2\%).

The practical reading: on the full 1.43B grid, the loss landscape is no longer approximately additive in any single hyperparameter. At 380M and 600M, $\lambda$ and $\eta$ are approximately independent ($\lambda \times \eta < 5\%$ of variance), so a practitioner can tune them along separate 1D sweeps and compose the results. At 1.43B this factorisation breaks. The $\lambda \times \eta$ interaction alone carries 24\% of variance, more than twice the sum of the individual $\lambda$ and $\eta$ main effects (11\%). Concretely, the best $\lambda$ depends on which $\eta$ is chosen and vice versa: marginalising over one axis returns a coordinate that is suboptimal at the other. A practitioner who tunes $\eta$ first at fixed $\lambda$, then $\lambda$ at fixed $\eta$, can miss the joint $(\lambda, \eta)$ optimum by several percent at this scale. The coupling extends to $R_\text{max}$: the $R_\text{max} \times \lambda$ and $R_\text{max} \times \eta$ interactions add another 25\% of variance, so the joint $(\lambda, \eta)$ optimum itself shifts with the mixing ratio. The remaining 39\% sits in the three-way residual, capturing variance no pair can predict; the next paragraph shows that this residual is largely concentrated in 5 outlier cells, so the bulk of the grid is pairwise-structured ($R_\text{max} \times \lambda$ and $\lambda \times \eta$) rather than fully non-separable. In operational terms, per-scale tuning at 1.43B still couples all three axes: the $R_\text{max} \times \lambda$ interaction means that the best $R_\text{max}$ is not strictly independent of $(\lambda, \eta)$. The $\mu$P recipe keeps this penalty small: transferring HPs from the 150M proxy and tuning $R_\text{max}$ at those HPs lands within 0--3.4\% of the per-scale optimum (Table~\ref{tab:mup_transfer}), and the 0--3.4\% residual is exactly what the $R_\text{max} \times \lambda$ coupling predicts.

\begin{figure}[t]
  \centering
  \includegraphics[width=0.55\linewidth]{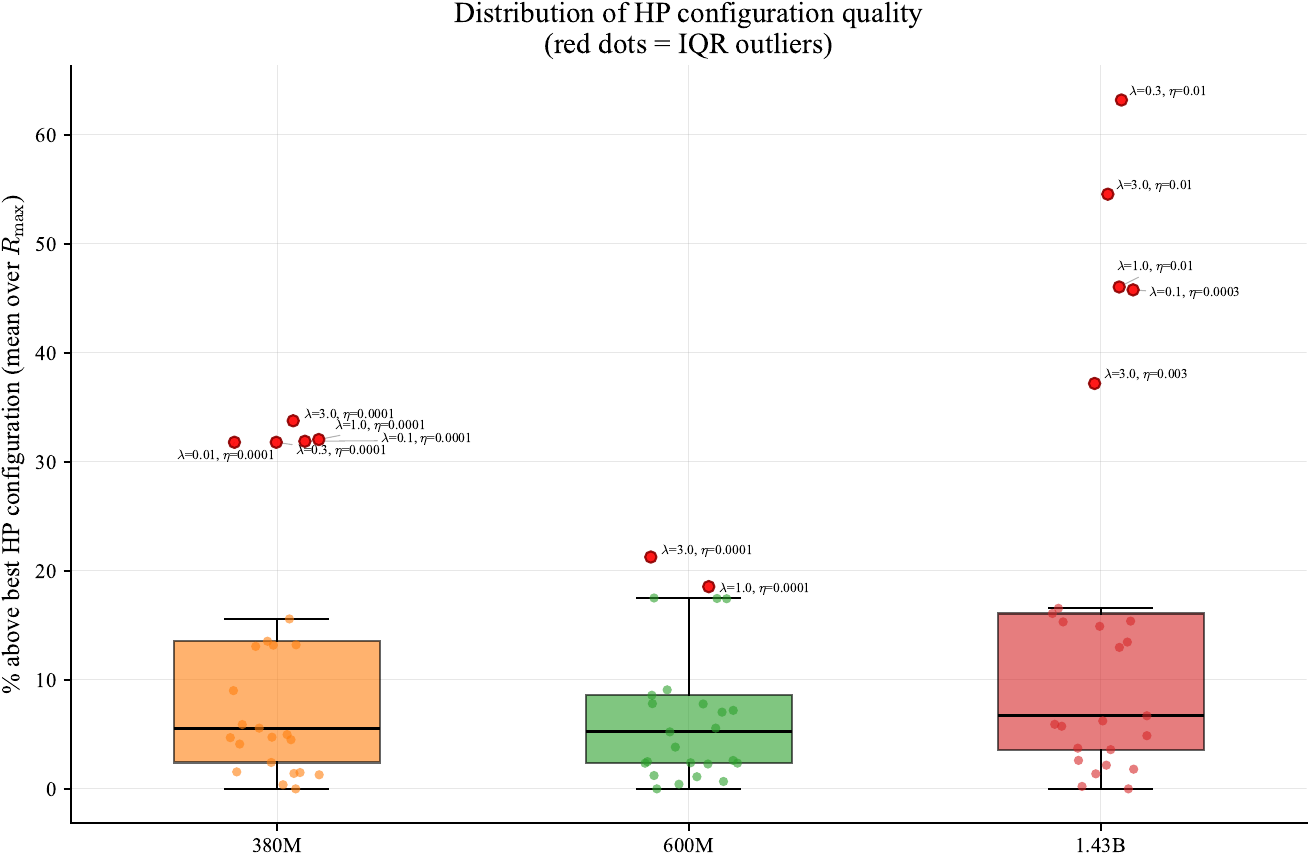}
  \caption{Distribution of HP configuration quality at each scale. Each point is one $(\lambda, \eta)$ combination, measured as \% above the best. Boxes show the IQR; red dots are statistical outliers (beyond $1.5 \times \text{IQR}$).}
  \label{fig:hp_boxplot}
\end{figure}

\paragraph{Is the 1.43B sharpening driven by a few pathological cells?}
The 39\% three-way residual raises a natural question: is the non-separability spread across the grid, or concentrated in a few cells? We flag IQR outliers on the full grid: for each $(\lambda, \eta)$ cell, we take its mean Arabic VL across all $R_\text{max}$ levels as a percentage above the best configuration, and mark cells beyond $1.5 \times$ IQR. Figure~\ref{fig:hp_boxplot} shows the resulting quality distribution: 5 outliers at 380M, 2 at 600M, and 5 at 1.43B.

\begin{figure}[t]
  \centering
  \includegraphics[width=0.9\linewidth]{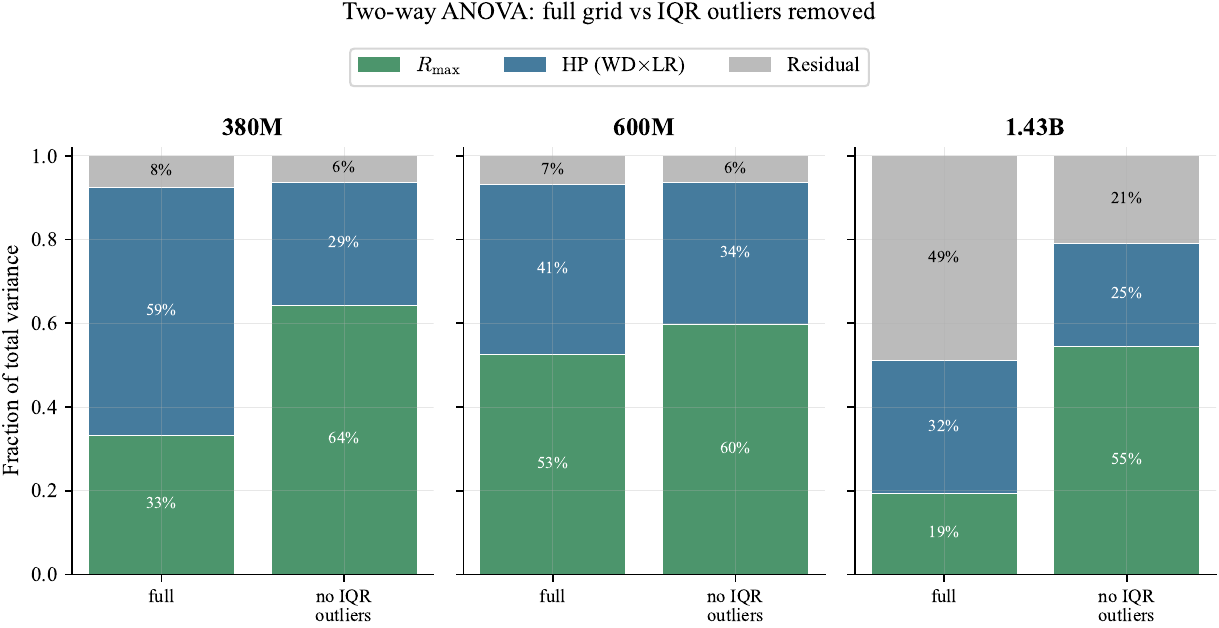}
  \vspace{0.5em}
  \includegraphics[width=0.9\linewidth]{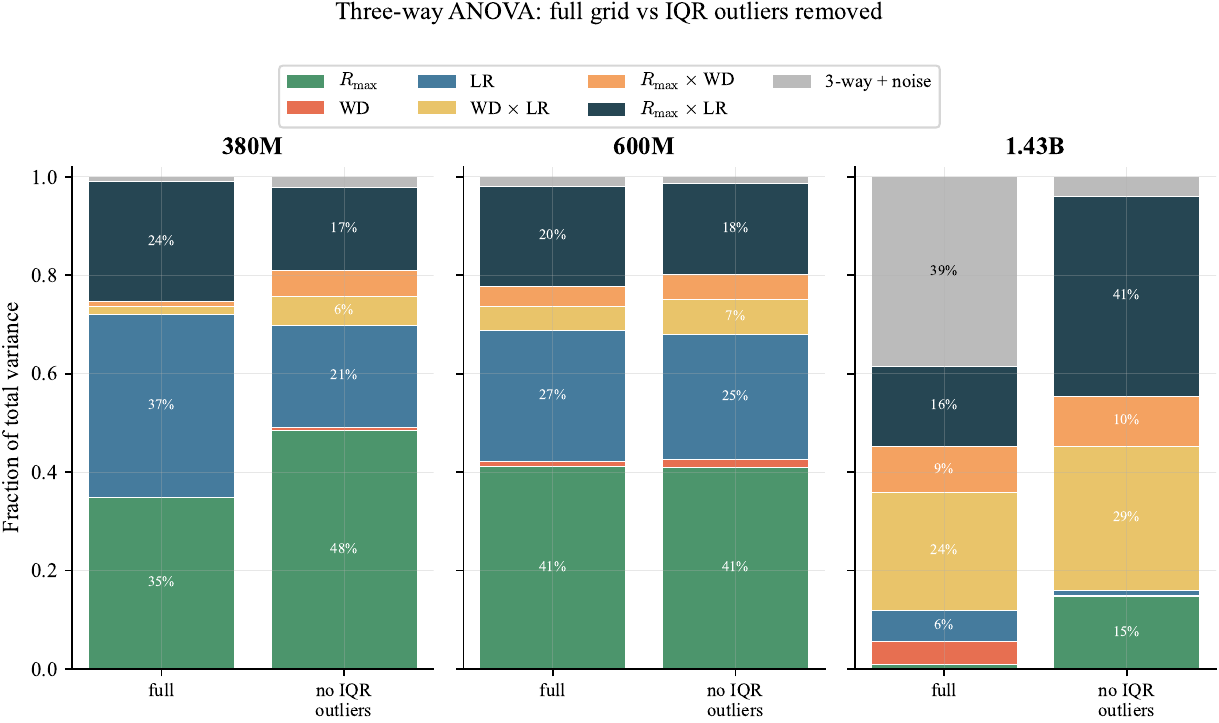}
  \caption{ANOVA decomposition of Arabic VL variance, full grid vs.\ IQR outliers removed. \emph{Top}: two-way ANOVA treating $(\lambda, \eta)$ as a single HP factor. \emph{Bottom}: three-way ANOVA with $R_\text{max}$, $\lambda$, $\eta$ as separate factors plus all two-way interactions. Removing outliers flips the 380M picture entirely ($R_\text{max}$ from 30\% to 63\% in two-way) and recovers $R_\text{max}$ as the dominant factor at 1.43B (19\% to 55\%); the 1.43B three-way residual collapses from 39\% to $\sim$4\%, with what remains dominated by $R_\text{max} \times \lambda$ (41\%) and $\lambda \times \eta$ (29\%).}
  \label{fig:outlier_anova}
\end{figure}

Figure~\ref{fig:outlier_anova} re-runs the ANOVA with and without those outliers, under both the two-way decomposition ($R_\text{max}$ vs.\ the HP factor) and the three-way decomposition. The answer to the opening question splits along the two views.

\emph{Three-way: outlier-driven.} The three-way residual at 1.43B collapses from 39\% to $\sim$4\% once the 5 cells are removed, and the rest of the grid becomes pairwise-dominated ($R_\text{max} \times \lambda$ and $\lambda \times \eta$). Mechanically, those 5 cells fail at particular $(\lambda, \eta, R_\text{max})$ triples rather than along any single axis or pair, and an ANOVA routes variance of that kind into the three-way residual, which is why removing them empties it.

\emph{Two-way: broad.} The HP-factor share at 1.43B drops from 32\% to 25\% when the outliers are removed, not to zero. The HP-dominance flagged in the main body (Q1) is therefore a broad property of the 1.43B grid, not an artifact of a handful of cells. The threshold-based and optimum-centered re-centering views in App.~\ref{app:operational_sensitivity} agree on independent definitions, and the $\mu$P cell $(\lambda{=}0.01, \eta{=}0.01)$ sits far from the outlier region at every scale, so the HP-dominance that survives outlier removal is what shapes the landscape around $\mu$P.

\paragraph{Conclusion: the narrowing is genuine, but $\mu$P transfer still works.}
This section built up a working definition of ``narrowing'' and tested it with independent views. The narrowing is the shrinking of the near-optimal basin: at a relative threshold $\tau$, fewer HP configurations lie within $\tau\%$ of the best. Table~\ref{tab:hp_landscape_flatness} shows this directly (1.43B has fewer configurations within the 2--15\% thresholds than 380M or 600M); Fig.~\ref{fig:outlier_anova} shows the sharpening is not fully explained by outliers (the two-way HP share at 1.43B drops from 32\% to 25\% after removing the 5 IQR outliers, not to zero); Table~\ref{tab:hp_axis_decomposition} shows the $\lambda$ range jumps $5\times$ between 600M and 1.43B while the $\eta$ optimum drifts an order of magnitude below $\eta_\text{base}$ (Fig.~\ref{fig:optimal_hp_drift}, Fig.~\ref{fig:hp_axis_decomposition}); and the three-way ANOVA shows main effects dropping from 84\% at 380M to 12\% at 1.43B, with interactions and a three-way residual absorbing the rest (Table~\ref{tab:three_way_anova}). The narrowing is therefore a genuine property of the 1.43B loss landscape, built from a $\lambda$-sensitivity jump, an $\eta$-optimum drift, and the emergence of a $\lambda \times \eta$ interaction.

\begin{figure}[t]
  \centering
  \begin{minipage}[t]{0.48\linewidth}
    \centering
    \includegraphics[width=\linewidth]{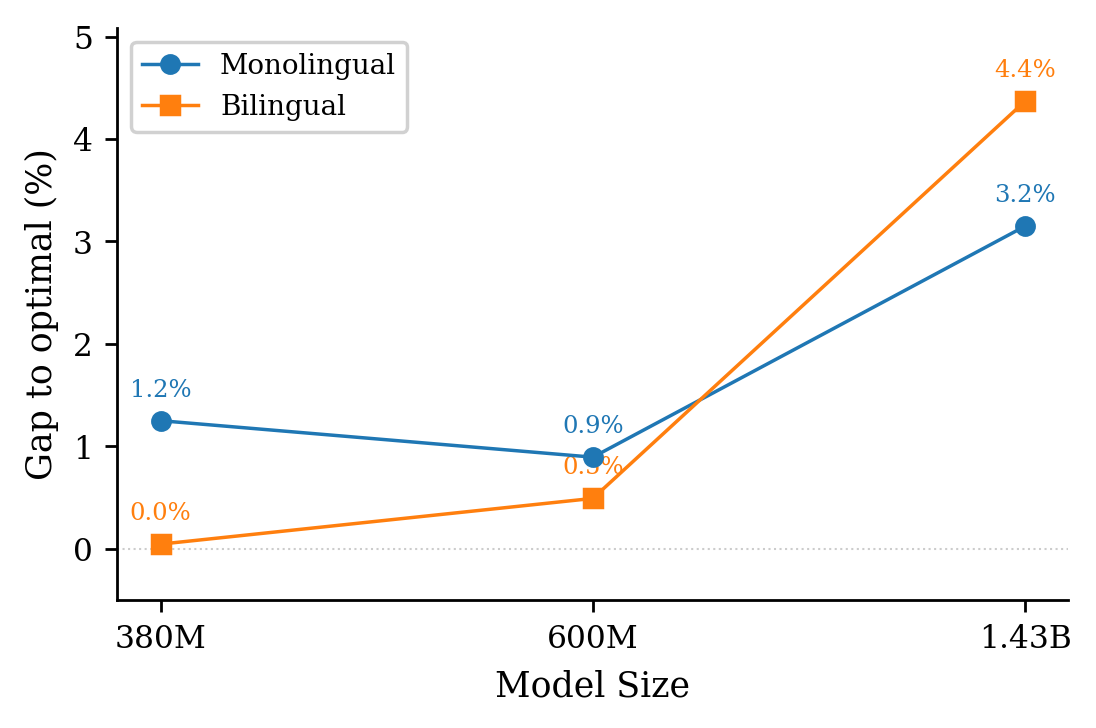}
  \end{minipage}
  \hfill
  \begin{minipage}[t]{0.48\linewidth}
    \centering
    \includegraphics[width=\linewidth]{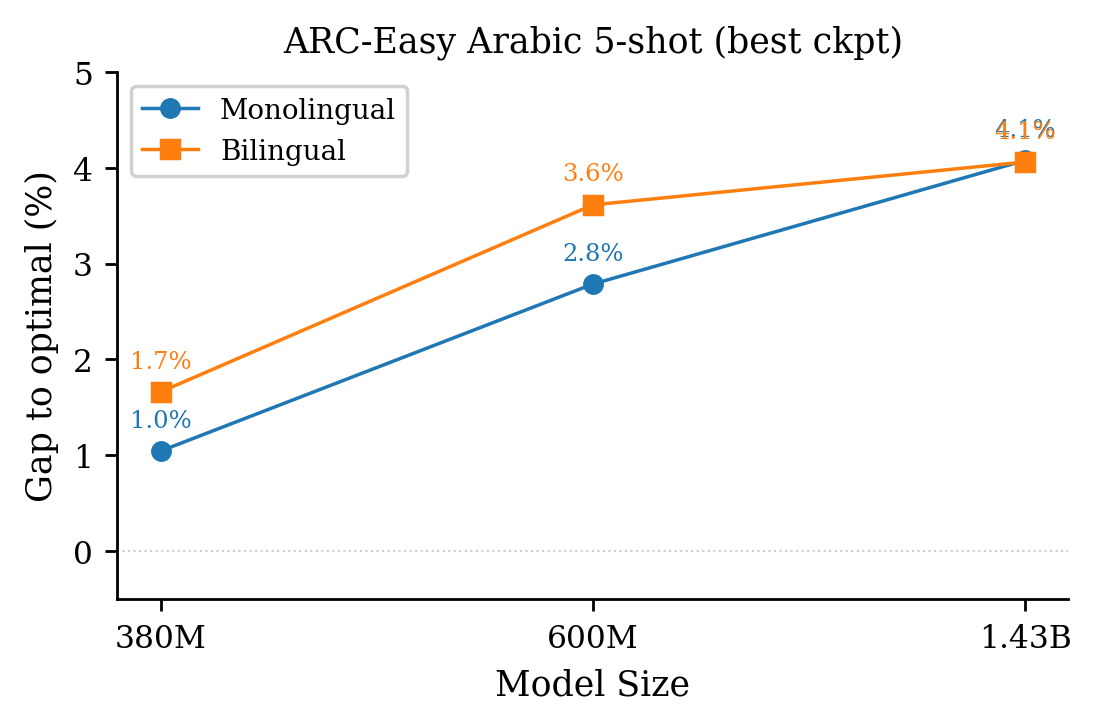}
  \end{minipage}
  \caption{Relative gap between $\mu$P HPs and the per-scale optimum. \textbf{Left}: Arabic validation loss. \textbf{Right}: ARC-Easy Arabic 5-shot accuracy (best checkpoint). The benchmark gap tracks the validation loss gap, confirming that $\mu$P transfer generalizes to downstream performance. In the bilingual setting, the gap is $<$0.6\% on VL at all scales up to 1.43B.}
  \label{fig:mup_transfer_gap}
\end{figure}

The $\mu$P prediction that the optimum sits at a scale-invariant base LR does not hold strictly: the nominal optimum $\eta^\star$ drops from $10^{-2}$ at 380M and 600M to $3 \times 10^{-3}$ at 1.43B (Fig.~\ref{fig:optimal_hp_drift}, left panel). What still holds is the weaker claim that matters in practice: the $\mu$P HP stays inside the near-optimal basin even as that basin shrinks and shifts. Table~\ref{tab:mup_transfer} shows the $\mu$P cell ($\lambda{=}0.01$, $\eta{=}0.01$) is within 0--3.4\% of the per-$R_\text{max}$ best at every scale, and Fig.~\ref{fig:mup_transfer_gap} shows the same gap is below 0.6\% on bilingual Arabic VL at all scales up to 1.43B, with ARC-Easy Arabic 5-shot accuracy tracking the validation-loss gap. Sharpening raises the cost of a \emph{bad} HP choice (extreme $\lambda$ or far-from-optimum $\eta$) but not the cost of the $\mu$P choice specifically. The practical risk at 1.43B is therefore that a wide grid search observes HP-dominated variance as far corners of the grid become pathological, not that $\mu$P transfer itself fails.

\begin{table}[t]
  \caption{Performance of the $\mu$P HP ($\lambda{=}0.01$, $\eta{=}0.01$) vs.\ the per-$R_\text{max}$ best at each scale. Gap is expressed as \% of the best loss. Even though the per-scale optimum drifts with scale (Fig.~\ref{fig:optimal_hp_drift}), the $\mu$P HP stays within a few percent of it; the cost of deviating from the $\mu$P cell grows with scale only at the grid extremes.}
  \label{tab:mup_transfer}
  \centering
  \begin{tabular}{llccc}
    \toprule
    Model & $R_\text{max}$ & Best loss & $\lambda{=}0.01$, $\eta{=}0.01$ & Gap (\%) \\
    \midrule
    380M & 1   & 3.572 & 3.572 & 0.0 \\
    380M & 10  & 3.097 & 3.097 & 0.0 \\
    380M & 50  & 3.065 & 3.151 & 2.8 \\
    \midrule
    600M & 1   & 3.396 & 3.396 & 0.0 \\
    600M & 10  & 2.914 & 2.914 & 0.0 \\
    600M & 50  & 2.897 & 2.962 & 2.2 \\
    600M & 100 & 2.998 & 3.053 & 1.8 \\
    \midrule
    1.43B & 5  & 2.973 & 3.007 & 1.1 \\
    1.43B & 10 & 2.836 & 2.866 & 1.1 \\
    1.43B & 20 & 2.749 & 2.749 & 0.0 \\
    1.43B & 50 & 2.678 & 2.737 & 2.2 \\
    1.43B & 100 & 2.681 & 2.774 & 3.4 \\
    \bottomrule
  \end{tabular}
\end{table}

The practical implication is twofold. First, $\mu$P hyperparameter transfer works: a practitioner who selects $\lambda{=}0.01$, $\eta{=}0.01$ at small scale and transfers it to 1.43B obtains near-optimal performance at every $R_\text{max}$. Second, the risk at larger scales is not that the optimal HP shifts, but that the penalty for choosing a poor HP (especially high weight decay) grows. A practitioner who trusts the $\mu$P HP is safe; a practitioner who runs a wide grid search at 1.43B will observe HP-dominated variance because the far corners of the grid become increasingly pathological.

\FloatBarrier

\section{Data quantity vs.\ hyperparameter variance decomposition}
\label{app:dlr_variance}

The main text shows that data mixing dominates HP tuning in the bilingual setting (\S\ref{sec:hp_sensitivity}, Q1). Here we ask the more direct question: does \emph{raw unique data quantity} dominate HP choice in monolingual training? We answer with two complementary analyses: a three-way ANOVA at 150M with all two-way interactions and a re-centering robustness sweep, at the only scale where we have a full $D_\text{LR} \times \lambda \times \eta$ factorial grid; and an effect-size comparison across all four scales using the partial grids that exist at larger scales. Both follow the ANOVA framework described in App.~\ref{app:anova_method}.

\paragraph{Three-way ANOVA at 150M.}
We run a balanced factorial experiment: 7 Arabic corpus sizes ($D_\text{LR} \in \{25\text{M}, \ldots, 2\text{B}\}$) $\times$ 5 weight decay values $\times$ 5 learning rates = 175 monolingual runs at 150M, each trained to 100 repetitions. We fit a full three-way ANOVA with main effects and all two-way interactions (Type III sums of squares) and decompose Arabic VL variance across the seven resulting terms. Table~\ref{tab:anova_dlr} reports the result.

\begin{table}[h]
  \caption{Three-way ANOVA at 150M on the complete $7 \times 5 \times 5$ grid ($n = 175$, no missing cells): Type III variance decomposition with main effects and all two-way interactions. \emph{Left}: variance fractions (the four terms contributing $<0.1\%$ — $\lambda$, $\lambda \times \eta$, $D_\text{LR} \times \lambda$, three-way$+$noise — are omitted). \emph{Right}: same decomposition as a stacked bar.}
  \label{tab:anova_dlr}
  \centering
  \begin{minipage}[c]{0.30\linewidth}
    \centering
    \begin{tabular}{lc}
      \toprule
      Term & Variance fraction \\
      \midrule
      $D_\text{LR}$             & 58.1\% \\
      $D_\text{LR} \times \eta$ & 37.5\% \\
      $\eta$                     & 4.4\% \\
      \bottomrule
    \end{tabular}
  \end{minipage}%
  \hfill
  \begin{minipage}[c]{0.68\linewidth}
    \centering
    \includegraphics[width=\linewidth]{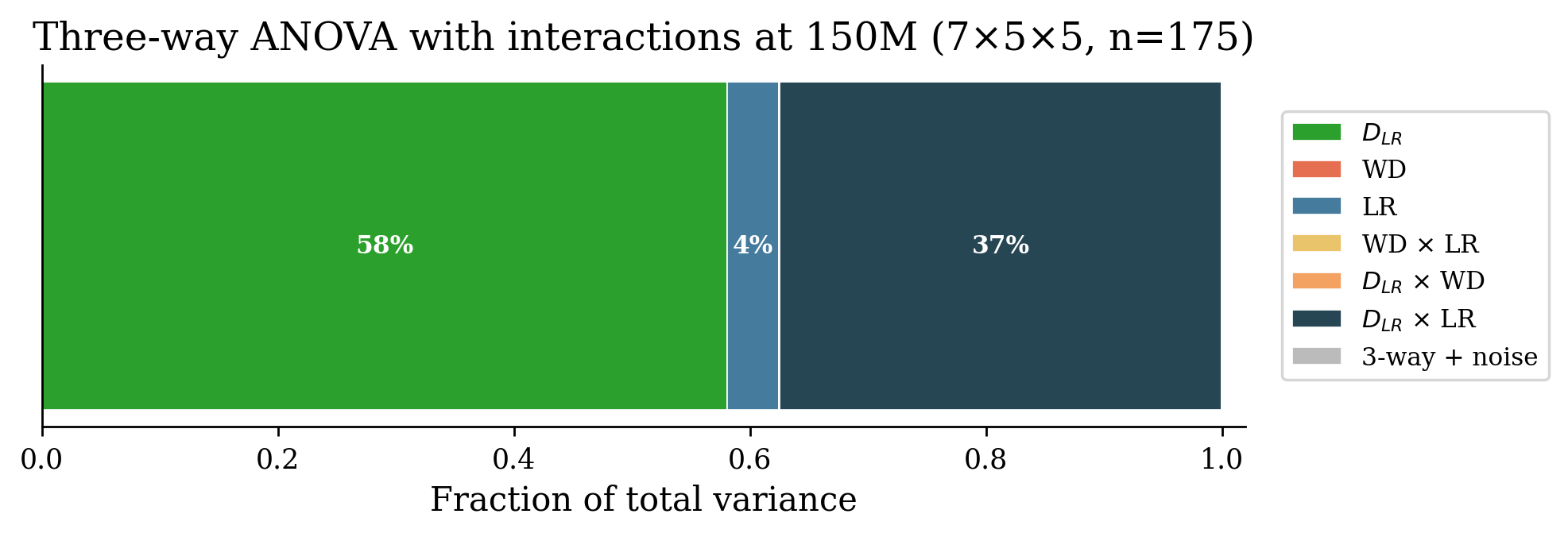}
  \end{minipage}
\end{table}

$D_\text{LR}$ alone explains 58\% of Arabic VL variance, $\eta$ explains 4\%, and a large $D_\text{LR} \times \eta$ interaction contributes another 37\%. The interaction is structured, not residual noise: the effect of $\eta$ depends strongly on $D_\text{LR}$, with $\eta{=}0.0001$ severely undertraining the model at large $D_\text{LR}$ (100 repetitions under a tiny LR) while being less harmful at small $D_\text{LR}$. Weight decay and its interactions contribute essentially nothing. Combining the main effect and its interaction with $\eta$, $D_\text{LR}$ accounts for 96\% of the variance; everything involving $\eta$ alone or $\lambda$ sums to 4\%. The bilingual ANOVA in \S\ref{sec:hp_sensitivity} uses $R_\text{max}$ (repetition count) on the data axis rather than unique Arabic tokens, and is not reported at 150M. With unique Arabic tokens as the data axis, data dominates already at 150M.

\begin{figure}[h]
  \centering
  \includegraphics[width=\linewidth]{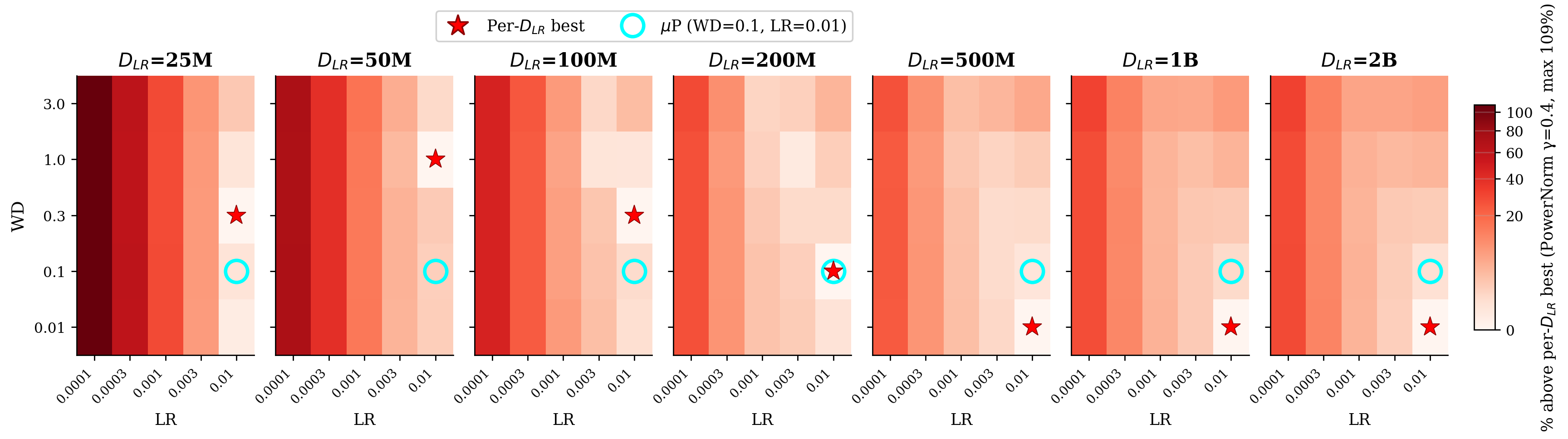}
  \caption{HP landscape at 150M, faceted by Arabic corpus size $D_\text{LR}$. Each panel is the $5 \times 5$ $(\lambda, \eta)$ grid at one $D_\text{LR}$; cell color is \% above the per-$D_\text{LR}$ best (PowerNorm $\gamma{=}0.4$, capped at 100\%). Red star: per-$D_\text{LR}$ best. Cyan circle: $\mu$P HP $(\lambda{=}0.1, \eta{=}0.01)$. The per-$D_\text{LR}$ best $\lambda$ tracks repetition pressure: high $\lambda$ at small $D_\text{LR}$ (heavy repetition), dropping to the smallest grid value $\lambda{=}0.01$ once $D_\text{LR} \geq 500\text{M}$. The dark $\eta{=}0.0001$ column at every $D_\text{LR}$ is the $D_\text{LR} \times \eta$ interaction the ANOVA reports.}
  \label{fig:dlr_hp_compact_150m}
\end{figure}

\begin{figure}[h]
  \centering
  \includegraphics[width=0.7\linewidth]{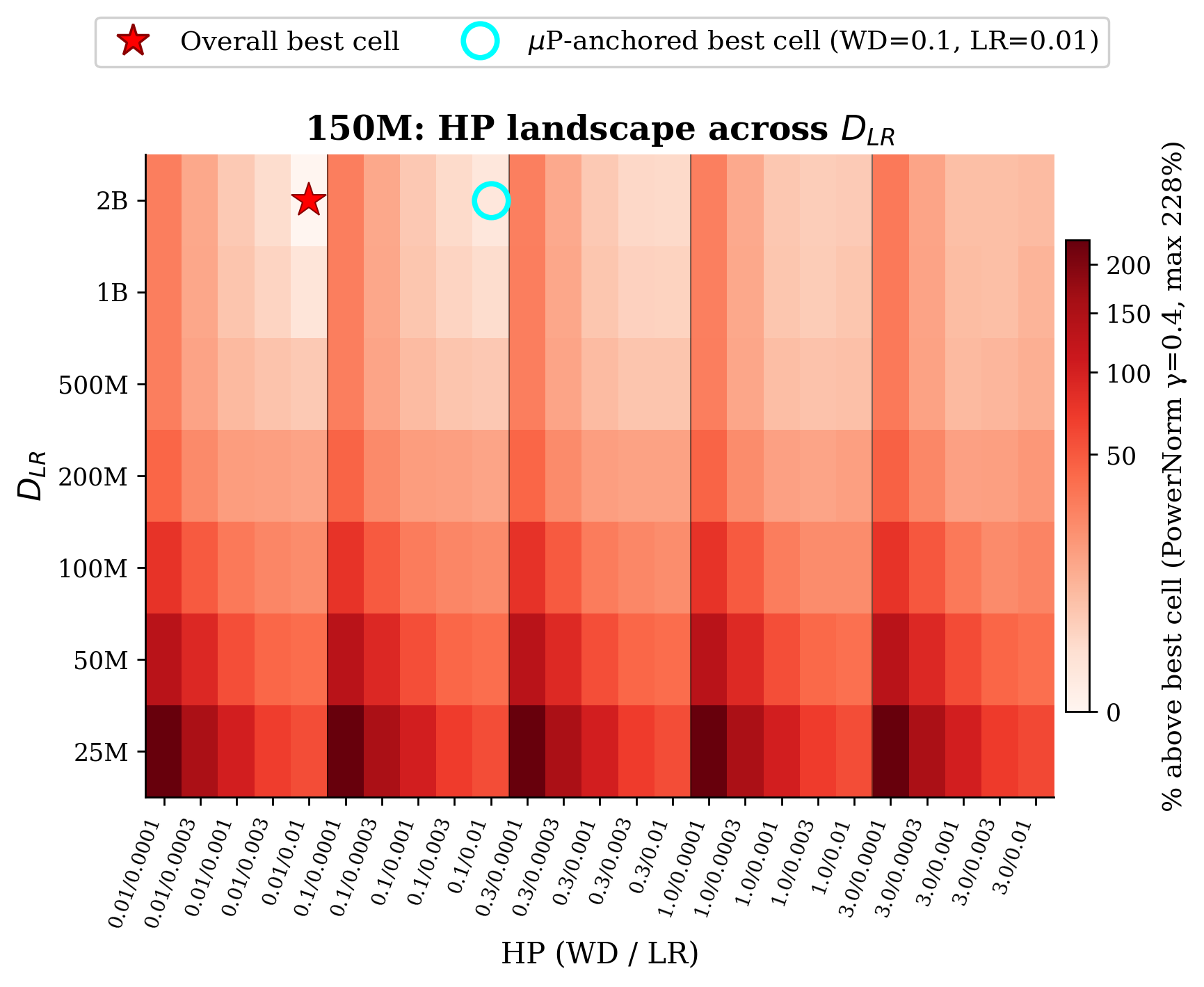}
  \caption{Same data as Fig.~\ref{fig:dlr_hp_compact_150m}, arranged as a single heatmap: $D_\text{LR}$ on the y-axis, the 25 $(\lambda, \eta)$ HP cells on the x-axis. Cell color is \% above the overall best. Red star: overall best at $D_\text{LR}{=}2\text{B}$, $\lambda{=}0.01$, $\eta{=}0.01$. Cyan circle: $\mu$P-anchored best at $\lambda{=}0.1$, $\eta{=}0.01$, also at $D_\text{LR}{=}2\text{B}$.}
  \label{fig:dlr_hp_combined_150m}
\end{figure}

\begin{figure}[h]
  \centering
  \includegraphics[width=\linewidth]{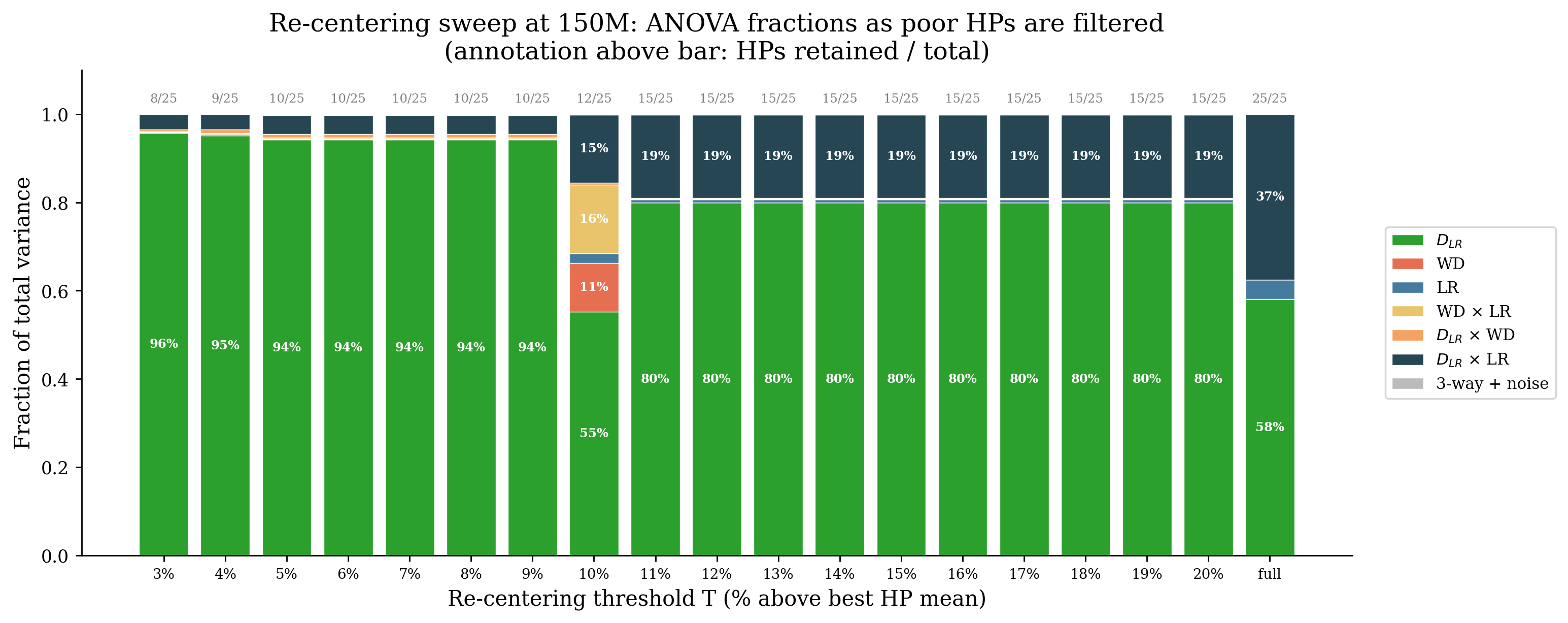}
  \caption{Re-centering sweep at 150M: three-way ANOVA variance fractions as poorly-performing HPs are filtered out. At each threshold $T$ we retain HPs whose marginal mean Arabic VL (averaged over $D_\text{LR}$) is within $T\%$ of the best HP; the right-most ``full'' bar is the unfiltered $7 \times 5 \times 5$ grid. Annotations above each bar: retained HPs / 25 total. $D_\text{LR}$ dominates at every threshold (55--96\%). The $T{=}10\%$ bar is a balanced-grid artifact (see text), not a substantive finding.}
  \label{fig:anova_recentering}
\end{figure}

\paragraph{Re-centering sweep.}
The $D_\text{LR} \times \eta$ interaction concentrates in pathological corners of the HP grid (mainly $\eta{=}0.0001$ paired with large $D_\text{LR}$). We verify $D_\text{LR}$ dominance via a re-centering sweep: for each threshold $T \in \{3\%, \ldots, 20\%\}$, retain only the HPs whose marginal mean Arabic VL (averaged over $D_\text{LR}$) is within $T\%$ of the best HP, then refit the three-way ANOVA on the filtered subgrid. Figure~\ref{fig:anova_recentering} plots the decomposition as a function of $T$.

$D_\text{LR}$ dominance is stable across the sweep: 94--96\% at tight thresholds ($T \leq 9\%$, retaining 8--10 of 25 HPs near the per-axis optimum), 80\% at moderate thresholds ($T \geq 11\%$, retaining 15 of 25 HPs), and 58\% on the full grid. The complement is the $D_\text{LR} \times \eta$ interaction, which grows as more pathological $\eta$ settings re-enter the subgrid. The $T{=}10\%$ point is a balanced-factorial artifact rather than a finding: the cutoff retains 12 of 25 HPs but only 2 of 5 weight decays at $\eta{=}0.001$, so the filtered subgrid is a staircase rather than a rectangle, and Type III SS cannot cleanly separate $\lambda$ from $\eta$ under this geometry (spurious $\lambda$ and $\lambda \times \eta$ mass appears). At $T{=}11\%$ the subgrid becomes rectangular again and the decomposition returns to $D_\text{LR}=80\%$.

\paragraph{Effect-size comparison across scales.}
At 380M, 600M, and 1.43B we lack a full $D_\text{LR} \times \lambda \times \eta$ grid, so a three-way ANOVA is not available. We substitute a model-free proxy for variance dominance: the ratio of the Arabic VL range along the data axis to the Arabic VL range along the HP axis. Let $\ell(D, \lambda, \eta; N)$ denote the best Arabic validation loss for a model of size $N$ trained on $D$ unique Arabic tokens with weight decay $\lambda$ and learning rate $\eta$, and let $\mathcal{H}_T(N)$ denote the set of HPs $(\lambda, \eta)$ whose marginal mean VL (averaged over $D$) lies within a fraction $T$ of the best HP's marginal mean, matching the re-centering used for the ANOVA above. Define
\begin{align}
  \operatorname{Range}_D(N)          &= \max_{D}\, \ell\bigl(D, \lambda^\star(D), \eta^\star(D); N\bigr) - \min_{D}\, \ell\bigl(D, \lambda^\star(D), \eta^\star(D); N\bigr), \label{eq:range_d} \\
  \operatorname{Range}_\text{HP}(N; T) &= \max_{(\lambda, \eta) \in \mathcal{H}_T(N)}\, \ell(200\text{M}, \lambda, \eta; N) - \min_{(\lambda, \eta) \in \mathcal{H}_T(N)}\, \ell(200\text{M}, \lambda, \eta; N), \label{eq:range_hp}
\end{align}
where $D \in \{25\text{M}, 50\text{M}, 100\text{M}, 200\text{M}, 500\text{M}, 1\text{B}, 2\text{B}\}$ with per-size tuned HP $(\lambda^\star(D), \eta^\star(D))$ and $(\lambda, \eta)$ ranges over the $5 \times 5$ grid at fixed $D = 200\text{M}$. Taking $T = \infty$ recovers the full HP grid (no filtering). The data-axis dominance ratio
\begin{align}
  \rho(N; T) = \frac{\operatorname{Range}_D(N)}{\operatorname{Range}_\text{HP}(N; T)} \label{eq:dominance_ratio}
\end{align}
compares the two effect sizes: $\rho > 1$ means sweeping the data axis moves Arabic VL more than sweeping the (post-thresholding) HP axis. Two asymmetries bias $\rho$ downward, against our claim: (i) with up to 25 cells on the HP side vs.\ 7 on the data side, the HP range has more opportunity to reach an extreme loss, and (ii) the data-axis HPs are retuned per $D$, keeping every point near its local optimum and compressing $\operatorname{Range}_D$. Any observed $\rho > 1$ is therefore a conservative lower bound on the true data-axis dominance.

\begin{table}[h]
  \caption{Data-axis dominance ratio $\rho(N; T)$ at each model scale, on the full HP grid and after re-centering at $T \in \{10\%, 20\%\}$; $n_\text{HP}$ is the number of HPs retained out of 25. On the full grid the ratio drops at 1.43B; after re-centering, dominance is stable at $\sim 5\times$ and is largest at 1.43B.}
  \label{tab:effect_size}
  \centering
  \begin{tabular}{lccccc}
    \toprule
    & & & \multicolumn{3}{c}{$\rho(N; T)$} \\
    \cmidrule(lr){4-6}
    Model & $\operatorname{Range}_D$ & $\operatorname{Range}_\text{HP}$ (full) & Full grid & $T{=}10\%$ ($n_\text{HP}$) & $T{=}20\%$ ($n_\text{HP}$) \\
    \midrule
    150M  & 1.71 & 0.96 & 1.8$\times$ & 5.1$\times$ (20) & 5.1$\times$ (20) \\
    380M  & 1.59 & 0.45 & 3.6$\times$ & 5.1$\times$ (22) & 3.6$\times$ (25) \\
    600M  & 1.66 & 0.31 & 5.4$\times$ & 5.4$\times$ (25) & 5.4$\times$ (25) \\
    1.43B & 1.90 & 0.97 & 2.0$\times$ & 6.7$\times$ (15) & 5.0$\times$ (24) \\
    \bottomrule
  \end{tabular}
\end{table}

\begin{figure}[h]
  \centering
  \includegraphics[width=0.85\linewidth]{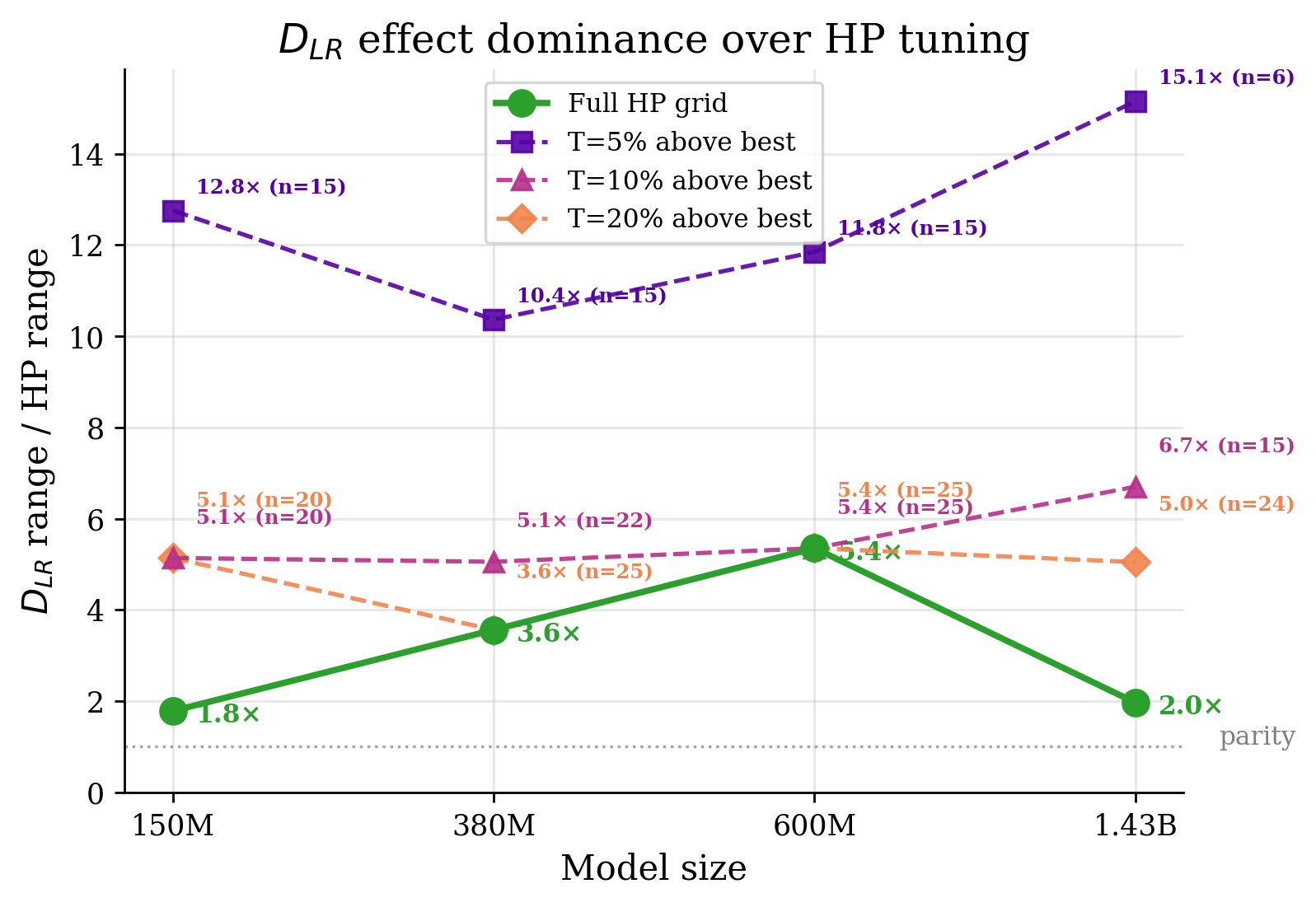}
  \caption{Data-axis dominance ratio $\rho(N; T) = \operatorname{Range}_D(N) / \operatorname{Range}_\text{HP}(N; T)$ by model scale (Eqs.~\ref{eq:range_d}--\ref{eq:dominance_ratio}). Green solid ($T = \infty$): full HP grid. Purple, pink, orange (dashed): HP range recomputed after retaining HPs within $T \in \{5\%, 10\%, 20\%\}$ of the best. Annotations: $\rho$ value and $n_\text{HP}$. The full-grid dip at 1.43B is absent from every re-centered curve; once pathological HPs are filtered, $\rho$ is stable ($\sim 5\times$) across scales and largest at 1.43B, consistent with 1.43B having the sharpest HP landscape rather than the weakest data effect.}
  \label{fig:effect_size}
\end{figure}

On the full grid, $\rho$ rises from $1.8\times$ at 150M to $5.4\times$ at 600M and then drops to $2.0\times$ at 1.43B. After re-centering, the drop at 1.43B disappears: at $T = 10\%$, $\rho$ is in the range $5.1\text{--}6.7\times$ at every scale and is in fact \emph{largest} at 1.43B ($6.7\times$), reaching $15.1\times$ at $T = 5\%$. The full-grid drop was therefore driven by HP pathologies specific to 1.43B, where only 15 of 25 HPs survive a 10\% cutoff (vs.\ 20--25 at smaller scales): the HP range is inflated by a small number of bad cells rather than by a broadly sharper landscape. Data-axis dominance at 1.43B is not a counterexample to the scaling trend but a consequence of the same pathology flagged in \S\ref{sec:hp_sensitivity} (Q3). Figure~\ref{fig:effect_size} plots $\rho(N; T)$ for $T \in \{5\%, 10\%, 20\%, \infty\}$.

\section{Concurrent work}
\label{app:concurrent}

\citet{anonymous2026mixturepretraining} study mixture pre-training under data constraints, deriving a repetition-aware scaling law that predicts target-domain loss as a joint function of model size, total compute, target-pool size, and mixture weight, fitted at small scales and extrapolated to larger ones. They validate the law across more than 2000 runs spanning multilingual (German, French, Swahili), multi-domain (mathematics, scientific papers, Wikipedia), and quality-filtered settings at 101M--936M parameters, and use it to recover optimal repetition counts that reach 15--20 for the constrained source. Our paper addresses the prior question: comparing data mixing against hyperparameter tuning (weight decay, learning rate) as competing strategies for data-constrained pre-training, and quantifying mixing's advantage over tuning in equivalent unique target tokens across model scales. We further show that target-language validation loss systematically underestimates mixing's downstream value: the gap between validation loss and benchmark accuracy widens with scale, with mixing yielding the equivalent of $2$--$3\times$ more unique target data on validation loss but $2$--$13\times$ on downstream task accuracy.
   
\section{Limitations}
\label{app:limitations}

\paragraph{Single benchmark.}
All downstream claims rest on a single benchmark, ARC Easy, evaluated in both English and Arabic. We chose ARC Easy because monolingual models at 150M--1.43B score measurably above the random baseline on it, while pilot evaluations on MMLU and Arabic MMLU at the same scales remained at random across the full grid. Our data-multiplier analysis (\S\ref{sec:data_equiv}) fits a log-linear curve through the monolingual points and inverts it at the bilingual target; this fit requires signal above random, which MMLU-class benchmarks did not deliver at our scales. ARC Easy is also knowledge-intensive. Benchmarks that emphasize target-language syntax, generative fluency, or culturally specific knowledge may show different multiplier magnitudes, and our results do not speak to them. The natural follow-up is to scale the evaluation in benchmark difficulty by running stronger training recipes on natively authored Arabic benchmarks such as Arabic MMLU, Belebele-AR, or AlGhafa, on which our current models cannot yet produce usable signal. Our scope here is pre-training dynamics, for which a stable benchmark signal across scales was the primary requirement; extending to natively authored evaluations at stronger scales is left to future work.

\paragraph{Translated evaluation set.}
The Arabic version of ARC Easy is machine-translated from English with NLLB-200-3.3B \citep{nllbteam2022languageleftbehindscaling}. Translated text retains English-aligned syntactic and statistical artifacts (``translationese''), and since bilingual models are trained on English, an artifact-driven component of the bilingual advantage cannot be ruled out from our data alone. 
A natively authored Arabic benchmark would still separate target-language capability from translation effects more cleanly than our current design; we flag this as the cleanest follow-up experiment.

\paragraph{Single language pair.}
We evaluate a single pair, Arabic--English. These two languages are typologically distant (different language families, scripts, and morphology), which makes our measurements a relatively hard case for cross-lingual transfer; we expect closer pairs such as German--English or French--English to show effects at least as strong. Extending the evaluation to additional pairs is a natural follow-up. A second open question is the tokenizer: we use the XGLM tokenizer, which has strong subword coverage on Arabic, English, and a broad set of other languages. Its shared vocabulary favors Arabic--English transfer, so the multiplier magnitudes we report likely absorb a tokenizer-specific contribution. Studying other tokenizers, in particular on very low-resource languages where subword coverage is weaker, would isolate how much of the mixing advantage survives under less favorable tokenization.

\paragraph{Static mixing ratios.}
We restrict to static mixing ratios; dynamic schedules that shift the ratio across phases are an orthogonal axis we leave to future work.

\paragraph{Single-seed runs.}
Each cell of our grid is trained once. We ran approximately 1000 models in total across the four scales; replicating the grid across multiple seeds would have multiplied total compute beyond our budget, and we do not report error bars across seeds. As a partial noise bound, the ANOVA in App.~\ref{app:hp_landscape} reports residual variance at each scale: the two-way residual is 4--6\% across 380M, 600M, and 1.43B, and the three-way residual stays below 7\% at 380M and 600M. At 1.43B the three-way residual is larger but is driven by genuine interaction structure between $R_\text{max}$, $\lambda$, and $\eta$ rather than by seed noise (\S\ref{app:hp_landscape}). Full seed replication is the cleaner measurement and is left to future work.

\paragraph{HP grid design.}
Our $5 \times 5$ HP grid was fixed across all model scales, and the per-scale optimum frequently lands on grid boundaries (App.~\ref{app:operational_sensitivity}): $\eta = 0.01$ (the largest grid value) at smaller scales, and $\lambda = 0.01$ (the smallest grid value) at every scale. The true optimum may therefore lie outside our grid, and the HP sensitivity we measure at 1.43B is partly a property of the loss landscape and partly an artifact of the grid being off-center. A more informative analysis would have used a per-scale grid centered on the $\mu$P-predicted HP, with sufficient range in both directions to bracket the per-scale optimum. We recommend this design for follow-up work.

\section{Compute resources}
\label{app:compute}

We trained approximately 1000 models in total across the four scales. Each run was performed on a single GPU node, using either NVIDIA A100 or H100 GPUs. For the monolingual 20B-token runs, wall-clock times on A100 were 18.8 hours at 150M, 35.9 hours at 380M, 51.0 hours at 600M, and 95.0 hours at 1.43B. Bilingual runs varied in token budget across scales: 2.28B tokens in 2.7 hours (A100) at 150M, 12.17B tokens in 19.6 hours (A100) at 380M, 28.51B tokens in 62.0 hours (A100) at 600M, and 97.31B tokens in 191.0 hours (H100) at 1.43B.

\applefootnote{\textcolor{textgray}{\sffamily Apple and the Apple logo are trademarks of Apple Inc., registered in the U.S. and other countries and regions.}}

\end{document}